\documentclass[runningheads]{llncs}
\usepackage{graphicx}
\usepackage{amsmath,amssymb} 
\usepackage{color}
\usepackage{hhline}
\usepackage{appendix}
\usepackage{multirow}
\usepackage{adjustbox}
\usepackage{booktabs}
\usepackage{cite}
\usepackage[table]{xcolor}
\usepackage[linesnumbered,ruled]{algorithm2e}
\usepackage[pagebackref=true,breaklinks=true,letterpaper=true,colorlinks,bookmarks=false]{hyperref}
\usepackage[width=122mm,left=12mm,paperwidth=146mm,height=193mm,top=12mm,paperheight=217mm]{geometry}

\hypersetup{linkcolor=[rgb]{0.8551,0.2333,0.2333}}
\hypersetup{citecolor=[rgb]{0.3333,0.2333,0.7551}}

\DeclareMathOperator*{\argmax}{arg\,max}

\definecolor{city_color_1}{rgb}{0.5020,0.2510,0.5020}
\definecolor{city_color_2}{rgb}{0.9569,0.1373,0.9098}
\definecolor{city_color_3}{rgb}{0.2745,0.2745,0.2745}
\definecolor{city_color_4}{rgb}{0.4000,0.4000,0.6118}
\definecolor{city_color_5}{rgb}{0.7451,0.6000,0.6000}
\definecolor{city_color_6}{rgb}{0.6000,0.6000,0.6000}
\definecolor{city_color_7}{rgb}{0.9804,0.6667,0.1176}
\definecolor{city_color_8}{rgb}{0.8627,0.8627,0.0000}
\definecolor{city_color_9}{rgb}{0.4196,0.5569,0.1373}
\definecolor{city_color_10}{rgb}{0.5961,0.9843,0.5961}
\definecolor{city_color_11}{rgb}{0.2745,0.5098,0.7059}
\definecolor{city_color_12}{rgb}{0.8627,0.0784,0.2353}
\definecolor{city_color_13}{rgb}{1.0000,0.0000,0.0000}
\definecolor{city_color_14}{rgb}{0.0000,0.0000,0.5569}
\definecolor{city_color_15}{rgb}{0.0000,0.0000,0.2745}
\definecolor{city_color_16}{rgb}{0.0000,0.2353,0.3922}
\definecolor{city_color_17}{rgb}{0.0000,0.3137,0.3922}
\definecolor{city_color_18}{rgb}{0.0000,0.0000,0.9020}
\definecolor{city_color_19}{rgb}{0.4667,0.0431,0.1255}

\begin{document}
\title{Domain Adaptation for Semantic Segmentation via Class-Balanced Self-Training}
\titlerunning{Domain Adaptation via Class-Balanced Self-Training}

\author{
Yang Zou\inst{1}$^{\star}$\and 
Zhiding Yu\inst{2}\thanks{indicates equal contribution.} \and
B.V.K. Vijaya Kumar\inst{1} \and 
Jinsong Wang\inst{3}
}
\authorrunning{Yang Zou$^{\star}$, Zhiding Yu$^{\star}$, B.V.K. Vijaya Kumar, Jinsong Wang}

\institute{
Carnegie Mellon University, Pittsburgh, PA 15213\\
\email{yzou2@andrew.cmu.edu, kumar@ece.cmu.edu} \and
NVIDIA, Santa Clara, CA 95051~~\email{zhidingy@nvidia.com} \and
General Motors R \& D, Warren, MI 48092~~\email{jinsong.wang@gm.com}\\
{[\href{https://github.com/yzou2/cbst}{GitHub}]~~
 [\href{https://yzou2.github.io/project/self-training/}{Project Page}]~~
 [\href{https://yzou2.github.io/pdf/CBST_slides.pdf}{Slides}]~~
 [\href{https://yzou2.github.io/pdf/CBST_poster.pdf}{Poster}]}
}

\maketitle
\begin{abstract}
Recent deep networks achieved state of the art performance on a variety of semantic segmentation tasks. Despite such progress, these models often face challenges in real world ``wild tasks'' where large difference between labeled training/source data and unseen test/target data exists. In particular, such difference is often referred to as ``domain gap'', and could cause significantly decreased performance which cannot be easily remedied by further increasing the representation power. Unsupervised domain adaptation (UDA) seeks to overcome such problem without target domain labels. In this paper, we propose a novel UDA framework based on an iterative self-training procedure, where the problem is formulated as latent variable loss minimization, and can be solved by alternatively generating pseudo labels on target data and re-training the model with these labels. On top of self-training, we also propose a novel class-balanced self-training framework to avoid the gradual dominance of large classes in pseudo-label generation, and introduce spatial priors to refine the generated pseudo-labels. Comprehensive experiments show that the proposed methods achieve state of the art semantic segmentation performance under multiple major UDA settings.
\end{abstract}
\section{Introduction}\label{introduction}
\begin{figure}[!t]
 \centering
\resizebox{1.0\textwidth}{!}{
\begin{tabular}{@{}cccccccccc@{}}
\cellcolor{city_color_1}\textcolor{white}{~~road~~} &
\cellcolor{city_color_2}~~sidewalk~~&
\cellcolor{city_color_3}\textcolor{white}{~~building~~} &
\cellcolor{city_color_4}\textcolor{white}{~~wall~~} &
\cellcolor{city_color_5}~~fence~~ &
\cellcolor{city_color_6}~~pole~~ &
\cellcolor{city_color_7}~~traffic lgt~~ &
\cellcolor{city_color_8}~~traffic sgn~~ &
\cellcolor{city_color_9}~~vegetation~~ \\
\cellcolor{city_color_10}~~terrain~~ &
\cellcolor{city_color_11}~~sky~~ &
\cellcolor{city_color_12}\textcolor{white}{~~person~~} &
\cellcolor{city_color_13}\textcolor{white}{~~rider~~} &
\cellcolor{city_color_14}\textcolor{white}{~~car~~} &
\cellcolor{city_color_15}\textcolor{white}{~~truck~~} &
\cellcolor{city_color_16}\textcolor{white}{~~bus~~} &
\cellcolor{city_color_17}\textcolor{white}{~~train~~} &
\cellcolor{city_color_18}\textcolor{white}{~~motorcycle~~} &
\cellcolor{city_color_19}\textcolor{white}{~~bike~~}
\end{tabular}
}
  \includegraphics[width=\linewidth]{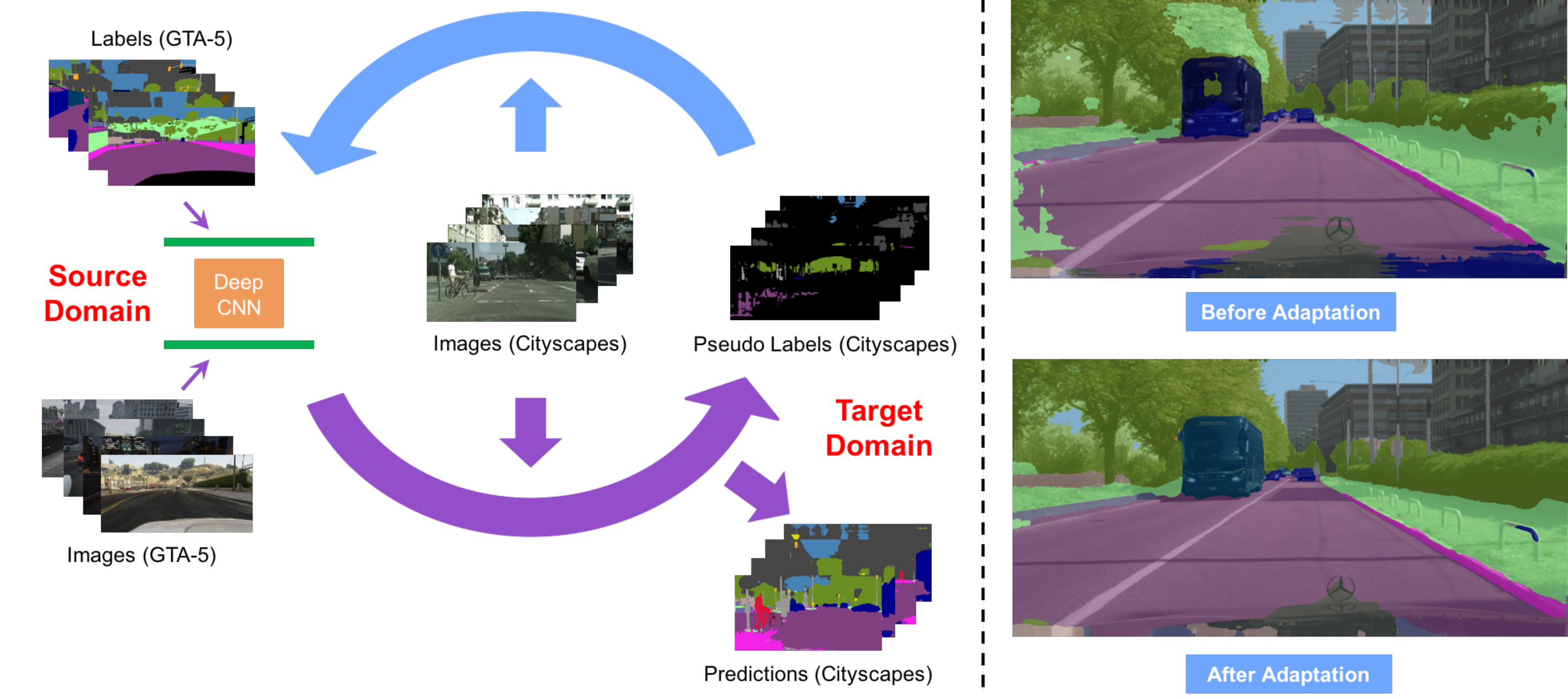}
  \caption{Illustration of the proposed itertive self-training framework for unsupervised domain adaptation. Left: algorithm workflow. Right figure: semantic segmentation results on Cityscapes before and after adaptation.}
  \label{flow}
\end{figure}

Semantic segmentation is a core computer vision task where one aims to densely assign labels to each pixel in the input image. In the past decade, significant amount of effort has been devoted to this area~\cite{cordts2016cityscapes,geiger2012we,zhou2016semantic,Chen_2017_ICCV,nexar,long2015fully,Zhao_2017_CVPR,wu2016wider,chen2018deeplab,chen2017rethinking,wang2017understanding}, leading to considerable progress with the recent advance of deep representation learning ~\cite{krizhevsky2012imagenet, Simonyan15, he2016deep}. The competition on major open benchmark datasets~\cite{cordts2016cityscapes} have resulted in a number of more powerful models that tend to overfit to the benchmark data. While the boundaries of benchmark performance have been pushed to new limits, these models often encounter challenges in practical applications such as autonomous driving, where one needs ubiquitous good performance of the perception module. This is because benchmark datasets are usually biased to specific environments, while the testing scenario may encounter large domain differences caused by a number of factors, including change of geological position, illumination, camera, weather condition, etc. In this case, even the performance of a powerful model often drops dramatically, and such issue can not be easily remediated by further building up the model power \cite{Chen_2017_ICCV,hoffman2016fcns,hoffman2018cycada}.

A natural idea to improve network's generalization ability is to collect and annotate data covering more diverse scenes. However, densely annotating image is time-consuming and labor-intensive. An example is the Cityscapes dataset, where each image on average takes about 90 minutes to annotate~\cite{cordts2016cityscapes}. To overcome such limitation, efforts were made to efficiently generate densely annotated images from rendered scenes, such as the SYNTHIA Dataset~\cite{ros2016synthia}, the Grand Theft Auto V (GTA5) Dataset~\cite{richter2016playing} and the VIPER Dataset~\cite{richter2017playing}. However, the large appearance gap across the simulated domain and the real domain can significantly degrade the performance of synthetically trained models.

In light of the above issues, we focus on the challenging problem of unsupervised domain adaptation for semantic segmentation in this paper, aiming to unsupervisedly adapt a segmentation model trained on a labeled source domain to a target domain without knowing target labels. Recently, the problem of unsupervised domain adaptation has been widely explored on classification/segmentation/detection tasks. There exist a predominant trend to use adversarial training based methods to match image-level/feature-level/prediction-level distributions of both source and target domains~\cite{ganin2016domain,bousmalis2016domain,hoffman2016fcns,Chen_2017_ICCV,sankaranarayanan2017unsupervised, tsai2018learning}. In particular, these methods aim to minimize a domain adversarial loss to reduce the discrepancy between source and target feature distributions, while retaining good performance on source domain task by minimizing the task-specific loss.

While adversarial training based methods have recently achieved great success in domain adaptation, in this work we show that comparable or even better adaptation performance can be achieved by taking an alternative way without adversarial training. Our proposed method is a self-training based learning framework where one predicts in the target domain and in turn uses the predictions to update the model. In this way, class-wise feature space alignment and task-specific learning are implicitly unified together under a single unified loss. This differs from adversarial training based methods which seek to adapt by confusing the domain discriminator, with domain alignment standalone from task-specific learning under a separate loss.

The idea of self-training is not new. Traditional self-training methods have been commonly used in semi-supervised learning (SSL) problems as a framework towards learning better classifier decision boundaries with unlabeled data~\cite{grandvalet2005semi}. However, most of these methods deal with handcrafted features which are fixed inputs to classifiers. A subtle difference between these methods and deep self-training is that the latter allows learnable/flexible deep embeddings as input to a classifier. Upon minimizing the self-training loss, the limited representation power of classifier not only leads to better decision boundaries, but also drives the feature embeddings across domains to be similar. As a result, deep self-training can become a powerful class-wise domain adaptation method. The method also coincides with the recent trend of class-wise/conditional adversarial domain adaptation~\cite{Chen_2017_ICCV} in the sense that they all can be broadly regarded as EM-like adaptation frameworks with network predictions being latent variables.

To the best of our knowledge, this is one of the early works applying deep self-training to the recent tasks of domain adaptation for semantic segmentation. We propose a deep self-training (ST) framework of which the workflow is shown in Figure~\ref{flow} with GTA5 $\rightarrow$ Cityscapes being an example. Self-training is performed by alternately generating a set of pseudo-labels corresponding to confident\footnote{In this paper, we define ``confidence'' as the maximum class probability at each pixel.} predictions in the target domain, and then fine-tuning the network model based on these pseudo-labels together with the labeled source data. Note that such framework implicitly assumes that target samples with more confident predictions tend to have higher prediction accuracies.

Vanilla ST treats every class equally when measuring the prediction confidence. However, the domain gaps caused by appearance difference and scale difference can vary significantly across different classes. For instance, different countries may have quite dissimilar construction layouts, but traffic lights and vehicles are often more or less similar. As a result, it is harder for a source models to transfer knowledge on the construction class than on traffic light and vehicle. Such issue leads to different difficulties for a network to learn transferable knowledge as well as non-unified confidence levels for various classes in the target domain. We observe vanilla ST tends to bias towards easy classes while ignoring other difficult ones, since ST universally chooses pseudo-labels with high confidence. The problem causes decreased adaptation performance as pseudo-labels of the difficult classes diminish during training. To address this problem, we also propose a class-balanced self-training (CBST) framework. Our contributions in this paper can be summarized as follows:

\begin{itemize}
  \item We introduce a deep self-training framework with iterative self-paced learning policy towards domain adaptation. Specifically, we treat the unknown target domain labels as latent variables (pseudo-labels), and formulate pseudo-label estimation and network training as a unified loss minimization problem which can optimized end-to-end via mixed integer programming.
  \item To address the issue of imbalanced pseudo-labels in ST, we propose a novel class-balanced self-training framework towards more balanced pseudo-label generation. The proposed framework performs class-wise confidence normalization, by dividing the confidence of each pseudo-label with a rank-based reference confidence from that particular class.
  \item We also observe that traffic scenes often share similar spatial layouts. As a result we also introduce spatial priors (SP) to improve cross-domain adaptation. We incorporate spatial priors into the proposed CBST framework, leading to CBST with spatial priors (CBST-SP).
  \item Our approaches are comprehensively evaluated under both synthetic-to-real (SYNTHIA/GTA5 to Cityscapes) and real-to-real (Cityscapes-to-NTHU) settings with currently state-of-the-art performance.
\end{itemize}

\section{Related works}
\noindent\textbf{Domain adaptation:} UDA problems have been widely investigated for their importance in a number of real world tasks. A major idea to perform domain adaption is to learn domain invariant embeddings by reducing the difference between source and target domain distributions~\cite{tzeng2014deep,tzeng2015simultaneous,long2015learning,sun2016deep}. Among them, Maximum Mean Discrepancy (MMD) and its kernel variants has been a popular target towards minimizing the cross-domain difference of feature distributions. More recently, there has been an increasing interest in using adversarial training based methods to reduce the domain gaps~\cite{ajakan2014domain,tzeng2014deep,ganin2015unsupervised,ganin2016domain,Adda_CVPR2017}.

\noindent\textbf{Self-training for SSL:} Self-training methods have been widely studied in semi-supervised learning~\cite{chapelle2009semi,zhu2006semi}, with applications to vision and natural language processing~\cite{yarowsky1995unsupervised,riloff2003learning,maeireizo2004co,zhu2006semi}. Given the inherent relation between UDA and SSL in their forms, Tang et al.~\cite{tang2012shifting} addressed image-to-video detector adaptation, by incorporating self-paced sample selection into self-training with a weight shifting policy to gradually increase target domain samples. Chen et al.~\cite{chen2011co} proposed a variant of co-training~\cite{blum1998combining} for domain adaptation, by jointly learning target predictor, view split, and subset selection in a single optimization problem.

\noindent\textbf{Semantic segmentation:} Recent advances in deep learning have aroused broad interests in semantic segmentation using convolutional neural networks (CNNs). Long et al.~\cite{long2015fully} proposed fully convolutional network (FCN) towards pixel-level dense prediction. Since then, several powerful segmentation networks have been proposed, including DeepLab v2/v3~\cite{chen2018deeplab, chen2017rethinking}, ResNet-38~\cite{wu2016wider}, PSPNet~\cite{Zhao_2017_CVPR} etc.

\noindent\textbf{UDA for Semantic segmentation:} It is commonly observed that the representation power of a segmentation model does not transfer well to its cross-domain performance. As a result, domain adaptation for semantic segmentation recently emerged as a hot topic. A number of adaptation methods are proposed, including adversarial training at input image level~\cite{hoffman2018cycada}, feature level~\cite{hoffman2016fcns,Chen_2017_ICCV,hoffman2018cycada}, and network output level~\cite{tsai2018learning}. Specifically,~\cite{hoffman2018cycada} seeks to reduce domain gap by first transferring source images to target style with a cycle consistency loss, and then aligning the cross-domain feature distributions of the task network through adversarial training. In addition,~\cite{saito2017adversarial} propose to detect non-discriminative samples near decision boundaries through a critic network, and let the generator learn to generate more discriminative features by fooling the critic network with adversarial training. \cite{Zhang_2017_ICCV} proposed a curriculum adaption method to regularize the distributions of predicted labels in the target domain such that they follow the label distributions in source domain. 

\section{Preliminaries}
\subsection{Fine-tuning for supervised domain adaptation}
Assuming that the labels in both source and target are available, possibly the most direct way to perform domain adaptation is supervised fine-tuning the model on both domains. For semantic segmentation nets with softmax output, the adaptation problem can be formulated as minimizing the following loss function:
\begin{equation}
\begin{aligned}\label{finetune}
\min_{\mathbf{w}}\mathcal{L}_S(\mathbf{w})
= -\sum_{s=1}^{S}\sum_{n=1}^{N}\mathbf{y}_{s,n}^\top\log (\mathbf{p}_n(\mathbf{w},\mathbf{I}_s)) -\sum_{t=1}^{T}\sum_{n=1}^{N}\mathbf{y}_{t,n}^\top\log (\mathbf{p}_n(\mathbf{w},\mathbf{I}_t))
\end{aligned}
\end{equation}
where $\mathbf{I}_s$ denotes the image in source domain indexed by $s = 1,2,...,S$, $\mathbf{y}_{s,n}$ the ground truth label for the $n$-th pixel ($n = 1,2,...,N$) in $I_s$, and $\mathbf{w}$ contains the network weights. $\mathbf{p}_n(\mathbf{w},\mathbf{I}_s)$ is the softmax output containing the class probabilities at pixel $n$. Similar definitions apply for $\mathbf{I}_t$, $\mathbf{y}_{t,n}$ and $\mathbf{p}_n(\mathbf{w},\mathbf{I}_t)$.

\subsection{Self-training for unsupervised domain adaptation}
In the case of unsupervised domain adaptation, the target ground truth labels are not available. An alternate way to fine-tune the segmentation model is to consider the target labels as hidden variables that can be learned. Accordingly, the problem can be formulated as follows:
\begin{equation}\label{uda}
\begin{aligned}
& \min_{\mathbf{w},\hat{\mathbf{y}}}\mathcal{L}_{U}(\mathbf{w},\hat{\mathbf{y}}) = -\sum_{s=1}^{S}\sum_{n=1}^{N}\mathbf{y}_{s,n}^\top\log (\mathbf{p}_n(\mathbf{w},\mathbf{I}_s)) - \sum_{t=1}^{T}\sum_{n=1}^{N}\hat{\mathbf{y}}_{t,n}^\top\log (\mathbf{p}_n(\mathbf{w},\mathbf{I}_t))\\
& ~s.t. ~ \hat{\mathbf{y}}_{t,n} \in \{\mathbf{e}^{(i)}|\mathbf{e}^{(i)} \in \mathbb{R}^C\}, \forall t,n\\
\end{aligned}
\end{equation}
where $\hat{\mathbf{y}}$ indicates the set of target labels, $C$ is the number of classes, and $\mathbf{e}^{(i)}$ indicates a one-hot vector whose $i$-th entry is 1. By minimizing the loss in Eq. (\ref{uda}) with respect to $\hat{\mathbf{y}}$, the optimized $\hat{\mathbf{y}}$ should approximate the underlying true target ground truth. Domain adaptation can then be performed similarly to Eq. (\ref{finetune}). We call $\hat{\mathbf{y}}$ ``pseudo-labels'', and regard such training strategy as self-training.

\section{Proposed methods}\label{sect_st}
\subsection{Self-training (ST) with self-paced learning}
Jointly learning the model and optimizing pseudo-labels on unlabeled data is naturally difficult as it is not possible to completely guarantee the correctness of the generated pseudo-labels. A better strategy is to follow an ``easy-to-hard'' scheme via self-paced curriculum learning, where one seeks to generate pseudo-labels from the most confident predictions and hope they are mostly correct. Once the model is updated and better adapted to the target domain, the scheme then explores the remaining pseudo-labels with less confidence. To incorporate curriculum learning, we consider the following revised self-training formulation:
\begin{equation}\label{st}
\begin{aligned}
& \min_{\mathbf{w},\hat{\mathbf{y}}}\mathcal{L}_{ST}(\mathbf{w},\hat{\mathbf{y}}) = -\sum_{s=1}^{S}\sum_{n=1}^{N}\mathbf{y}_{s,n}^\top\log (\mathbf{p}_n(\mathbf{w},\mathbf{I}_s)) \\
&~~~~~~~~~~~~~~~~~~~~~~~ - \sum_{t=1}^{T}\sum_{n=1}^{N}\big[\hat{\mathbf{y}}_{t,n}^\top\log (\mathbf{p}_n(\mathbf{w},\mathbf{I}_t))+k|\hat{\mathbf{y}}_{t,n}|_1\big]\\
& ~s.t. ~ \hat{\mathbf{y}}_{t,n} \in \{\{\mathbf{e}^{(i)}|\mathbf{e}^{(i)} \in \mathbb{R}^C\} \cup \mathbf{0}\}, \forall t,n\\
&~~~~~~k > 0\\
\end{aligned}
\end{equation}
where assigning $\mathbf{y}_{s,n}$ as $\mathbf{0}$ leads to ignoring this pseudo-label in model training, and the $L_1$ regularization serves as a negative sparse promoting term to prevent the trivial solution of ignoring all pseudo-labels. $k$ is a hyperparameter controlling the amount of ignored pseudo-labels. A larger $k$ encourages the selection of more pseudo-labels for model training. To minimize the loss in Eq. (\ref{st}), we take the following alternative block coordinate descent algorithm:
\begin{itemize}
\item a) Fix (initialize) $\mathbf{w}$ and minimize the loss in Eq. \ref{st} with respect to $\hat{\mathbf{y}}_{t,n}$.
\item b) Fix $\hat{\mathbf{y}}_{t,n}$ and optimize the objective in Eq. \ref{st} with respect to $\mathbf{w}$.
\end{itemize}
We call one step of a) followed by one step of b) as one \textbf{round}. In this work, we propose a self-training algorithm where step a) and step b) are alternately repeated for multiple rounds. Intuitively, step a) selects a certain portion of most confident pseudo-labels from the target domain, while step b) trains the network model given the pseudo-labels selected in step a). Fig.~\ref{flow} illustrates the proposed algorithm flow in the domain adaptation example of GTA5 $\rightarrow$ Cityscapes.

Solving step b) leads to network learning with stochastic gradient descent. However, solving step a) requires a nonlinear integer programming given the optimization over discrete variables. Given $k>0$, step a) can be rewritten as:
\begin{equation}\label{st_a}
\begin{split}
\min_{\hat{\mathbf{y}}}& -\sum_{t=1}^{T}\sum_{n=1}^{N}\Big[\sum_{c=1}^C\hat{y}_{t,n}^{(c)}\log (p_n(c|\mathbf{w},\mathbf{I}_t))+k|\hat{\mathbf{y}}_{t,n}|_1\Big]\\
\mathrm{s.t.}& ~\hat{\mathbf{y}}_{t,n}=\big[\hat{y}_{t,n}^{(1)},...,\hat{y}_{t,n}^{(C)}\big] \in \{\{\mathbf{e}^{(i)}|\mathbf{e}^{(i)} \in \mathbb{R}^C\} \cup \mathbf{0}\}, \forall~t,n\\
&~k>0
\end{split}
\end{equation}
Since $\hat{\mathbf{y}}_{t,n}$ is required to be either a discrete one-hot vector or a zero vector, the pseudo-label configuration can be optimized via the following solver:
\begin{equation}\label{st_solver}
\hat{y}_{t,n}^{(c)*}=\left\{
\begin{aligned}
1, &~\mathbf{if}~c=\argmax_{c}p_n(c|\mathbf{w},\mathbf{I}_t),\\ &~~~~p_n(c|\mathbf{w},\mathbf{I}_t)>\exp(-k)\\
0, &~\mathrm{otherwise}\\
\end{aligned}
\right.
\end{equation}
Unlike traditional self-training adaptation with handcrafted features that learn a domain-invariant classifier, CNN based self-training can learn not only domain-invariant classifier but also domain-invariant features. The softmax loss implicitly tries to reduce the domain difference in feature space. In addition, the self-training also has the missing value (pseudo-label) problem, similar to EM algorithm. The proposed alternate optimization method can learn the weights of models without prior observation of target domain labels.

One may note that the proposed framework is similar to~\cite{tang2012shifting} and several other related works. However, the proposed method presents a more unified model for self-training and self-paced learning, in the sense that pseudo-label generation is unified with curriculum learning under a single learning framework. More importantly, in terms of the specific application, the above self-training framework sheds light on a relatively new direction for adapting semantic segmentation models. We will show that self-training based methods lead to considerably better or competitive performance compared to many current state of the art methods that are predominantly based on adversarial training.

\subsection{Class-balanced self-training (CBST)}
As mentioned in section \ref{introduction}, the difference in visual domain gap and class distribution can cause different levels of domain-transfer difficulties among different classes, which on average results in relatively higher prediction confidence scores for easy-to-transfer classes in the target domain. A problem with vanilla self-training is that it does not take such issue into consideration, but selects pseudo-labels by referring to their confidence universally across different classes. A consequent issue is that the model tends to bias towards some initially well-transferred classes while ignoring other hard classes along the training process. It is thus difficult for ST to perform well in certain domain adaptation scenarios. To overcome this issue, we propose a framework towards class-balanced self-training, where class-wise confidence levels are normalized to cancel the influence from diverse confidence levels:
\begin{equation}\label{cbst}
\begin{aligned}
&\min_{\mathbf{w},\hat{\mathbf{y}}}\mathcal{L}_{CB}(\mathbf{w},\hat{\mathbf{y}}) = -\sum_{s=1}^{S}\sum_{n=1}^{N}\mathbf{y}_{s,n}^\top\log (\mathbf{p}_n(\mathbf{w},\mathbf{I}_s))\\
&~~~~~~~~~~~~~~~~~~~~~~~-\sum_{t=1}^T\sum_{n=1}^N\sum_{c=1}^C \big[\hat{y}_{t,n}^{(c)}\log(p_n(c|\mathbf{w},\mathbf{I}_t))+k_c\hat{y}_{t,n}^{(c)}\big]\\
&~s.t. ~ \hat{\mathbf{y}}_{t,n}=\big[\hat{y}_{t,n}^{(1)},...,\hat{y}_{t,n}^{(C)}\big] \in \{\{\mathbf{e}^{(i)}|\mathbf{e}^{(i)} \in \mathbb{R}^C\} \cup \mathbf{0}\}, \forall t,n\\
&~~~~~~k_c > 0, \forall c
\end{aligned}
\end{equation}
where each $k_c$ is a separate parameter determining the proportion of selected pseudo-labels in class $c$. As one may observe, it is the difference between $k_c$ that introduces different levels of class-wise bias for pseudo-label selection, and addresses the issue of inter-class balance.

The optimization flow of class-balanced self-training is the same as in Eq. (\ref{st}) except for the generation of pseudo-labels. Again, we can rewrite the step of pseudo-label optimization as:
\begin{equation}\label{cbst_a}
\begin{split}
\min_{\hat{\mathbf{y}}}& -\sum_{t=1}^T\sum_{n=1}^N\sum_{c=1}^C\big[\hat{y}_{t,n}^{(c)}\log(p_n(c|\mathbf{w},\mathbf{I}_t))+k_c\hat{y}_{t,n}^{(c)}\big]\\
s.t.&~\hat{\mathbf{y}}_{t,n}=\big[\hat{y}_{t,n}^{(1)},...,\hat{y}_{t,n}^{(C)}\big] \in \{\{\mathbf{e}|\mathbf{e} \in \mathbb{R}^C\} \cup \mathbf{0}\}, \forall~t,n\\
&~k_c > 0, \forall~c\\
\end{split}
\end{equation}
Note that the loss function in Eq. (\ref{cbst_a}) can not be trivially minimized by the solver of Eq. (\ref{st}). Instead, optimizing Eq. (\ref{cbst_a}) leads to the following solver:
\begin{equation}\label{cbst_solver}
\hat{y}_{t,n}^{(c)*}=\left\{
\begin{aligned}
1, &~\mathbf{if}~c=\argmax_{c}\frac{p_n(c|\mathbf{w},\mathbf{I}_t)}{\exp(-k_c)},\\ &~~~~\frac{p_n(c|\mathbf{w},\mathbf{I}_t)}{\exp(-k_c)}>1\\
0, &~\mathrm{otherwise}\\
\end{aligned}
\right.
\end{equation}
From Eq. (\ref{cbst_solver}), one can see that pseudo-label generation in Eq. (\ref{cbst}) is no longer dependent on the output $p_n(c|\mathbf{w},\mathbf{I}_t)$, but hinges on the normalized output $\frac{p_n(c|\mathbf{w},\mathbf{I}_t)}{\exp(-k_c)}$. Pseudo-label assignment using this normalized output owns the benefit of balancing towards the class with relatively low score but having high within-class confidence. As a result, $k_c$ should be set in a way that $\exp(-k_c)$ encodes the response strength of each class to balance different classes. In addition, for CBST, the pseudo-label of any pixel is only filtered when all the balanced responses are smaller than 1. There could also be multiple classes with $\frac{p_n(c|\mathbf{w},\mathbf{I}_t)}{\exp(-k_c)}>1$. In this case, the class with the maximum balanced response is selected.

\subsection{Self-paced learning policy design}
\subsubsection{Determination of $k$ in ST} \label{plg_st_1}
From the previous section, we know that $k$ essentially plays a key role in pseudo-label selection, by filtering out those with confidence lower than $\exp(-k)$. We can design the following policy on $k$ such that we gradually increase the proportion of selected pseudo-labels in each round:

We take the confidence of all pixels from the whole target set, and sort their confidence in a descending order. We then set the value of $k$ such that $\exp(-k)$ equals to the probability ranked at $\mathrm{round}(p*T*N)$, where $p$ indicates the pseudo-label proportion and is between $[0, 1]$. In this case, pseudo-label optimization produces $p\times100\%$ most confident pseudo-labels for network training. The above pseudo-label selection policy is summarized in Algorithm \ref{k}.

\begin{algorithm}
    \SetKwInOut{Input}{Input}
    \SetKwInOut{Output}{Output}

    \Input{Neural network $P(\textbf{w})$, target images $\textbf{I}_t$, pseudo-label portion $p$}
    \Output{k}
    \For{t=1 to T}
     { P$_{\textbf{I}_t}$ = P($\textbf{w}$,$\textbf{I}_t$) \\
        MP$_{\textbf{I}_t}$ = max(P$_{\textbf{I}_t}$,axis=0) \\
     M = [M, matrix$\_$to$\_$vector(MP$_{\textbf{I}_t}$)]
     }
     {
     M = sort(M,order=descending)\\
     len$_{th}$ = length(M) $\times$ p \\
     k = -log(M[len$_{th}$]) \\
     return k
     }
    \caption{Determination of k in ST}
    \label{k}
\end{algorithm}

We design the self-paced learning policy such that more pseudo-labels are incorporated for each additional round. In particular, we start $p$ from $20\%$, and empirically add $5\%$ to $p$ in each additional round of pseudo-label generation. The maximum portion is set to be $50\%$.

\subsubsection{Determination of $k_c$ in CBST} \label{plg_cbst_1}
The policy of $k_c$ in CBST can be defined similar to ST. Although CBST seemingly introduce much more parameters than ST, we propose a strategy to determine $k_c$ for every class with a single parameter $p$, while effectively encoding the class-wise confidence levels. The proposed strategy is described in Algorithm~\ref{kc}.

Note that Algorithm~\ref{kc} determines $k_c$ by ranking the class $c$ probabilities on all pixels predicted as class $c$, and setting $k_c$ such that $\exp(-k_c)$ equals to the probability ranked at $\mathrm{round}(p*N_c)$, where $N_c$ indicates the number of pixels predicted as class $c$. Such a strategy basically takes the probability ranked at $p\times 100\%$ separately from each class as a reference for both thresholding and confidence normalization. The pseudo-label proportion $p$ and its increasing policy is defined exactly the same to ST.

\begin{algorithm}
	\SetKwInOut{Input}{Input}
	\SetKwInOut{Output}{Output}
	
	\Input{Neural network $f(\textbf{w})$, target images $\textbf{I}_t$, pseudo-label portion $p$}
	\Output{$\textbf{k}_c$}
	\For{t=1 to T}
	{ P$_{\textbf{I}_t}$ = P($\textbf{w}$,$\textbf{I}_t$) \\
		LP$_{\textbf{I}_t}$ = argmax(P,axis=0) \\
		MP$_{\textbf{I}_t}$ = max(P,axis=0) \\
		\For{c=1 to C}
		{
			MP$_{c,\textbf{I}_t}$ = MP$_{\textbf{I}_t}$(LP$_{\textbf{I}_t}$ == c) \\
			M$_c$ = [M$_c$, matrix$\_$to$\_$vector(MP$_{c,\textbf{I}_t}$)]
		}
	}
	\For{c=1 to C}
	{
		M$_c$ = sort(M$_c$,order=descending)\\
		len$_{c,th}$ = length(M$_c$) $\times$ p \\
		k$_c$ = -log(M$_c$[len$_{c,th}$]) \\
	}
	{
		return $\textbf{k}_c$
	}
	\caption{Determination of $k_c$ in CBST}
	\label{kc}
\end{algorithm}

\begin{figure*}[!b]
\centering
\includegraphics[width=\textwidth]{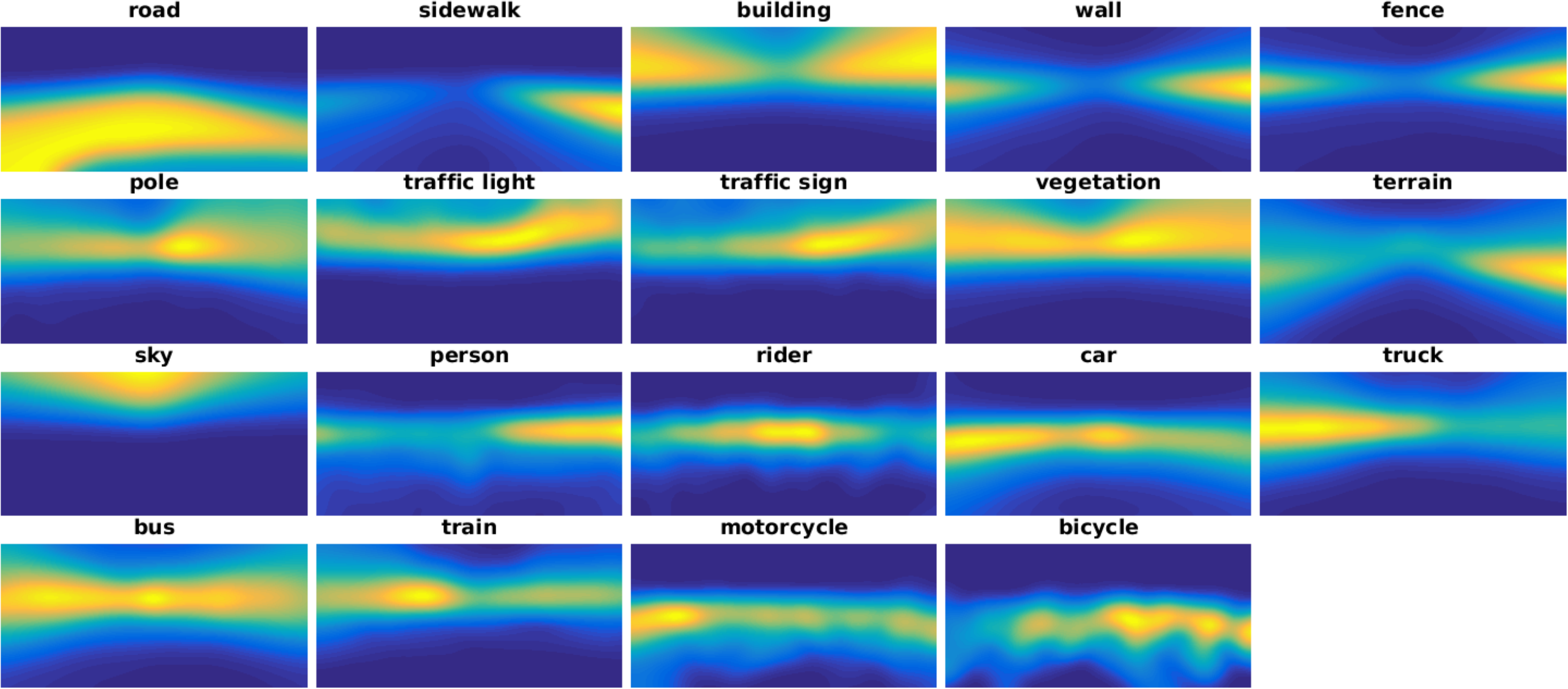}
\caption{Visualization of class-wise spatial priors on GTA5.}
\label{sp}
\end{figure*}

\subsection{Incorporating spatial priors}
For adapting models in the case of street scenes, we could take advantage of the spatial prior knowledge. Traffic scenes have common structures. For example, sky is not likely to appear at the bottom and road is not likely to appear at the top. If the image views in source domain and target domain are similar, we believe this knowledge can help to adapt source model. Thus we introduce spatial priors, similar to \cite{silberman2011indoor}, by counting the class frequencies in the source domain, followed by smoothing with a $70 \times 70$ Gaussian kernel. In particular, we use $q_n(c)$ to indicate the frequency of class $c$ at pixel $n$. Upon obtaining the class frequencies, we also normalize them by requiring $\sum_{i=1}^{N}q_n(c) = 1$. Fig.~\ref{sp} shows the heat map of spatial priors, calculated from GTA5 dataset, where yellow color indicates higher energy and blue color indicates lower energy.

To incorporate spatial priors into proposed CBST, we multiply the softmax output with the spatial priors, and consider the resulting potential as selection metric in pseudo-label generation:
\begin{equation}\label{cbst-sp}
\begin{aligned}
& \min_{\mathbf{w},\hat{\mathbf{y}}}\mathcal{L}_{SP}(\mathbf{w},\hat{\mathbf{y}}) = -\sum_{s=1}^{S}\sum_{n=1}^{N}\mathbf{y}_{s,n}^\top\log (\mathbf{p}_n(\mathbf{w},\mathbf{I}_s))\\
& ~~~~~~~~~~~~~~~~~~~~~~~-\sum_{t=1}^T\sum_{n=1}^N\sum_{c=1}^C\big[\hat{y}_{t,n}^{(c)}\log(q_n(c) p_n(c|\mathbf{w},\mathbf{I}_t))+k_c\hat{y}_{t,n}^{(c)}\big] \\
& ~s.t. ~ \hat{\mathbf{y}}_{t,n} \in \{\{\mathbf{e}|\mathbf{e} \in \mathbb{R}^C\} \cup \mathbf{0}\}, \forall t,n\\
&~~~~~~k_c > 0, \forall c
\end{aligned}
\end{equation}
We denote this as CBST-SP. The self-training and self-paced learning policy are identical to CBST, except that the potential $q_n(c)p_n(c|\mathbf{w},\mathbf{I}_t)$ is used to replace $p_n(c|\mathbf{w},\mathbf{I}_t)$ in CBST. It should be noted that incorporating the spatial prior does not change network training, since $q_n(c)$ can be taken out of $\log(\cdot)$ as a constant.

\section{Experiments}
In this section, we provide comprehensive evaluations of the proposed methods by performing experiments on three domain adaptation settings. We first consider a cross-city adaptation scenario by transferring segmentation models from Cityscapes training set to the NTHU Dataset~\cite{Chen_2017_ICCV}, where the dataset contains 400 images of size $1,024\times 2,048$ from 4 different cities: Rome, Rio, Tokyo and Taipei. We also consider another two challenging synthetic-to-real scenarios: SYNTHIA~\cite{ros2016synthia} to Cityscapes and GTA5 \cite{richter2016playing} to Cityscapes. We use the SYNTHIA-RAND-CITYSCAPES subset which includes 9,400 labeled images of size $760\times 1280$. The GTA5 Dataset includes 24,966 annotated images of size $1,052 \times 1,914$ rendered by the GTA5 game engine. In both of the above settings, the validation set of Cityscapes is treated as target domain.

We use FCN8s-VGG16~\cite{long2015fully} as one of the backbone networks in SYNTHIA to Cityscapes and GTA5 to Cityscapes to give a fair comparison with other methods using the same backbone. In addition, we boost our method performance with a better backbone network ResNet-38~\cite{wu2016wider}. Our implementations are based on MXNet~\cite{chen2015mxnet}, where we pretrain the networks on ImageNet~\cite{russakovsky2015imagenet} and fine-tune on source datasets with SGD. We also apply a hard sample mining strategy which mines the smallest classes according to target domain predictions. In particular, during random cropping on each target image for network input, priorities are given to classes whose predicted portions are less than $0.1\%$, by selectively cropping at locations containing pixels predicted to those classes.

\subsection{Small domain shift: Cross-city adaptation}
The NTHU dataset~\cite{Chen_2017_ICCV} contains 13 overlapping classes with Cityscapes. It should be noted that we train a 19 class model on Cityscapes and then evaluate it on the 13 NTHU classes. Similar to the rule of NTHU dataset, we consider pole, fence, wall as buildings, truck as car and terrain as vegetations. Following~\cite{Chen_2017_ICCV}, we split 100 images into 10 folds for each city, and report the cross-validation results by each time self-training with 90 unlabeled images and testing on the remaining 10. The results are shown in Table~\ref{t_citynthu}, which indicates that CBST achieves comparable or even better performance compared with state-of-the-art.

\begin{table*}[!t]
\centering
\caption{Experimental results for Cityscapes $\rightarrow$ NTHU dataset}\label{t_citynthu}
\vspace{1mm}
\resizebox{\linewidth}{!}{
\begin{tabular}{c|c|ccccccccccccc|c}
\hline
City                    & Method           & Road & SW   & Build & TL   & TS   & Veg. & Sky  & PR   & Rider & Car  & Bus  & Motor & Bike & Mean \\ \hline
\multirow{5}{*}{Rome}   & Source Dilation-Frontend \cite{Chen_2017_ICCV}     & 77.7 & 21.9 & 83.5  & 0.1  & 10.7 & 78.9 & 88.1 & 21.6 & 10.0  & 67.2 & 30.4 & 6.1   & 0.6  & 38.2 \\
                        & GCAA \cite{Chen_2017_ICCV}            & 79.5 & 29.3 & 84.5  & 0.0  & 22.2 & 80.6 & 82.8 & 29.5 & 13.0  & 71.7 & 37.5 & 25.9  & 1.0  & 42.9 \\ \cline{2-16}
& DeepLab-v2 \cite{tsai2018learning} & 83.9 & 34.3 & 87.7 & 13.0 & 41.9 & 84.6 & 92.5 & 37.7 & 22.4 & 80.8 & 38.1 & 39.1 & 5.3 & 50.9 \\
                        & MAA \cite{tsai2018learning}     & 83.9 & 34.2 & 88.3 & \textbf{18.8} & 40.2 & \textbf{86.2} & \textbf{93.1} & 47.8 & 21.7 & 80.9 & \textbf{47.8} & 48.3 & 8.6 & \textbf{53.8} \\  \cline{2-16}
                        & Source Resnet-38 & 86.0 & 21.4 & 81.5  & 14.3 & 47.4 & 82.9 & 59.8 & 30.8 & 20.9  & 83.1 & 20.2 & 40.0  & 5.6  & 45.7 \\
                        & ST               & 85.9 & 20.2 & 84.3  & 15.0 & 46.4 & 84.9 & 73.5 & \textbf{48.5} & 21.6  & 84.6 & 17.6 & 46.2  & 6.7  & 48.9 \\
                        & CBST             & \textbf{87.1} & \textbf{43.9} & \textbf{89.7} & 14.8 & \textbf{47.7} & 85.4 & 90.3 & 45.4 & \textbf{26.6} & \textbf{85.4} & 20.5 & \textbf{49.8} & \textbf{10.3} & 53.6 \\ \hline
\multirow{5}{*}{Rio}    & Source Dilation-Frontend \cite{Chen_2017_ICCV}     & 69.0 & 31.8 & 77.0  & 4.7  & 3.7  & 71.8 & 80.8 & 38.2 & 8.0   & 61.2 & 38.9 & 11.5  & 3.4  & 38.5 \\
                        & GCAA \cite{Chen_2017_ICCV}            & 74.2 & 43.9 & 79.0  & 2.4  & 7.5  & 77.8 & 69.5 & 39.3 & 10.3  & 67.9 & \textbf{41.2} & 27.9  & 10.9 & 42.5 \\ \cline{2-16}
                        & DeepLab-v2 \cite{tsai2018learning} & 76.6 & 47.3 & 82.5 & 12.6 & 22.5 & 77.9 & 86.5 & 43.0 & 19.8 & 74.5 & 36.8 & 29.4 & 16.7 & 48.2 \\
& MAA \cite{tsai2018learning}     & 76.2 & 44.7 & 84.6 & 9.3 & 25.5 & \textbf{81.8} & \textbf{87.3} & 55.3 & \textbf{32.7} & 74.3 & 28.9 & \textbf{43.0} & \textbf{27.6} & 51.6 \\  \cline{2-16}
                        & Source Resnet-38 & 80.6 & 36.0 & 81.8  & \textbf{21.0} & 33.1 & 79.0 & 64.7 & 36.0 & 21.0  & 73.1 & 33.6 & 22.5  & 7.8  & 45.4 \\
                        & ST               & 80.1 & 41.4 & 83.8  & 19.1 & \textbf{39.1} & 80.8 & 71.2 & \textbf{56.3} & 27.7  & \textbf{79.9} & 32.7 & 36.4  & 12.2 & 50.8 \\
                        & CBST             & \textbf{84.3} & \textbf{55.2} & \textbf{85.4} & 19.6 & 30.1 & 80.5 & 77.9 & 55.2 & 28.6 & 79.7 & 33.2 & 37.6 & 11.5 & \textbf{52.2} \\ \hline
\multirow{5}{*}{Tokyo}  & Source Dilation-Frontend \cite{Chen_2017_ICCV}     & 81.2 & 26.7 & 71.7  & 8.7  & 5.6  & 73.2 & 75.7 & 39.3 & 14.9  & 57.6 & 19.0 & 1.6   & 33.8 & 39.2 \\
                        & GCAA \cite{Chen_2017_ICCV}            & 83.4 & \textbf{35.4} & 72.8  & 12.3 & 12.7 & 77.4 & 64.3 & 42.7 & 21.5  & 64.1 & \textbf{20.8} & 8.9   & 40.3 & 42.8 \\ \cline{2-16}
                        & DeepLab-v2 \cite{tsai2018learning} & 83.4 & 35.4 & 72.8 & 12.3 & 12.7 & 77.4 & 64.3 & 42.7 & 21.5 & 64.1 & \textbf{20.8} & 8.9 & 40.3 & 42.8 \\
& MAA \cite{tsai2018learning}     &  81.5 & 26.0 & 77.8 & \textbf{17.8} & 26.8 & 82.7 & \textbf{90.9} & 55.8 & \textbf{38.0} & 72.1 & 4.2 & 24.5 & \textbf{50.8} & \textbf{49.9} \\  \cline{2-16}
                        & Source Resnet-38 & 83.8 & 26.4 & 73.0  & 6.5  & 27.0 & 80.5 & 46.6 & 35.6 & 22.8  & 71.3 & 4.2  & 10.5  & 36.1 & 40.3 \\
                        & ST               & 83.1 & 27.7 & 74.8  & 7.1  & 29.4 & \textbf{84.4} & 48.5 & \textbf{57.2} & 23.3  & \textbf{73.3} & 3.3  & 22.7  & 45.8 & 44.6 \\
                        & CBST             & \textbf{85.2} & 33.6 & \textbf{80.4} & 8.3 & \textbf{31.1} & 83.9 & 78.2 & 53.2 & 28.9 & 72.7 & 4.4 & \textbf{27.0} & 47.0 & 48.8 \\ \hline
\multirow{5}{*}{Taipei} & Source Dilation-Frontend \cite{Chen_2017_ICCV}     & 77.2 & 20.9 & 76.0  & 5.9  & 4.3  & 60.3 & 81.4 & 10.9 & 11.0  & 54.9 & 32.6 & 15.3  & 5.2  & 35.1 \\
                        & GCAA \cite{Chen_2017_ICCV}            & 78.6 & 28.6 & 80.0  & 13.1 & 7.6  & 68.2 & 82.1 & 16.8 & 9.4   & 60.4 & 34.0 & 26.5  & 9.9  & 39.6 \\ \cline{2-16}
                        & DeepLab-v2 \cite{tsai2018learning} & 78.6 & 28.6 & 80.0 & 13.1 & 7.6 & 68.2 & 82.1 & 16.8 & 9.4 & 60.4 & 34.0 & 26.5 & 9.9 & 39.6 \\
& MAA \cite{tsai2018learning}     &  81.7 & 29.5 & \textbf{85.2} & \textbf{26.4} & 15.6 & \textbf{76.7} & \textbf{91.7} & 31.0 & 12.5 & 71.5 & \textbf{41.1} & 47.3 & 27.7 & 49.1 \\  \cline{2-16}
                        & Source Resnet-38 & 84.9 & 26.0 & 80.1  & 8.3  & \textbf{28.0} & 73.9 & 54.4 & 18.9 & 26.8  & 71.6 & 26.0 & 48.2  & 14.7 & 43.2 \\
                        & ST               & 83.1 & 23.5 & 78.2  & 9.6  & 25.4 & 74.8 & 35.9 & \textbf{33.2} & 27.3 & 75.2 & 32.3 & 52.2  & 28.8 & 44.6 \\
                        & CBST             & \textbf{86.1} & \textbf{35.2} & 84.2 & 15.0 & 22.2 & 75.6 & 74.9 & 22.7 & \textbf{33.1} & \textbf{78.0} & 37.6 & \textbf{58.0} & \textbf{30.9} & \textbf{50.3} \\ \hline
\end{tabular}%
}
\end{table*}

\subsection{Large domain shift: Synthetic-to-real adaptation}
\subsubsection{SYNTHIA to Cityscapes}
We follow the same evaluation protocol as other works \cite{hoffman2016fcns,Zhang_2017_ICCV}, where we choose the 16 overlapping classes between SYNTHIA and Cityscapes as valid classes for evaluation. There is another setting which only considers 13 classes excluding wall, fence and pole~\cite{tsai2018learning}.

\begin{figure}[!t]
\centering
\resizebox{1.0\textwidth}{!}{
\begin{tabular}{@{}cccccccccc@{}}
\cellcolor{city_color_1}\textcolor{white}{~~road~~} &
\cellcolor{city_color_2}~~sidewalk~~&
\cellcolor{city_color_3}\textcolor{white}{~~building~~} &
\cellcolor{city_color_4}\textcolor{white}{~~wall~~} &
\cellcolor{city_color_5}~~fence~~ &
\cellcolor{city_color_6}~~pole~~ &
\cellcolor{city_color_7}~~traffic lgt~~ &
\cellcolor{city_color_8}~~traffic sgn~~ &
\cellcolor{city_color_9}~~vegetation~~ \\
\cellcolor{city_color_10}~~terrain~~ &
\cellcolor{city_color_11}~~sky~~ &
\cellcolor{city_color_12}\textcolor{white}{~~person~~} &
\cellcolor{city_color_13}\textcolor{white}{~~rider~~} &
\cellcolor{city_color_14}\textcolor{white}{~~car~~} &
\cellcolor{city_color_15}\textcolor{white}{~~truck~~} &
\cellcolor{city_color_16}\textcolor{white}{~~bus~~} &
\cellcolor{city_color_17}\textcolor{white}{~~train~~} &
\cellcolor{city_color_18}\textcolor{white}{~~motorcycle~~} &
\cellcolor{city_color_19}\textcolor{white}{~~bike~~}
\end{tabular}
}
\vspace{0.2mm}

\includegraphics[width=0.192\textwidth]{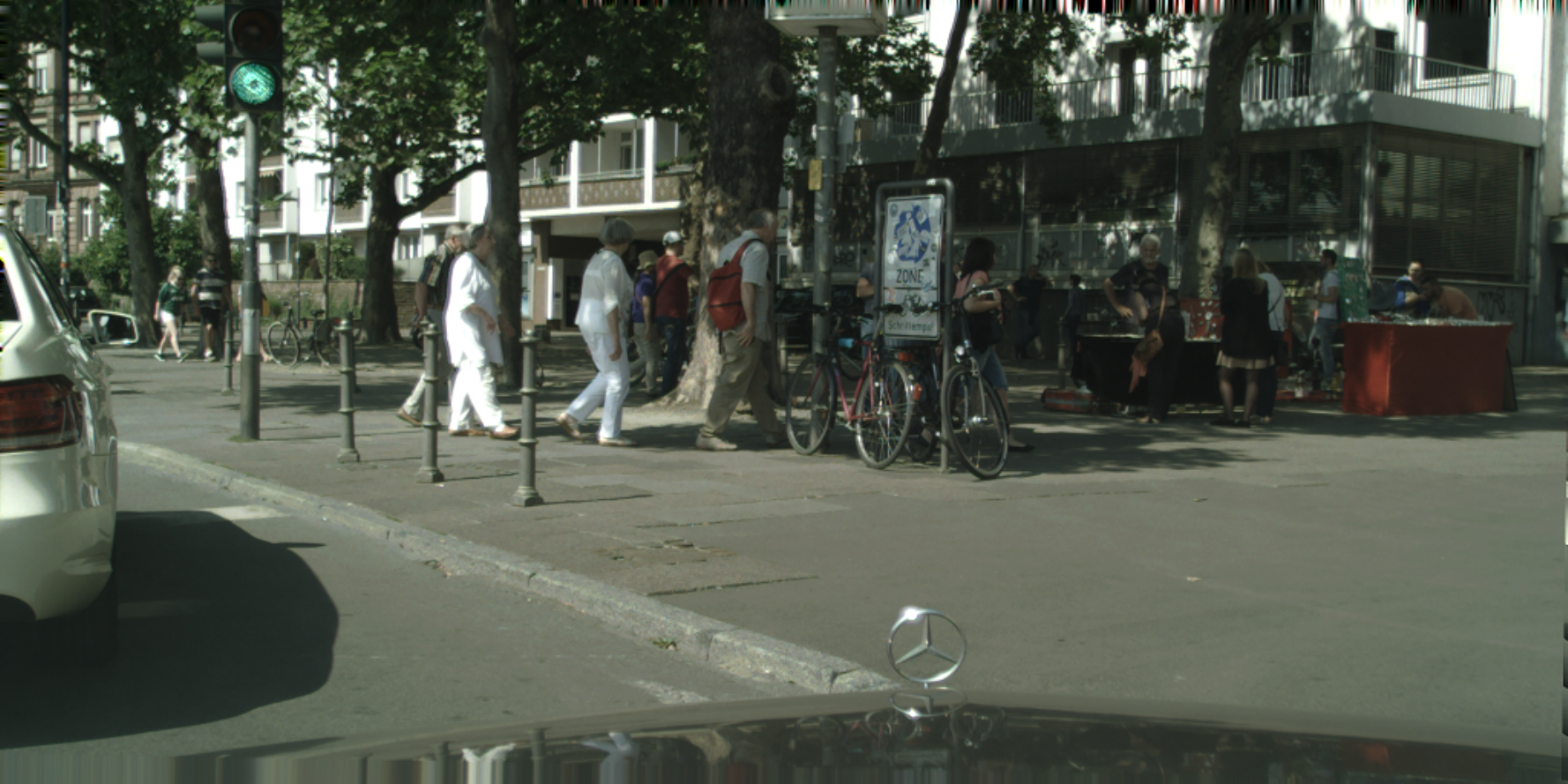}
\includegraphics[width=0.192\textwidth]{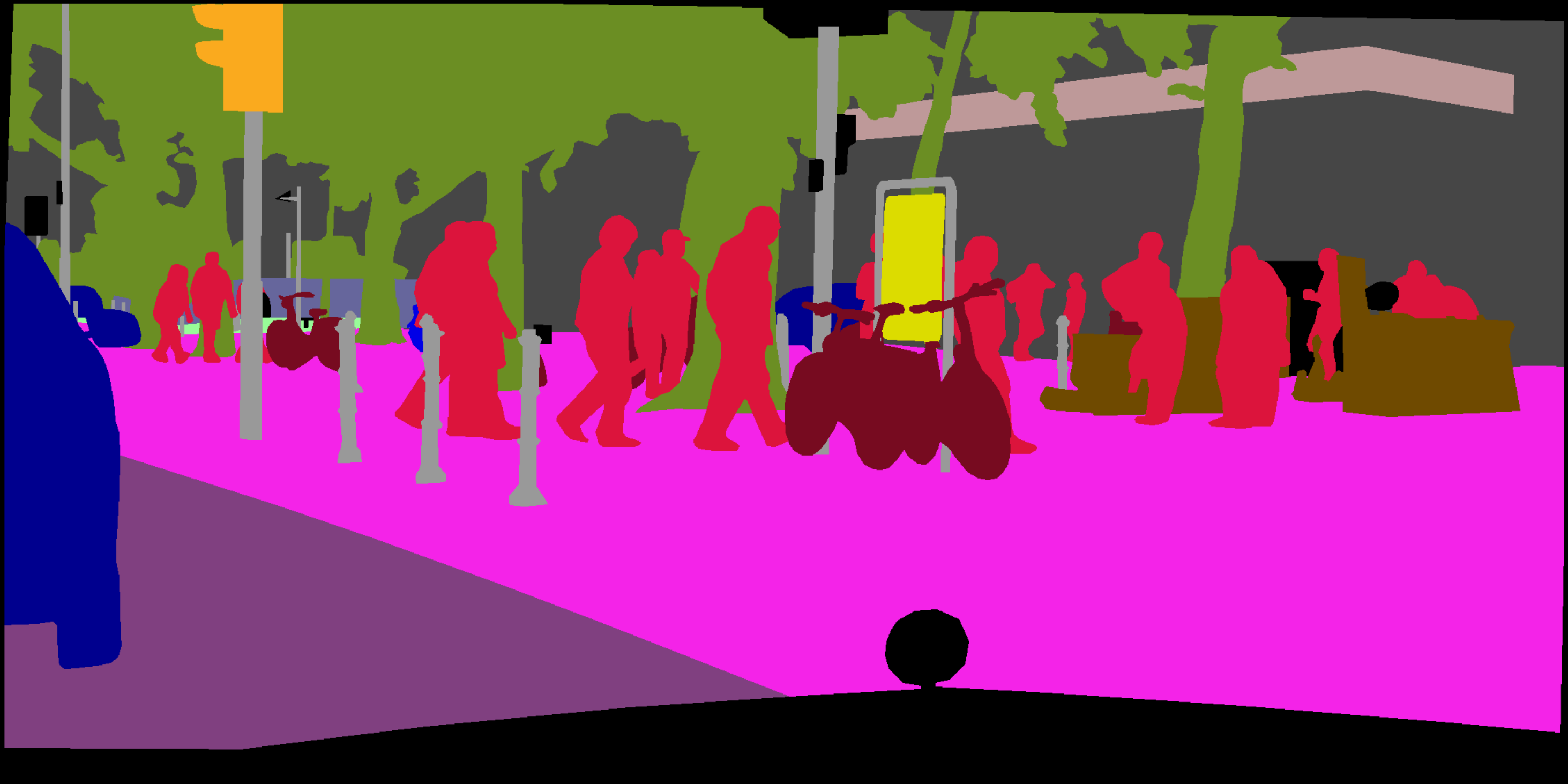}
\includegraphics[width=0.192\textwidth]{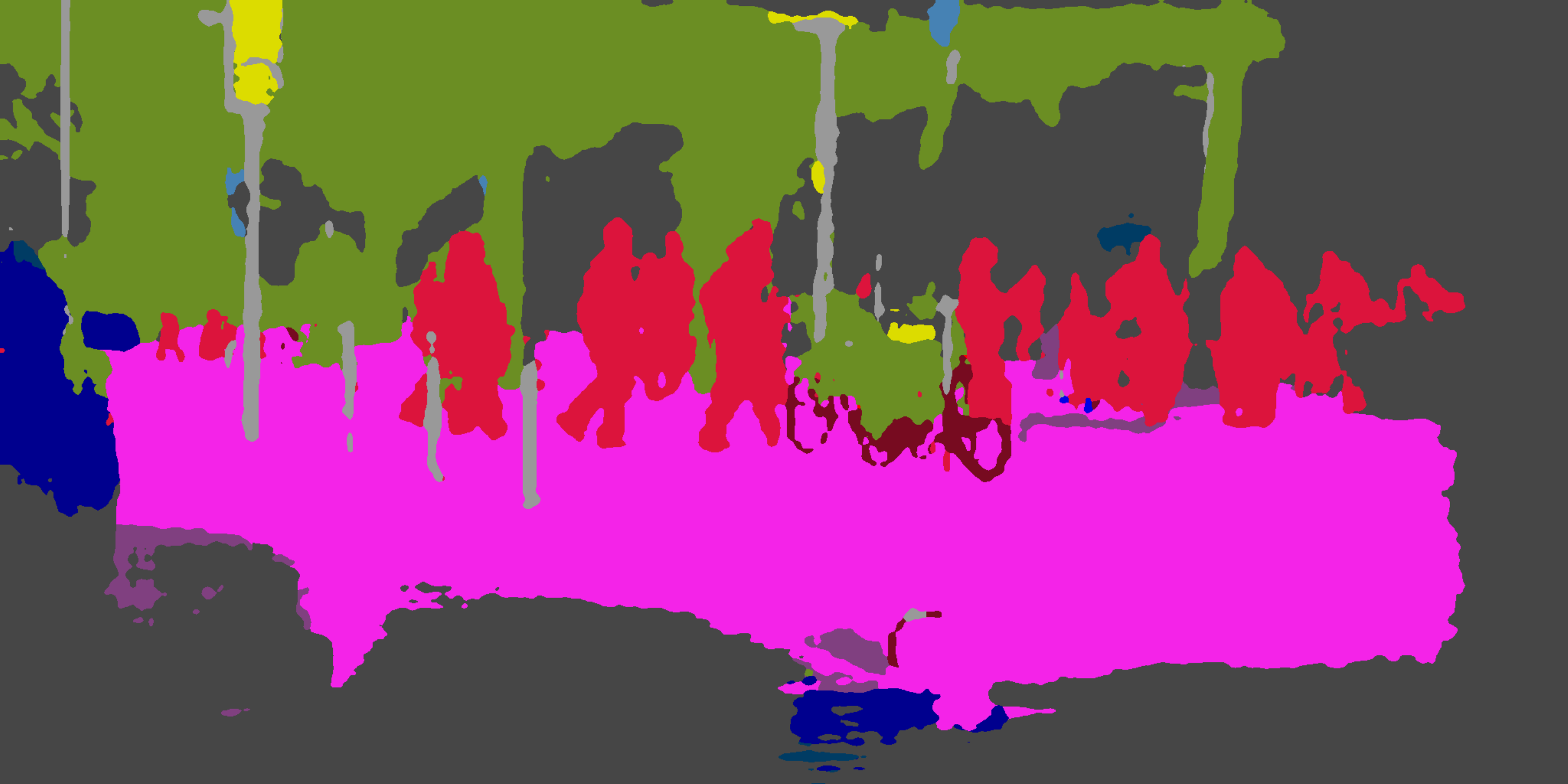}
\includegraphics[width=0.192\textwidth]{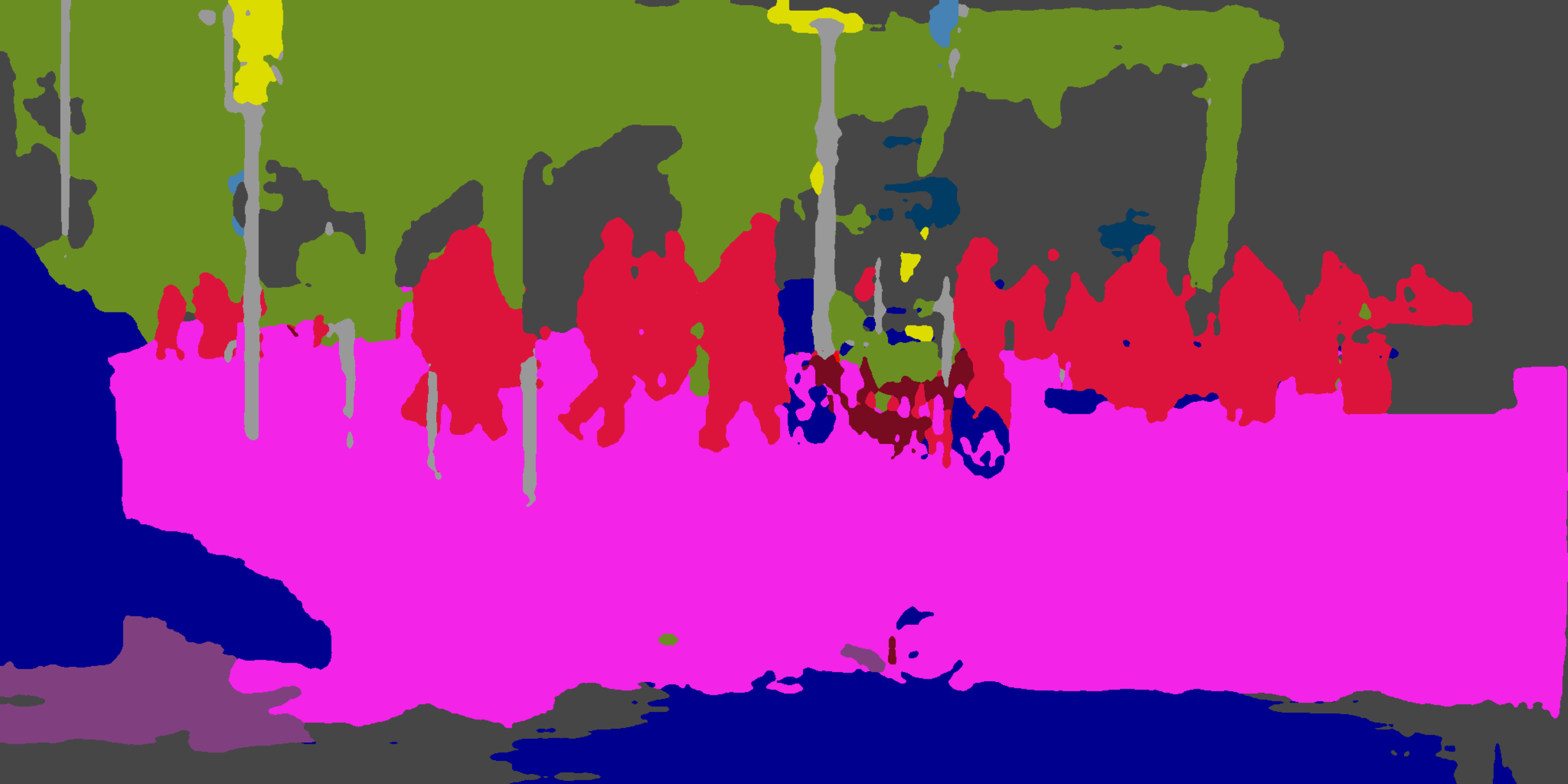}
\includegraphics[width=0.192\textwidth]{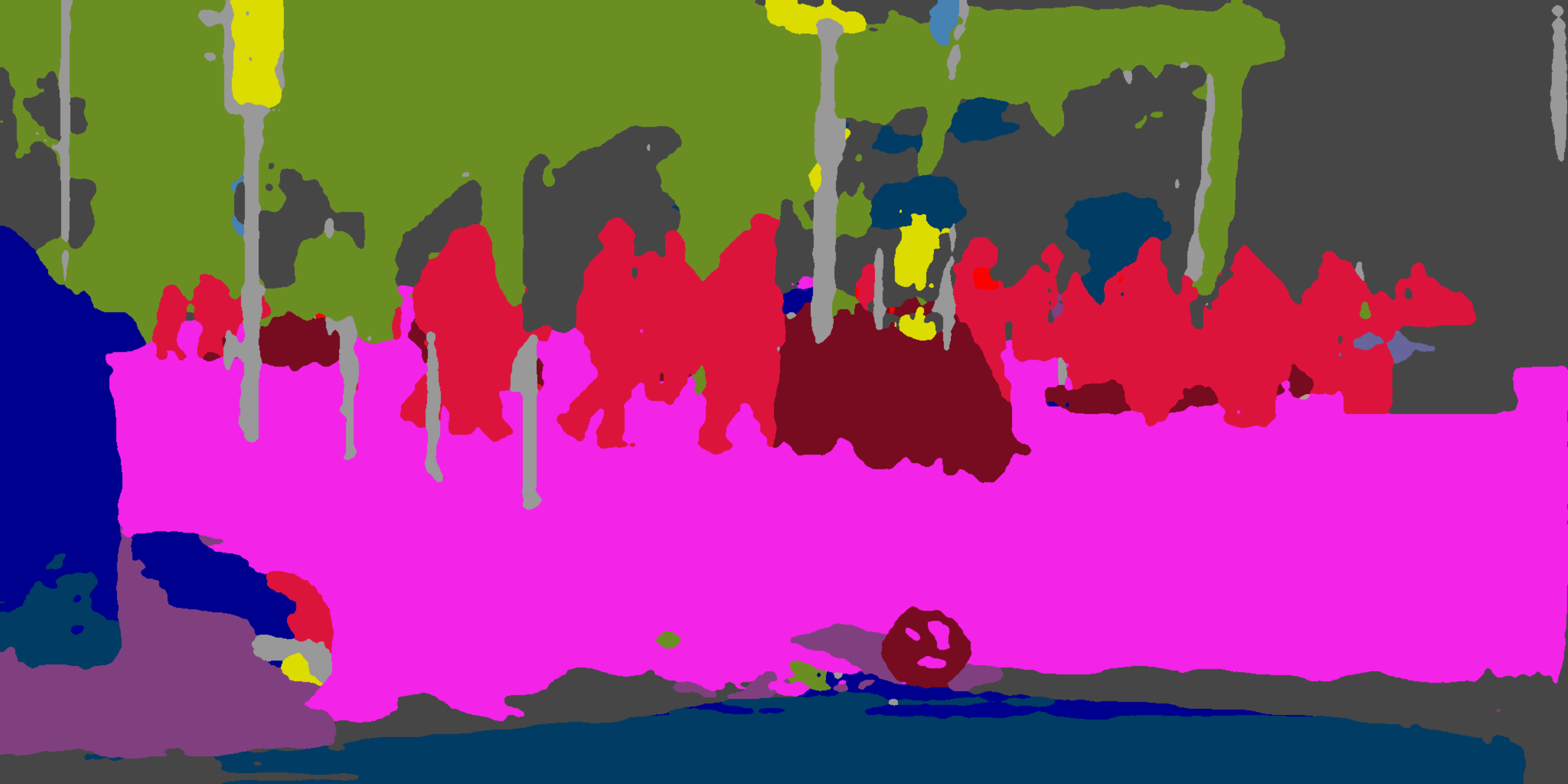}\\
\quad\\\vspace{-0.325cm}
\includegraphics[width=0.192\textwidth]{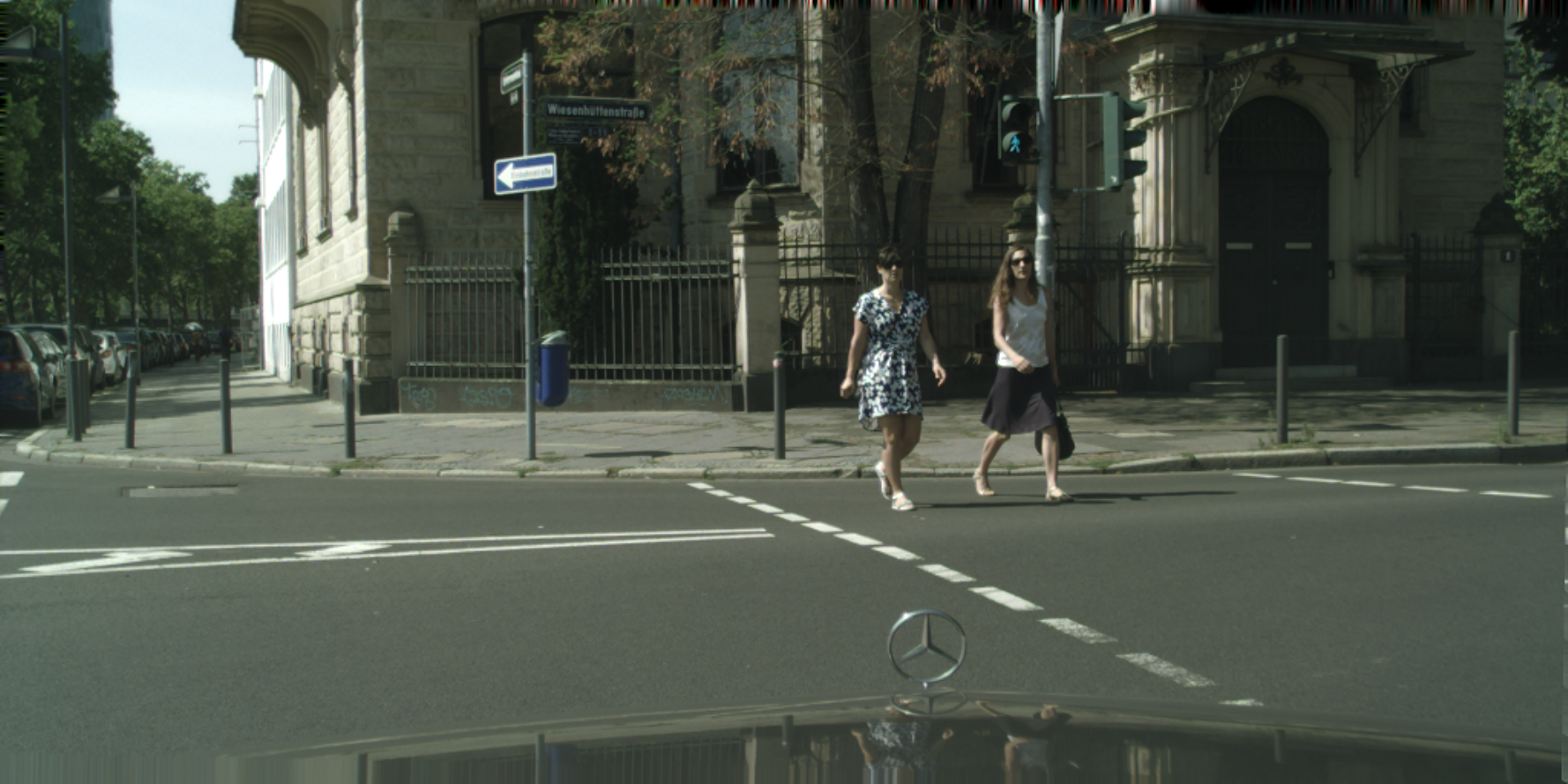}
\includegraphics[width=0.192\textwidth]{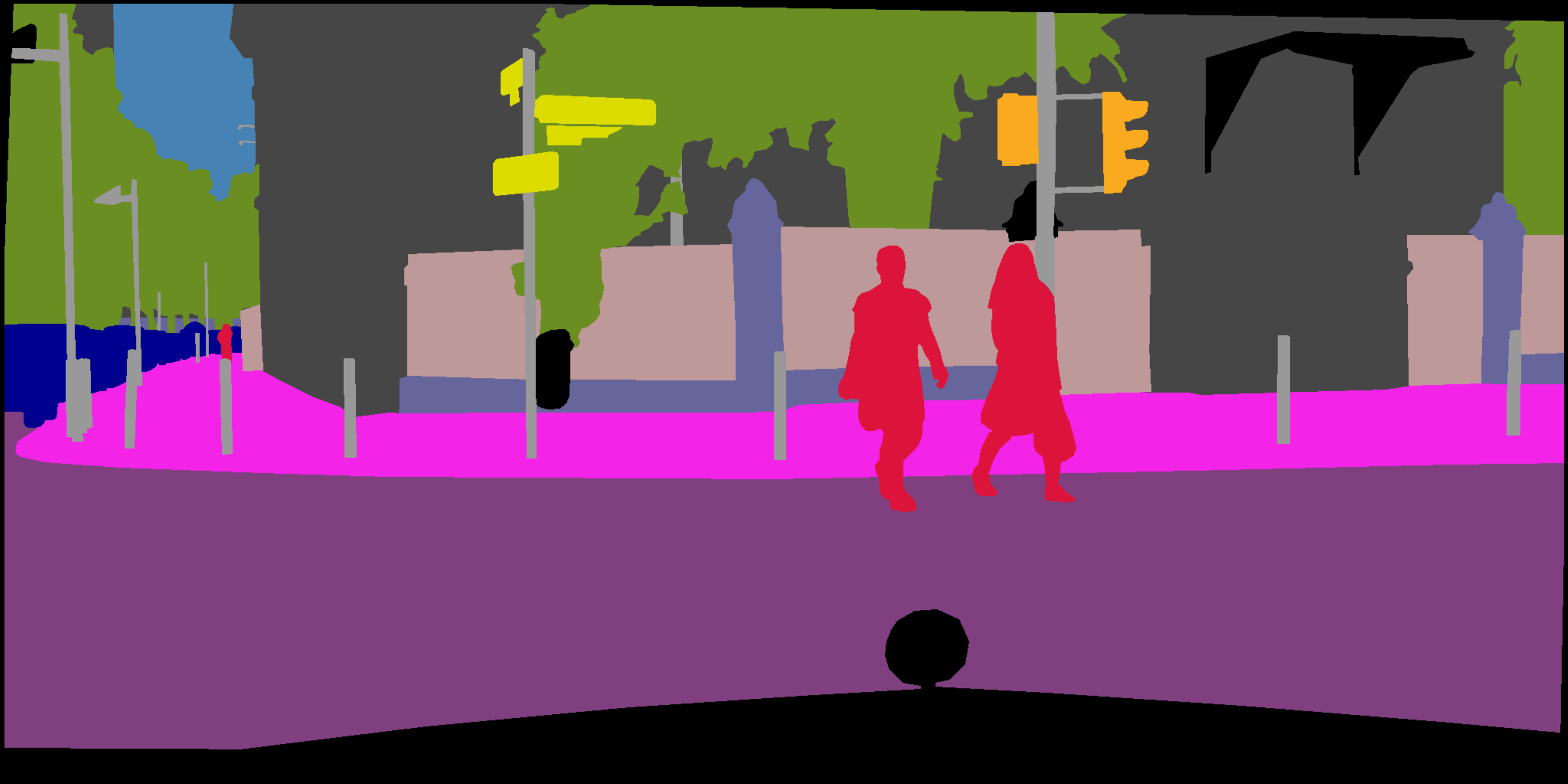}
\includegraphics[width=0.192\textwidth]{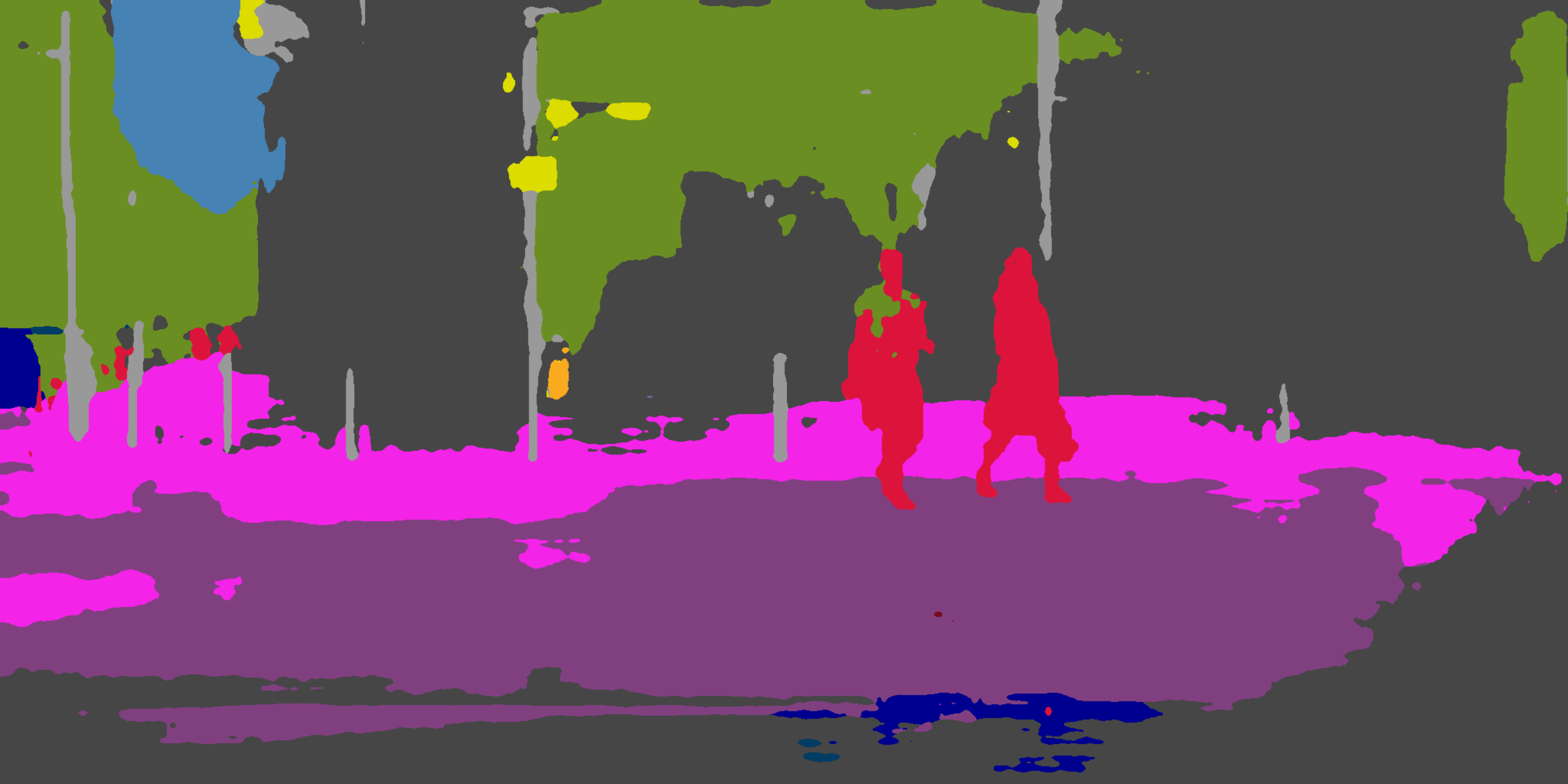}
\includegraphics[width=0.192\textwidth]{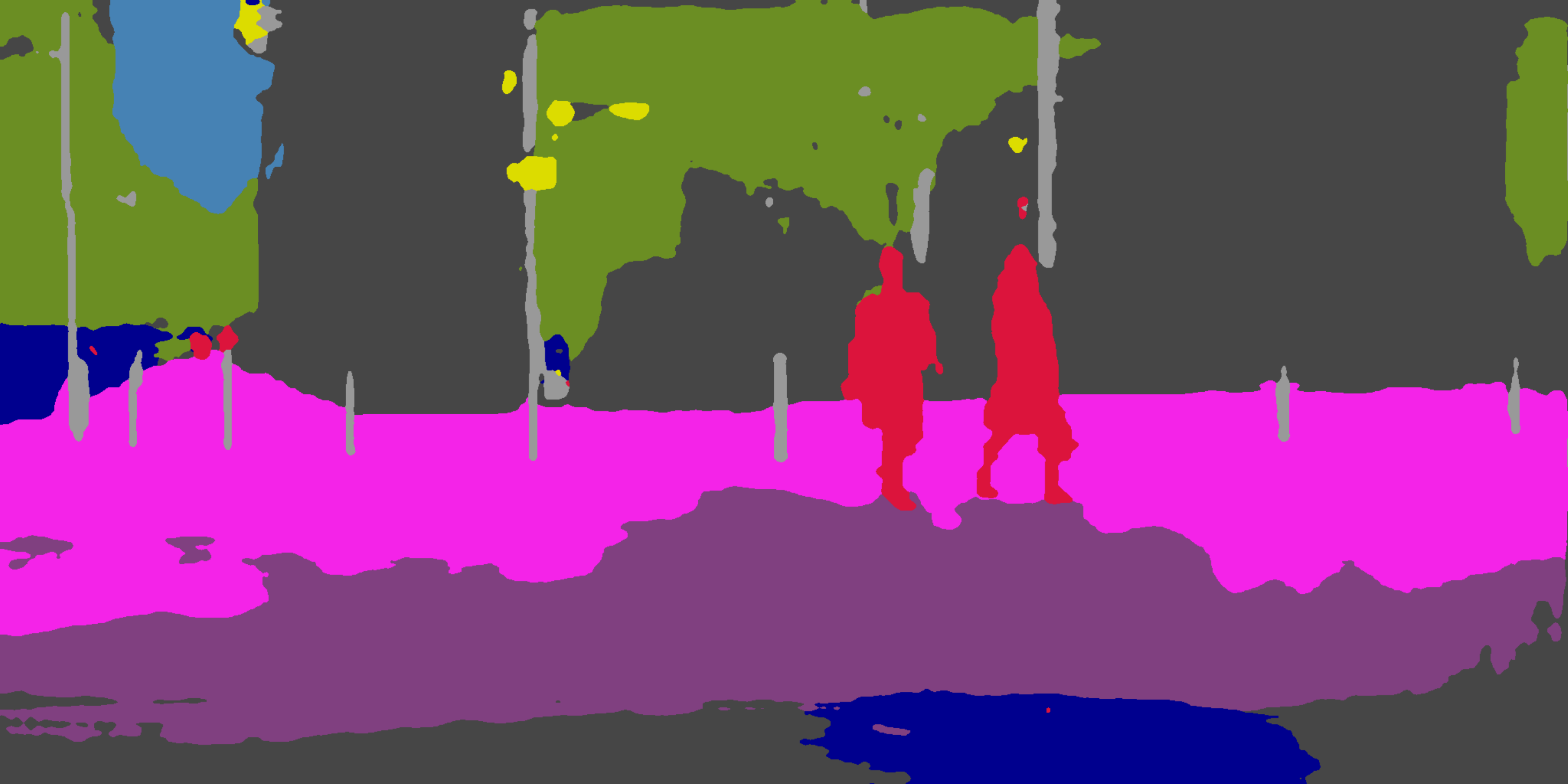}
\includegraphics[width=0.192\textwidth]{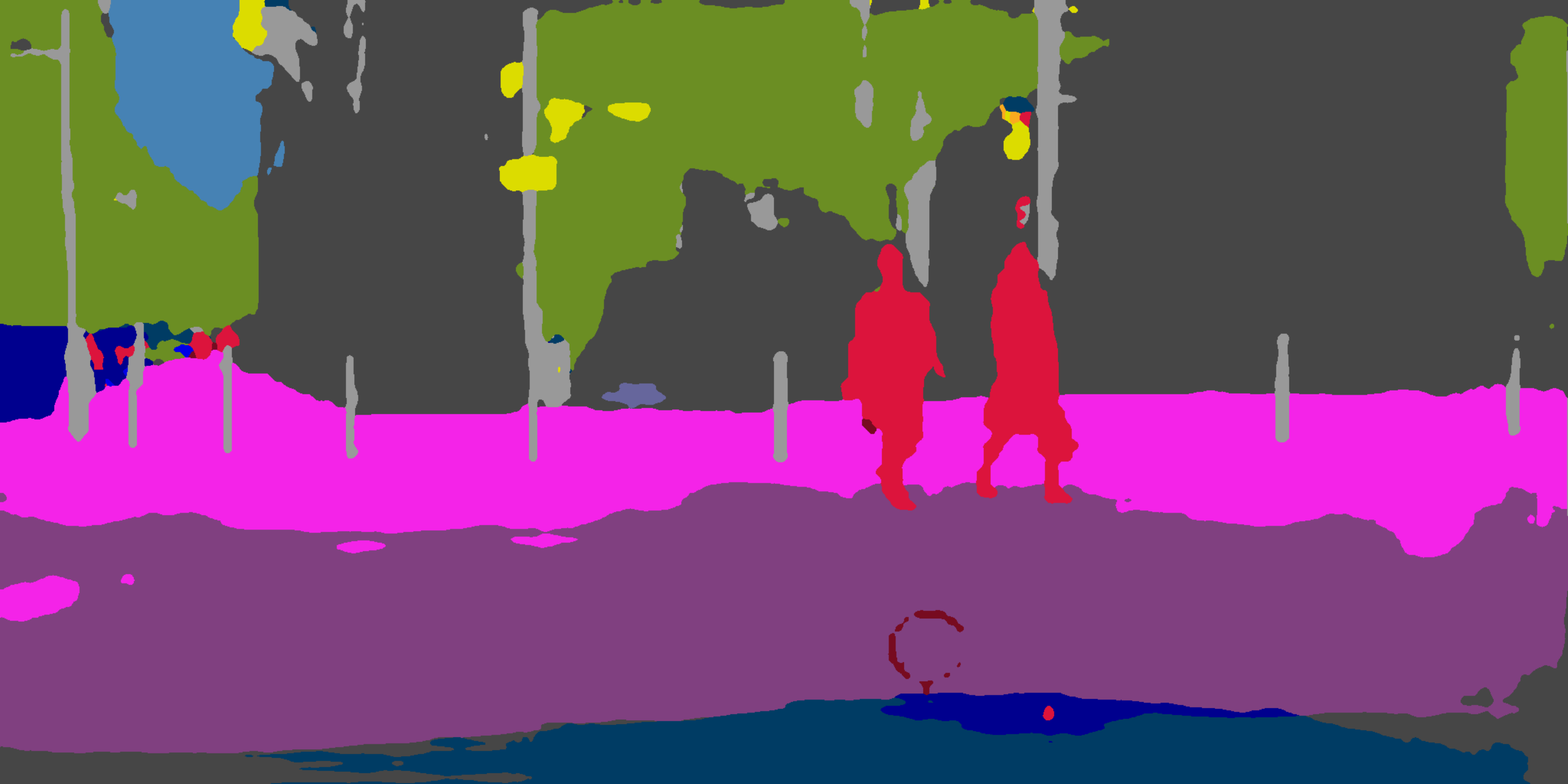}\\
\quad\\\vspace{-0.325cm}
\includegraphics[width=0.192\textwidth]{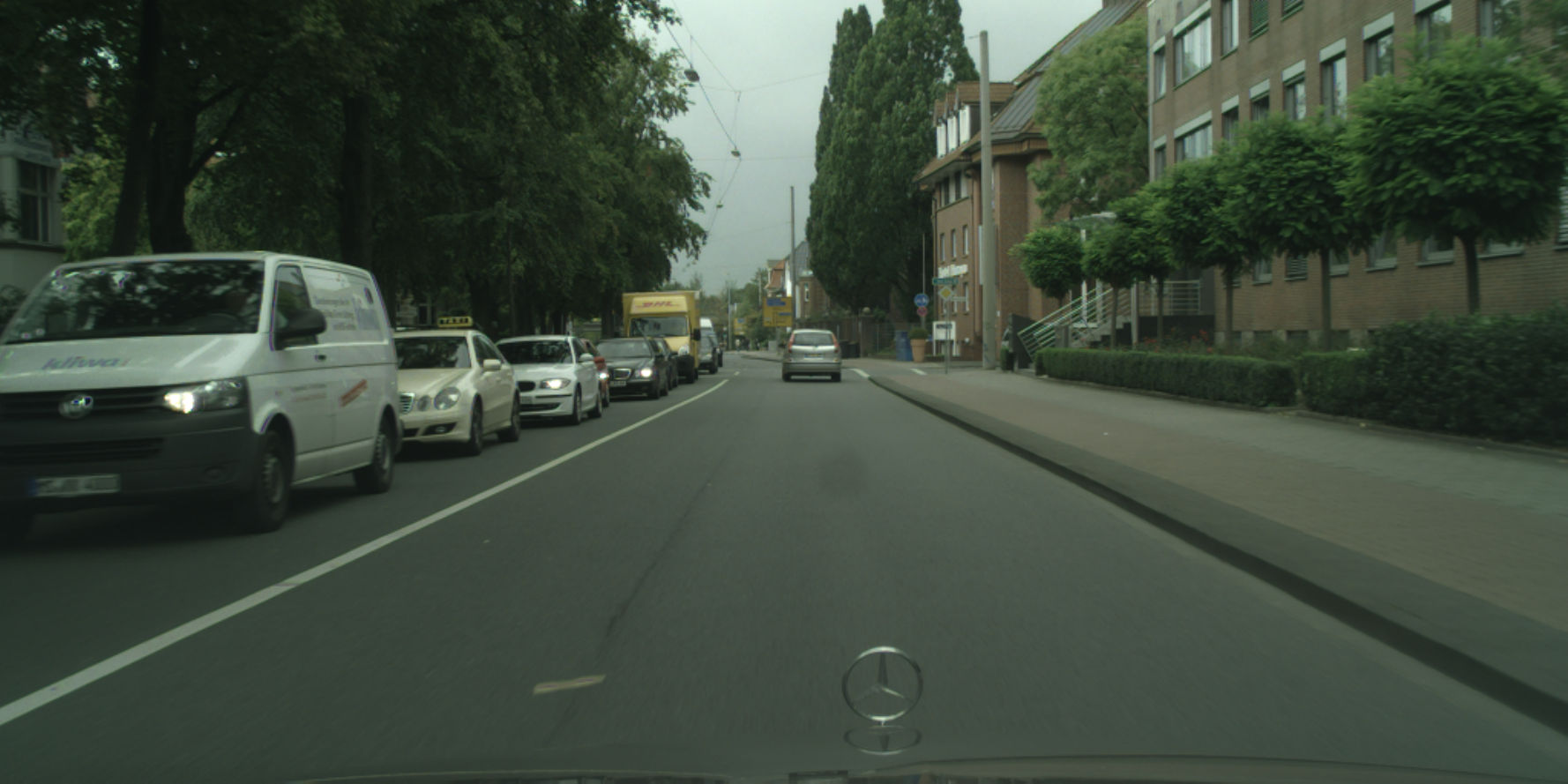}
\includegraphics[width=0.192\textwidth]{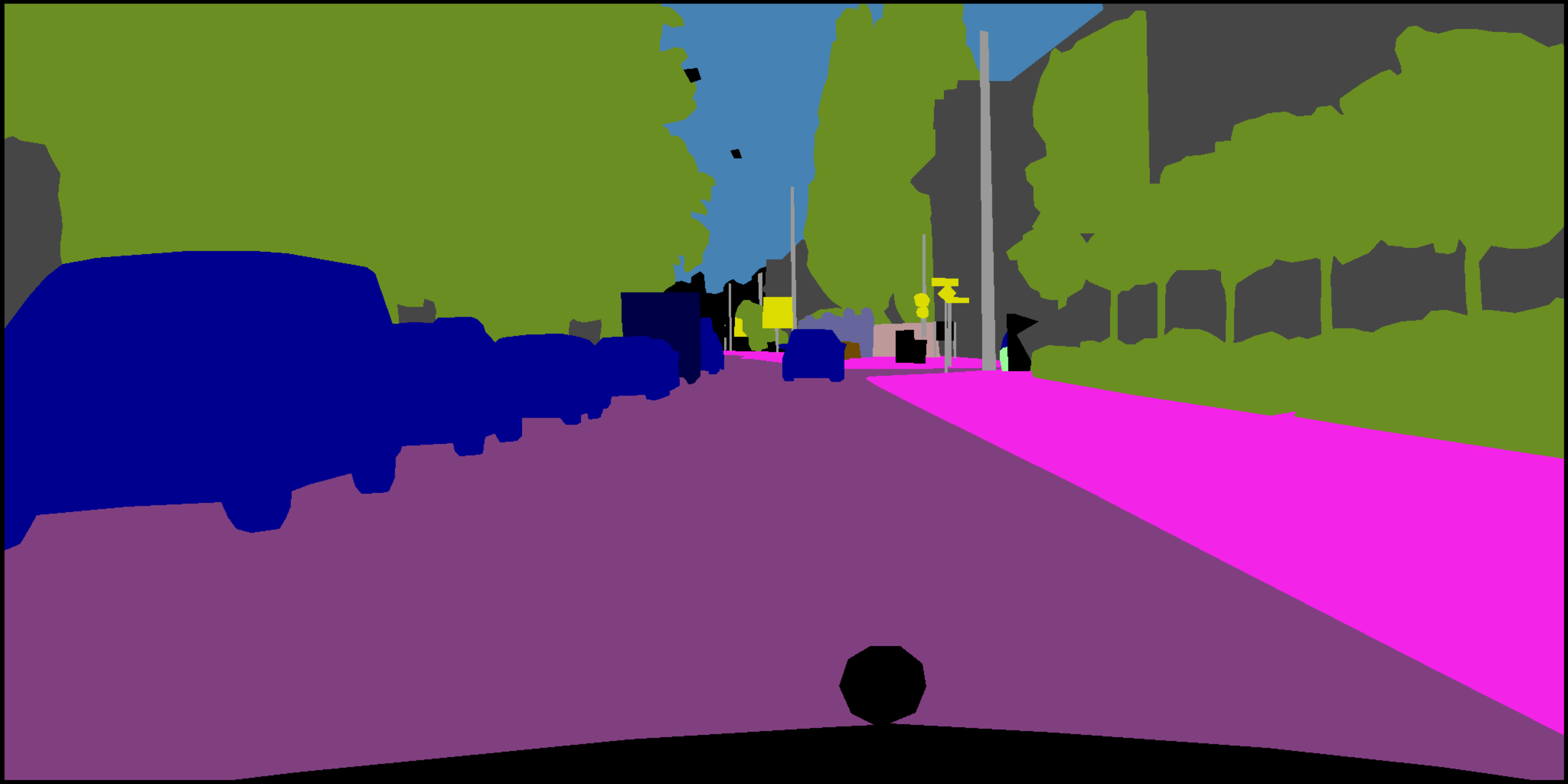}
\includegraphics[width=0.192\textwidth]{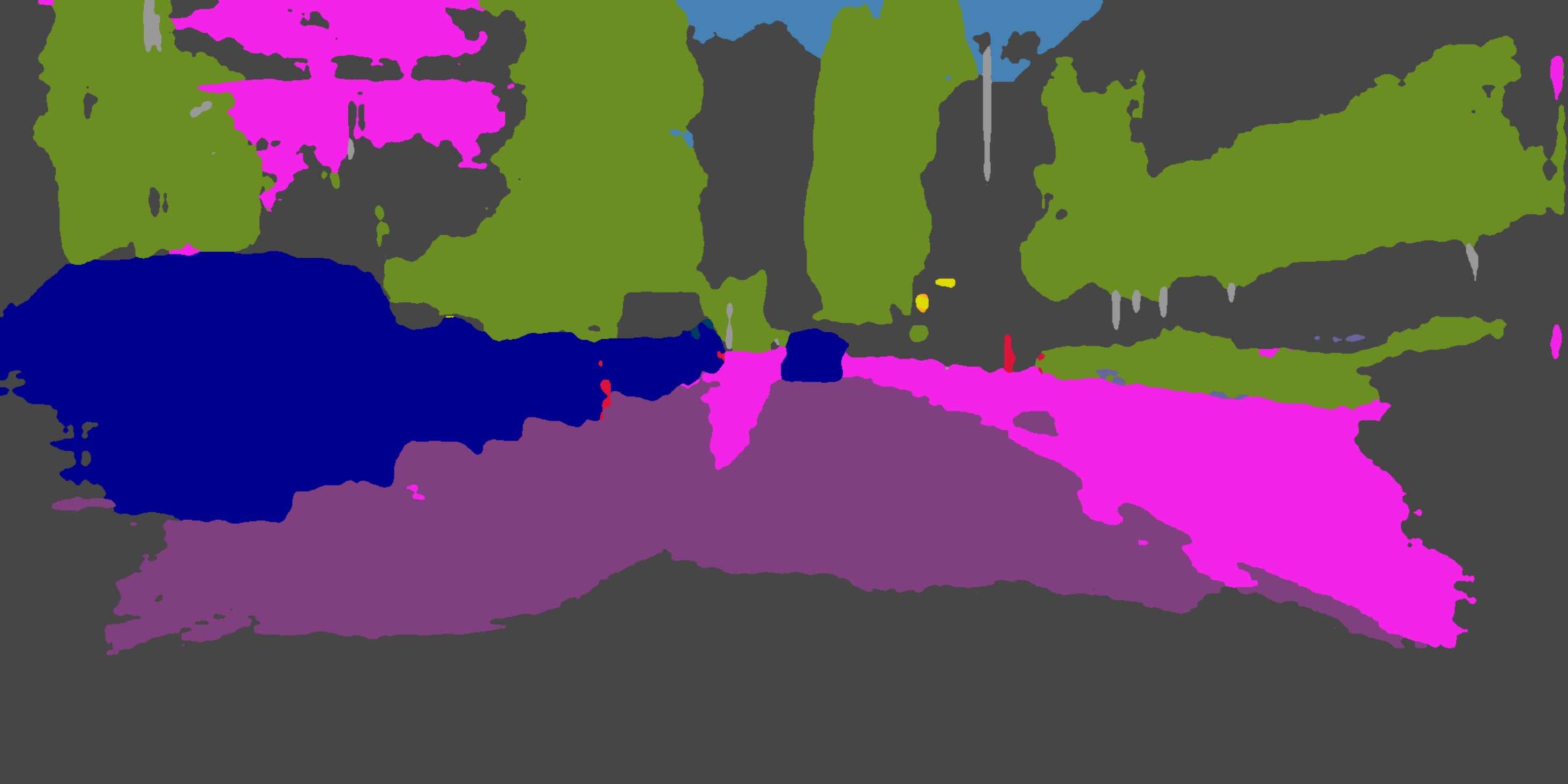}
\includegraphics[width=0.192\textwidth]{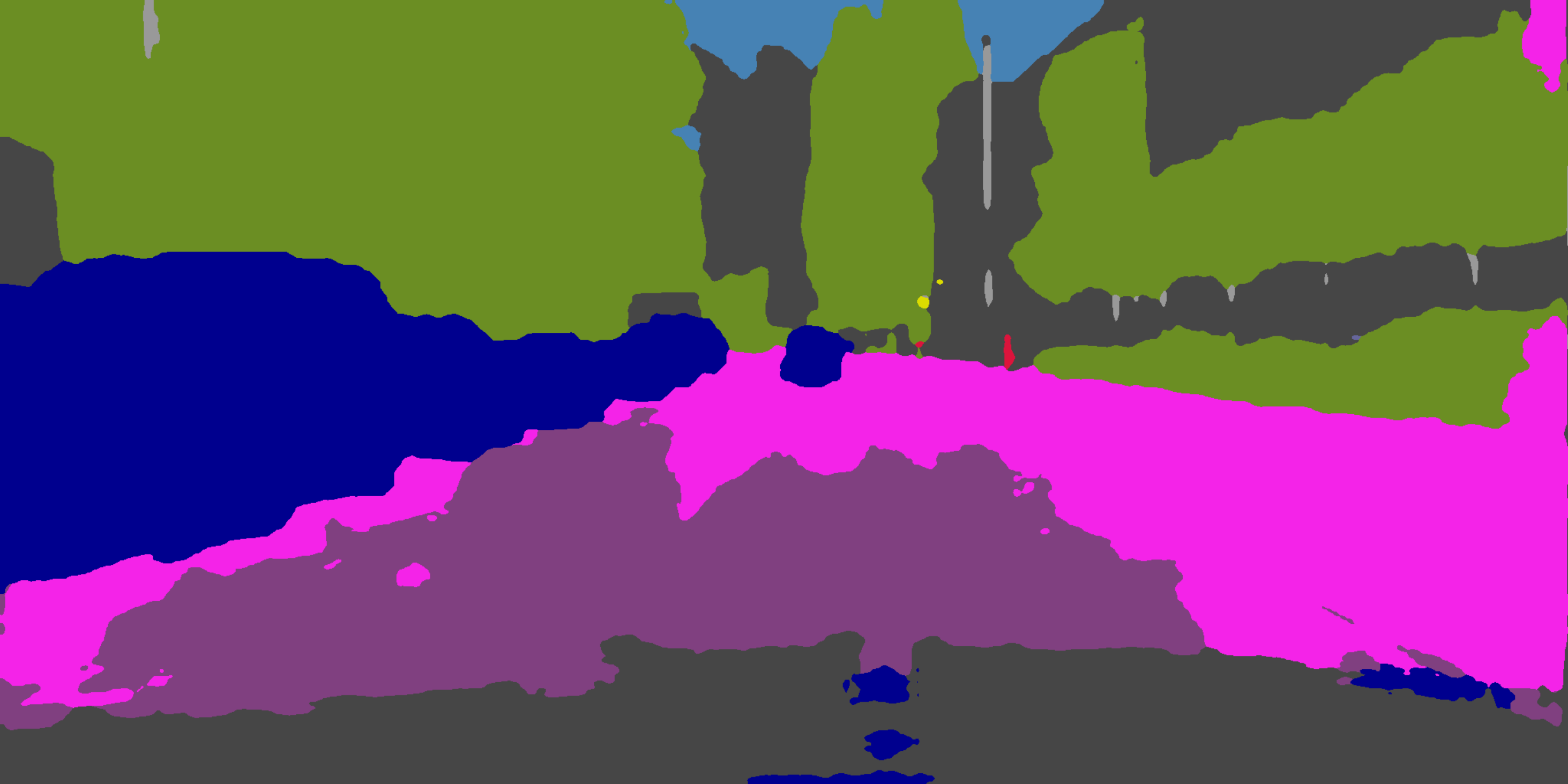}
\includegraphics[width=0.192\textwidth]{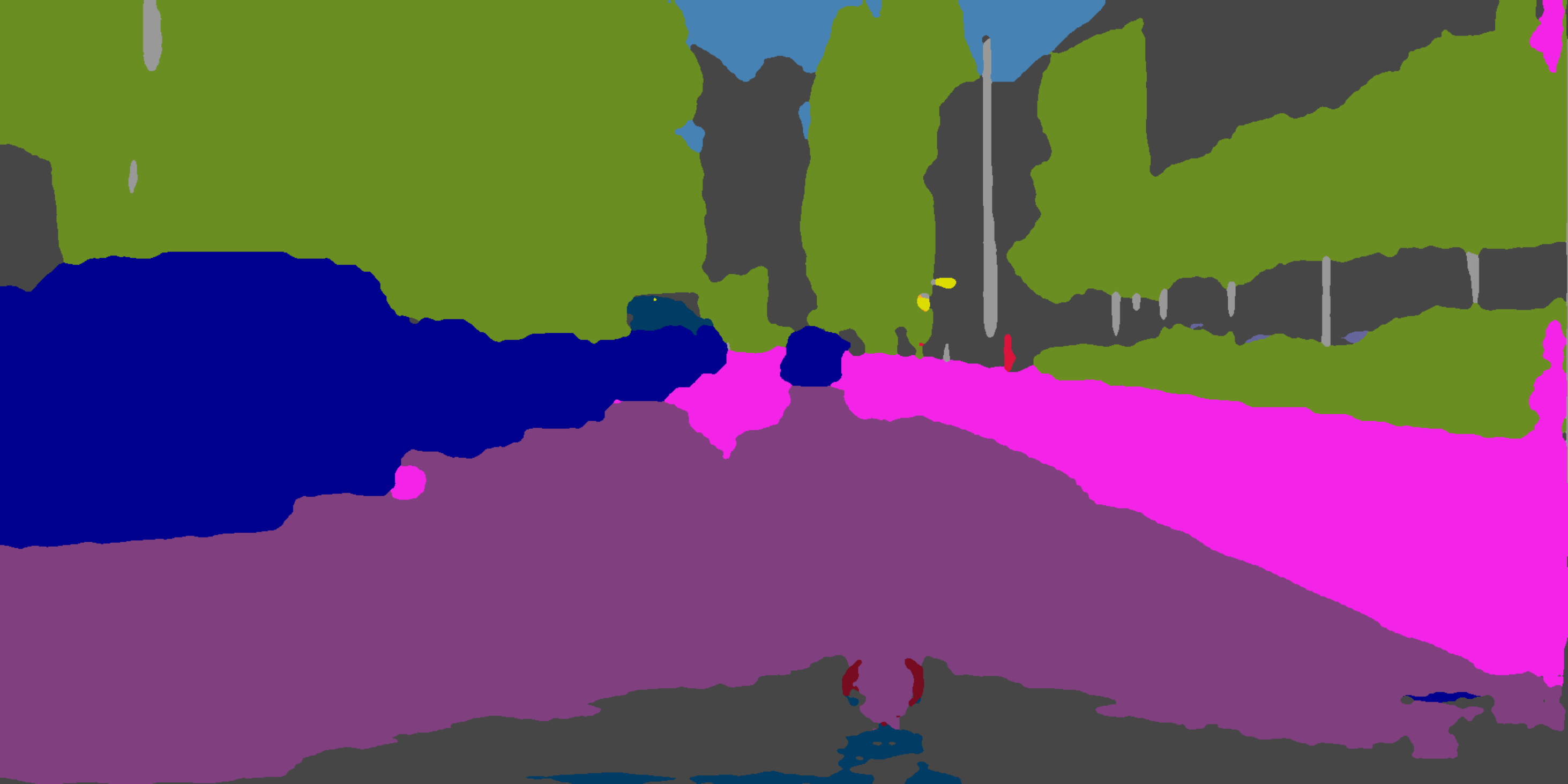}
\caption{Adaptation results on SYNTHIA $\rightarrow$ Cityscapes. Rows correspond to sampled images and predictions. Columns correspond to original images, ground truths, and results of source ResNet-38, ST, CBST. Best viewed in color.}\label{syn2city}
\end{figure}

Table \ref{t_syncity} reports the results, where mIoU* is the mean IoU of 13 classes, excluding the classes with *. With FCN8s-VGG16 as backbone, our CBST provides competitive performance compared with other methods. Equipped with a better backbone ResNet-38, CBST achieves the superior performance outperforming state-of-the-art by 1.7. Compared with ST, CBST with either FCN8s-VGG16 or ResNet-38 achieves better performance for mIoU and IoU of these initially not well-transfered classes, such as wall, rider, motorcycle and bike. The appearance of fence in SYNTHIA (car barriers) is extremely different from the fence in Cityscapes (pedestrian barriors) and it's very hard for the model to learn transferable knowledge for fence from SYNTHIA to Cityscapes. Figure \ref{syn2city} gives the visualization segmentation results in Cityscapes.

\begin{table*}[!t]
\centering
\caption{Experimental results of SYNTHIA $\rightarrow$ Cityscapes}\label{t_syncity}
\vspace{1mm}
\resizebox{\linewidth}{!}{%
\begin{tabular}{c|c|cccccccccccccccc|c|c}
\hline
Method      & Base Net          & Road & SW   & Build & Wall* & Fence* & Pole* & TL   & TS   & Veg. & Sky  & PR   & Rider & Car  & Bus  & Motor & Bike & mIoU & mIoU* \\ \hline
Source only \cite{hoffman2016fcns} & Dilation-Frontend & 6.4  & 17.7 & 29.7  & 1.2  & 0.0   & 15.1 & 0.0  & 7.2  & 30.3 & 66.8 & 51.1 & 1.5   & 47.3 & 3.9  & 0.1   & 0.0  & 17.4 & 20.2 \\
FCN wild \cite{hoffman2016fcns}   &  \cite{yu2016multi} & 11.5 & 19.6 & 30.8  & 4.4  & 0.0   & 20.3 & 0.1  & 11.7 & 42.3 & 68.7 & 51.2 & 3.8   & 54.0 & 3.2  & 0.2   & 0.6  & 20.2 & 22.1\\ \hline
Source only \cite{Zhang_2017_ICCV} & FCN8s-VGG16       & 5.6  & 11.2 & 59.6  & 8.0  & \textbf{0.5}   & 21.5 & 8.0  & 5.3  & 72.4 & 75.6 & 35.1 & 9.0   & 23.6 & 4.5  & 0.5   & 18.0 & 22.0 & 27.6 \\
Curr. DA \cite{Zhang_2017_ICCV}   &  \cite{long2015fully}    & 65.2 & 26.1 & 74.9  & 0.1  & \textbf{0.5} & 10.7 & 3.5  & 3.0  & 76.1 & 70.6 & 47.1 & 8.2   & 43.2 & 20.7 & 0.7   & 13.1 & 29.0 & 34.8 \\ \hline
Source only & FCN8s-VGG16       & 24.1 & 19.1 & 68.5 & 0.9 & 0.3 & 16.4 & 5.7 & 10.8 & 75.2 & 76.3 & 43.2 & 15.2 & 26.7 & 15.0 & 5.9 & 8.5 & 25.7 & 30.3\\
GAN DA    &  \cite{long2015fully}  & 79.1 & 31.1 & 77.1 & 3.0 & 0.2 & 22.8 & 6.6 & 15.2 & 77.4 & 78.9 & 47.0 & 14.8 & 67.5 & 16.3 & 6.9 & 13.0 & 34.8 & 40.8 \\ \hline
Source only & DeepLab-v2 \cite{tsai2018learning}       &  55.6 & 23.8 & 74.6 & $-$ & $-$ & $-$ & 6.1 & 12.1 & 74.8 & 79.0 & 55.3 & 19.1 & 39.6 & 23.3 & 13.7 & 25.0 & $-$ & 38.6\\
MAA    &  \cite{tsai2018learning}  & \textbf{84.3} & \textbf{42.7} & \textbf{77.5} & $-$ & $-$ & $-$ & 4.7 & 7.0 & 77.9 & \textbf{82.5} & 54.3 & \textbf{21.0} & 72.3 & \textbf{32.2} & \textbf{18.9} & 32.3 & $-$ & 46.7 \\ \hline
Source only & FCN8s-VGG16         & 17.2 & 19.7 & 47.3 & 1.1 & 0.0 & 19.1 & 3.0 & 9.1 & 71.8 & 78.3 & 37.6 & 4.7 & 42.2 & 9.0 & 0.1 & 0.9 & 22.6 & 26.2\\
ST        &  \cite{long2015fully} & 0.2 & 14.5 & 53.8 & 1.6 & 0.0 & 18.9 & 0.9 & 7.8 & 72.2 & 80.3 & 48.1 & 6.3 & 67.7 & 4.7 & 0.2 & 4.5 & 23.9 & 27.8\\
CBST      &  & 69.6 & 28.7 & 69.5 & 12.1 & 0.1 & 25.4 & 11.9 & 13.6 & 82.0 & 81.9 & 49.1 & 14.5 & 66.0 & 6.6 & 3.7 & 32.4 & 35.4 & 36.1\\ \hline
Source only & ResNet-38         & 32.6 & 21.5 & 46.5 & 4.8 & 0.1 & 26.5 & 14.8 & 13.1 & 70.8 & 60.3 & 56.6 & 3.5   & 74.1 & 20.4 & 8.9  & 13.1 & 29.2 & 33.6\\
ST        &  \cite{wu2016wider}  &  38.2 & 19.6 & 70.2 & 3.9 & 0.0 & 31.9 & 17.6 & 17.2 & 82.4 & 68.3 & 63.1 & 5.3 & 78.4 & 11.2 & 0.8 & 7.5 & 32.2 & 36.9 \\
CBST        &                   & 53.6 & 23.7 & 75.0 & \textbf{12.5} & 0.3 & \textbf{36.4} & \textbf{23.5} & \textbf{26.3} & \textbf{84.8} & 74.7 & \textbf{67.2} & 17.5 & \textbf{84.5} & 28.4 & 15.2 & \textbf{55.8} & \textbf{42.5} & \textbf{48.4}\\ \hline
\end{tabular}%
}
\end{table*}

\subsubsection{GTA5 to Cityscapes}
\begin{table*}[!t]
\centering
\caption{Experimental results for GTA5 $\rightarrow$ Cityscapes}\label{t_gtacity}
\vspace{1mm}
\resizebox{\linewidth}{!}{%
\begin{tabular}{c|c|ccccccccccccccccccc|c}
\hline
Method         & Base Net          & Road & SW   & Build & Wall & Fence & Pole & TL   & TS   & Veg. & Terrain & Sky  & PR   & Rider & Car  & Truck & Bus  & Train & Motor & Bike & mIoU \\ \hline
Source only \cite{hoffman2016fcns}   & Dilation-Frontend & 31.9 & 18.9 & 47.7  & 7.4  & 3.1   & 16.0 & 10.4 & 1.0  & 76.5 & 13.0    & 58.9 & 36.0 & 1.0   & 67.1 & 9.5   & 3.7  & 0.0   & 0.0   & 0.0  & 21.2 \\
FCN wild \cite{hoffman2016fcns}      & \cite{yu2016multi} & 70.4 & 32.4 & 62.1  & 14.9 & 5.4   & 10.9 & 14.2 & 2.7  & 79.2 & 21.3    & 64.6 & 44.1 & 4.2   & 70.4 & 8.0   & 7.3  & 0.0   & 3.5   & 0.0  & 27.1 \\ \hline
Source only \cite{Zhang_2017_ICCV}   & FCN8s-VGG16       & 18.1 & 6.8  & 64.1  & 7.3  & 8.7   & 21.0 & 14.9 & 16.8 & 45.9 & 2.4     & 64.4 & 41.6 & 17.5  & 55.3 & 8.4   & 5.0  & 6.9   & 4.3   & 13.8 & 22.3 \\
Curr. DA \cite{Zhang_2017_ICCV}      & \cite{long2015fully} & 74.9 & 22.0 & 71.7  & 6.0  & 11.9  & 8.4  & 16.3 & 11.1 & 75.7 & 13.3    & 66.5 & 38.0 & 9.3   & 55.2 & 18.8  & 18.9 & 0.0   & 16.8  & 16.6 & 28.9 \\ \hline
Source only \cite{hoffman2018cycada}   & FCN8s-VGG16       & 26.0 & 14.9 & 65.1  & 5.5  & 12.9  & 8.9  & 6.0  & 2.5  & 70.0 & 2.9     & 47.0 & 24.5 & 0.0   & 40.0 & 12.1  & 1.5  & 0.0   & 0.0   & 0.0  & 17.9 \\
CyCADA \cite{hoffman2018cycada}        & \cite{long2015fully} & 85.2 & 37.2 & 76.5  & 21.8 & 15.0  & 23.8 & 22.9 & 21.5 & 80.5 & 31.3    & 60.7 & 50.5 & 9.0   & 76.9 & 17.1  & 28.2 & 4.5   & 9.8   & 0.0  & 35.4 \\ \hline
Source only \cite{hoffman2018cycada}   & Dilated ResNet-26 & 42.7 & 26.3 & 51.7  & 5.5  & 6.8   & 13.8 & 23.6 & 6.9  & 75.5 & 11.5    & 36.8 & 49.3 & 0.9   & 46.7 & 3.4   & 5.0  & 0.0   & 5.0   & 1.4  & 21.7 \\
CyCADA \cite{hoffman2018cycada}        & \cite{yu2017dilated} & 79.1 & 33.1 & 77.9  & 23.4 & 17.3  & 32.1 & 33.3 & 31.8 & 81.5 & 26.7    & 69.0 & 62.8 & 14.7  & 74.5 & 20.9  & 25.6 & 6.9   & 18.8  & 20.4 & 39.5 \\ \hline
Source only \cite{saito2017adversarial}   & ResNet-50         & 64.5 & 24.9 & 73.7  & 14.8 & 2.5   & 18.0 & 15.9 & 0    & 74.9 & 16.4    & 72.0 & 42.3 & 0.0   & 39.5 & 8.6   & 13.4 & 0.0   & 0.0   & 0.0  & 25.3 \\
ADR \cite{saito2017adversarial} & \cite{he2016deep} & 87.8 & 15.6 & 77.4  & 20.6 & 9.7   & 19.0 & 19.9 & 7.7  & 82.0 & 31.5    & 74.3 & 43.5 & 9.0   & 77.8 & 17.5  & 27.7 & 1.8   & 9.7   & 0.0  & 33.3 \\ \hline
Source only \cite{murez2018image}   & DenseNet          & 67.3 & 23.1 & 69.4  & 13.9 & 14.4  & 21.6 & 19.2 & 12.4 & 78.7 & 24.5    & 74.8 & 49.3 & 3.7   & 54.1 & 8.7   & 5.3  & 2.6   & 6.2   & 1.9  & 29.0 \\
I2I Adapt \cite{murez2018image}  & \cite{huang2017densely} & 85.8 & 37.5 & 80.2  & 23.3 & 16.1  & 23.0 & 14.5 & 9.8  & 79.2 & \textbf{36.5} & \textbf{76.4} & 53.4 & 7.4   & 82.8 & 19.1  & 15.7 & 2.8   & 13.4  & 1.7  & 35.7 \\ \hline
Source only \cite{tsai2018learning}   & DeepLab-v2          &  75.8 & 16.8 & 77.2 & 12.5 & 21.0 & 25.5 & 30.1 & 20.1 & 81.3 & 24.6 & 70.3 & 53.8 & 26.4 & 49.9 & 17.2 & 25.9 & 6.5 & 25.3 & 36.0 & 36.6\\
MAA \cite{tsai2018learning}  & \cite{huang2017densely} & 86.5 & 36.0 & \textbf{79.9} & 23.4 & 23.3 & 23.9 & 35.2 & 14.8 & 83.4 & 33.3 & 75.6 & 58.5 & 27.6 & 73.7 & 32.5 & 35.4 & 3.9 & 30.1 & 28.1 & 42.4 \\ \hline
Source only    & FCN8s-VGG16       & 64.0 & 22.1 & 68.6  & 13.3 & 8.7   & 19.9 & 15.5 & 5.9  & 74.9 & 13.4    & 37.0 & 37.7 & 10.3  & 48.2 & 6.1   & 1.2  & 1.8   & 10.8  & 2.9  & 24.3 \\
ST             & \cite{hoffman2016fcns} & 83.8 & 17.4 & 72.1  & 14.3 & 2.9   & 16.5 & 16.0 & 6.8  & 81.4 & 24.2    & 47.2 & 40.7 & 7.6   & 71.7 & 10.2  & 7.6  & 0.5   & 11.1  & 0.9  & 28.1 \\
CBST           &                   & 66.7 & 26.8 & 73.7  & 14.8 & 9.5   & 28.3 & 25.9 & 10.1 & 75.5 & 15.7    & 51.6 & 47.2 & 6.2   & 71.9 & 3.7   & 2.2  & 5.4   & 18.9  & 32.4 & 30.9 \\
CBST-SP        &                   & \textbf{90.4} & 50.8 & 72.0  & 18.3 & 9.5   & 27.2 & 28.6 & 14.1 & 82.4 & 25.1    & 70.8 & 42.6 & 14.5  & 76.9 & 5.9   & 12.5 & 1.2   & 14.0  & 28.6 & 36.1 \\ \hline
Source only    & ResNet-38         & 70.0 & 23.7 & 67.8 & 15.4 & 18.1 & 40.2 & 41.9 & 25.3 & 78.8 & 11.7 & 31.4 & \textbf{62.9} & \textbf{29.8} & 60.1 & 21.5 & 26.8 & 7.7 & 28.1 & 12.0 & 35.4 \\
ST             & \cite{wu2016wider} & 90.1 & 56.8 & 77.9 & 28.5 & 23.0 & 41.5 & 45.2 & 39.6 & 84.8 & 26.4 & 49.2 & 59.0 & 27.4 & 82.3 & 39.7 & 45.6 & \textbf{20.9} & \textbf{34.8} & \textbf{46.2} & 41.5 \\
CBST           &                   & 86.8 & 46.7 & 76.9 & 26.3 & \textbf{24.8} & 42.0 & 46.0 & 38.6 & 80.7 & 15.7 & 48.0 & 57.3 & 27.9 & 78.2 & 24.5 & 49.6 & 17.7 & 25.5 & 45.1 & 45.2 \\
CBST-SP        &                   & 88.0 & 56.2 & 77.0 & 27.4 & 22.4 & 40.7 & 47.3 & \textbf{40.9} & 82.4 & 21.6 & 60.3 & 50.2 & 20.4 & \textbf{83.8} & 35.0 & \textbf{51.0} & 15.2 & 20.6 & 37.0 & 46.2 \\ 
CBST-SP+MST       &                   & 89.6 & \textbf{58.9} & 78.5 & \textbf{33.0} & 22.3 & \textbf{41.4} & \textbf{48.2} & 39.2 & \textbf{83.6} & 24.3 & 65.4 & 49.3 & 20.2 & 83.3 & \textbf{39.0} & 48.6 & 12.5 & 20.3 & 35.3 & \textbf{47.0} \\ \hline
\end{tabular}%
}
\end{table*}

\begin{figure}[!t]
\centering
\resizebox{1.0\textwidth}{!}{
	\begin{tabular}{@{}cccccccccc@{}}
		\cellcolor{city_color_1}\textcolor{white}{~~road~~} &
		\cellcolor{city_color_2}~~sidewalk~~&
		\cellcolor{city_color_3}\textcolor{white}{~~building~~} &
		\cellcolor{city_color_4}\textcolor{white}{~~wall~~} &
		\cellcolor{city_color_5}~~fence~~ &
		\cellcolor{city_color_6}~~pole~~ &
		\cellcolor{city_color_7}~~traffic lgt~~ &
		\cellcolor{city_color_8}~~traffic sgn~~ &
		\cellcolor{city_color_9}~~vegetation~~ \\
		\cellcolor{city_color_10}~~terrain~~ &
		\cellcolor{city_color_11}~~sky~~ &
		\cellcolor{city_color_12}\textcolor{white}{~~person~~} &
		\cellcolor{city_color_13}\textcolor{white}{~~rider~~} &
		\cellcolor{city_color_14}\textcolor{white}{~~car~~} &
		\cellcolor{city_color_15}\textcolor{white}{~~truck~~} &
		\cellcolor{city_color_16}\textcolor{white}{~~bus~~} &
		\cellcolor{city_color_17}\textcolor{white}{~~train~~} &
		\cellcolor{city_color_18}\textcolor{white}{~~motorcycle~~} &
		\cellcolor{city_color_19}\textcolor{white}{~~bike~~}
	\end{tabular}
}
\vspace{0.2mm}

\includegraphics[width=0.16\textwidth]{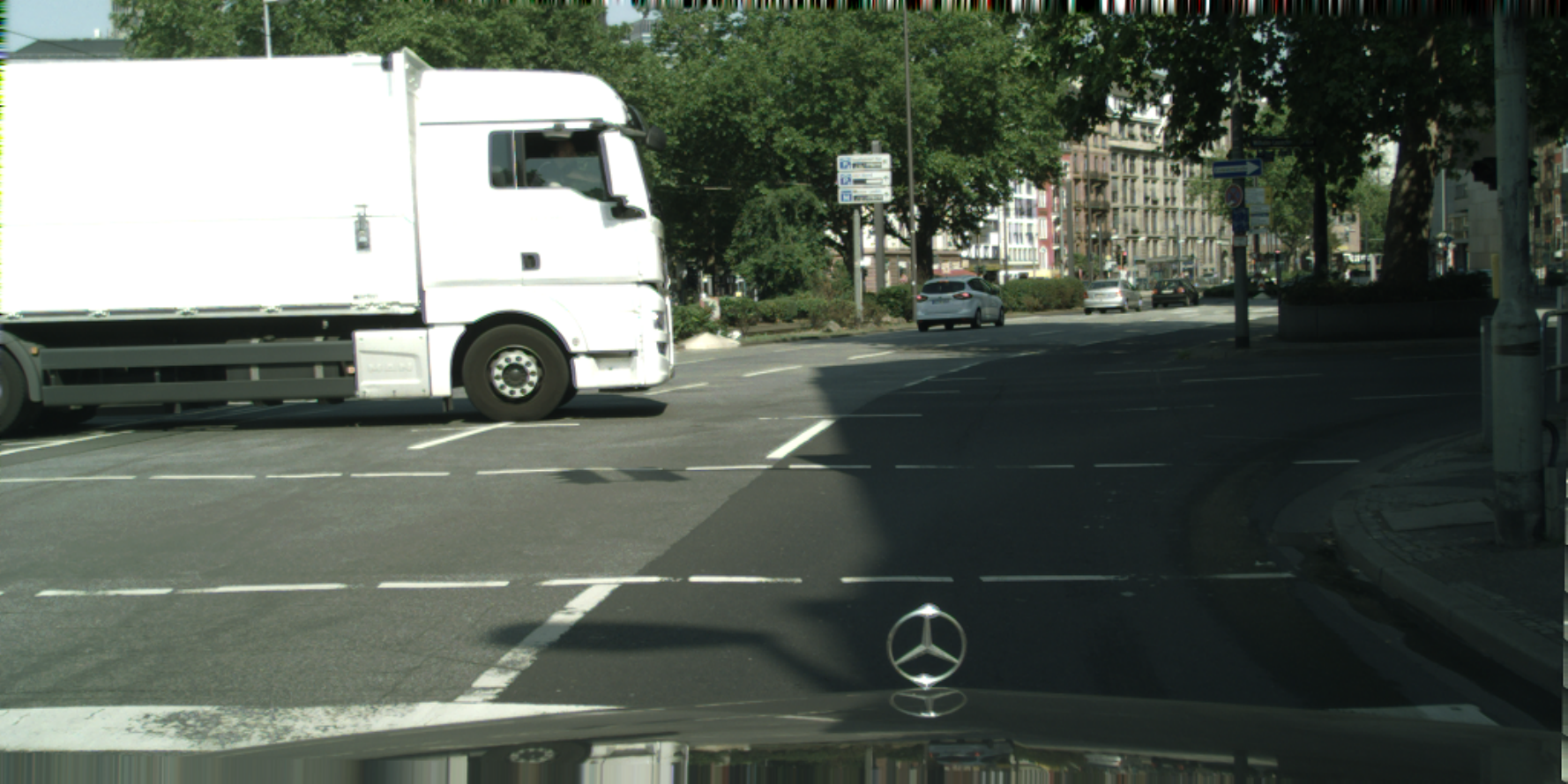}
\includegraphics[width=0.16\textwidth]{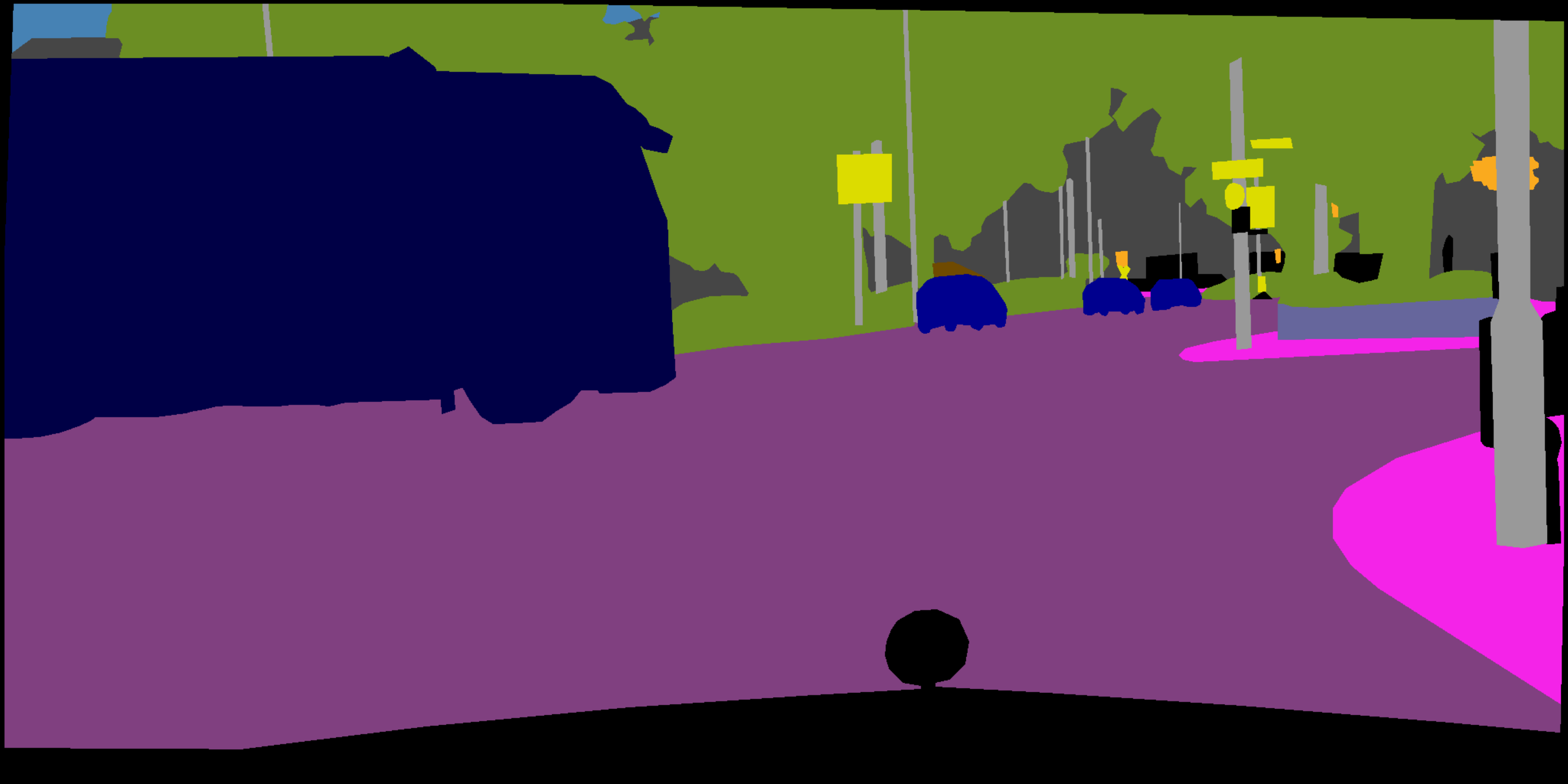}
\includegraphics[width=0.16\textwidth]{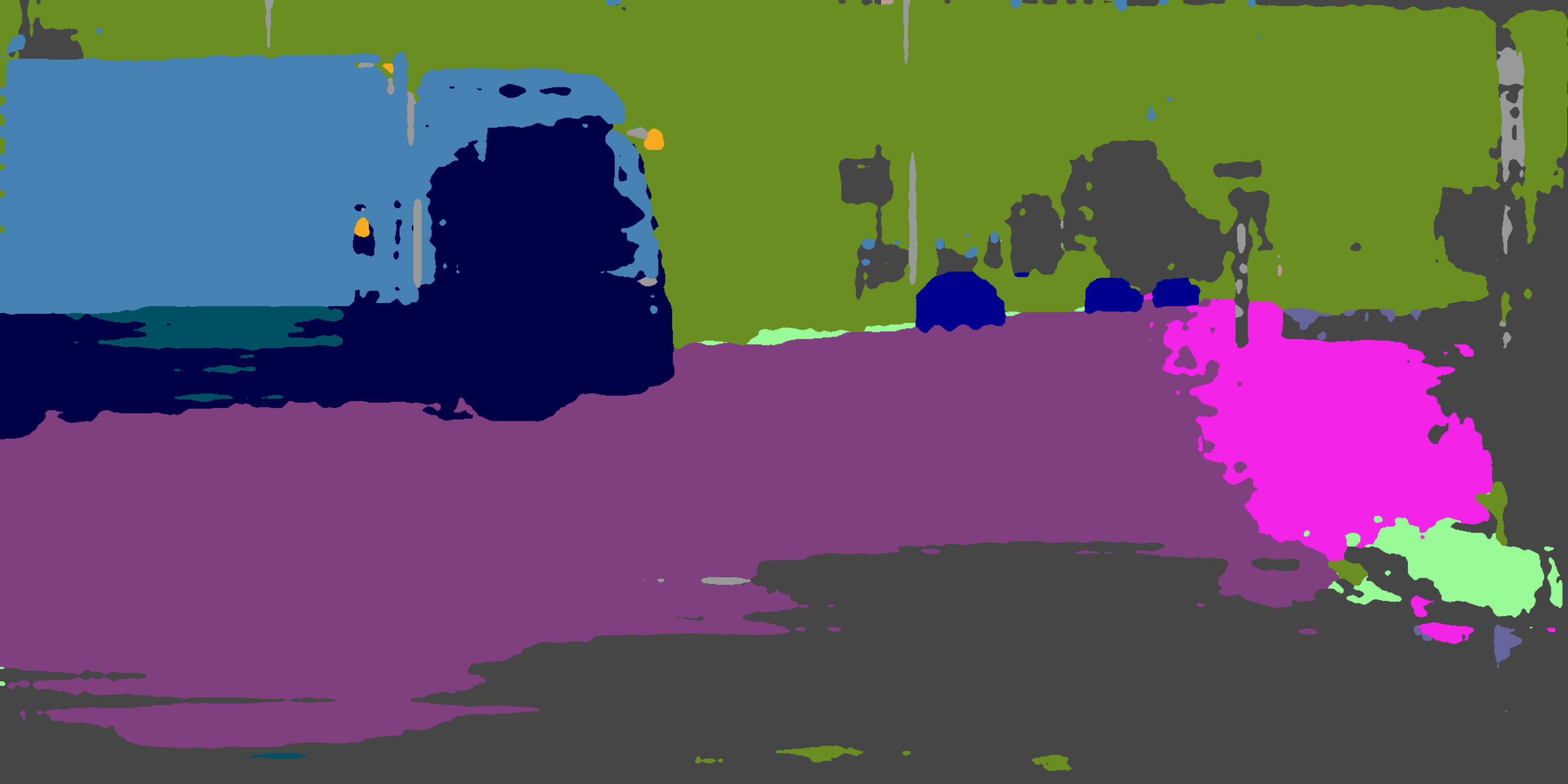}
\includegraphics[width=0.16\textwidth]{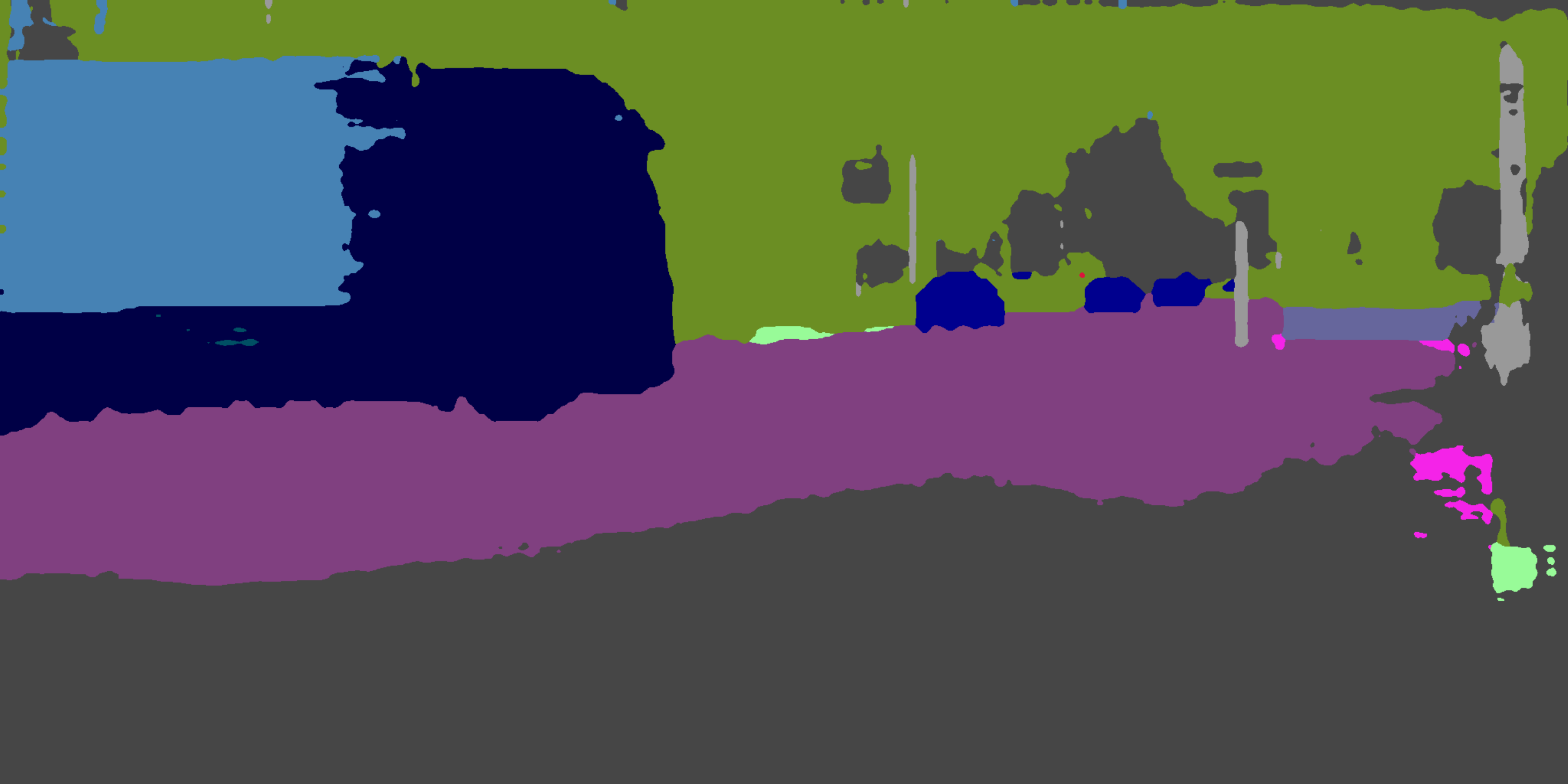}
\includegraphics[width=0.16\textwidth]{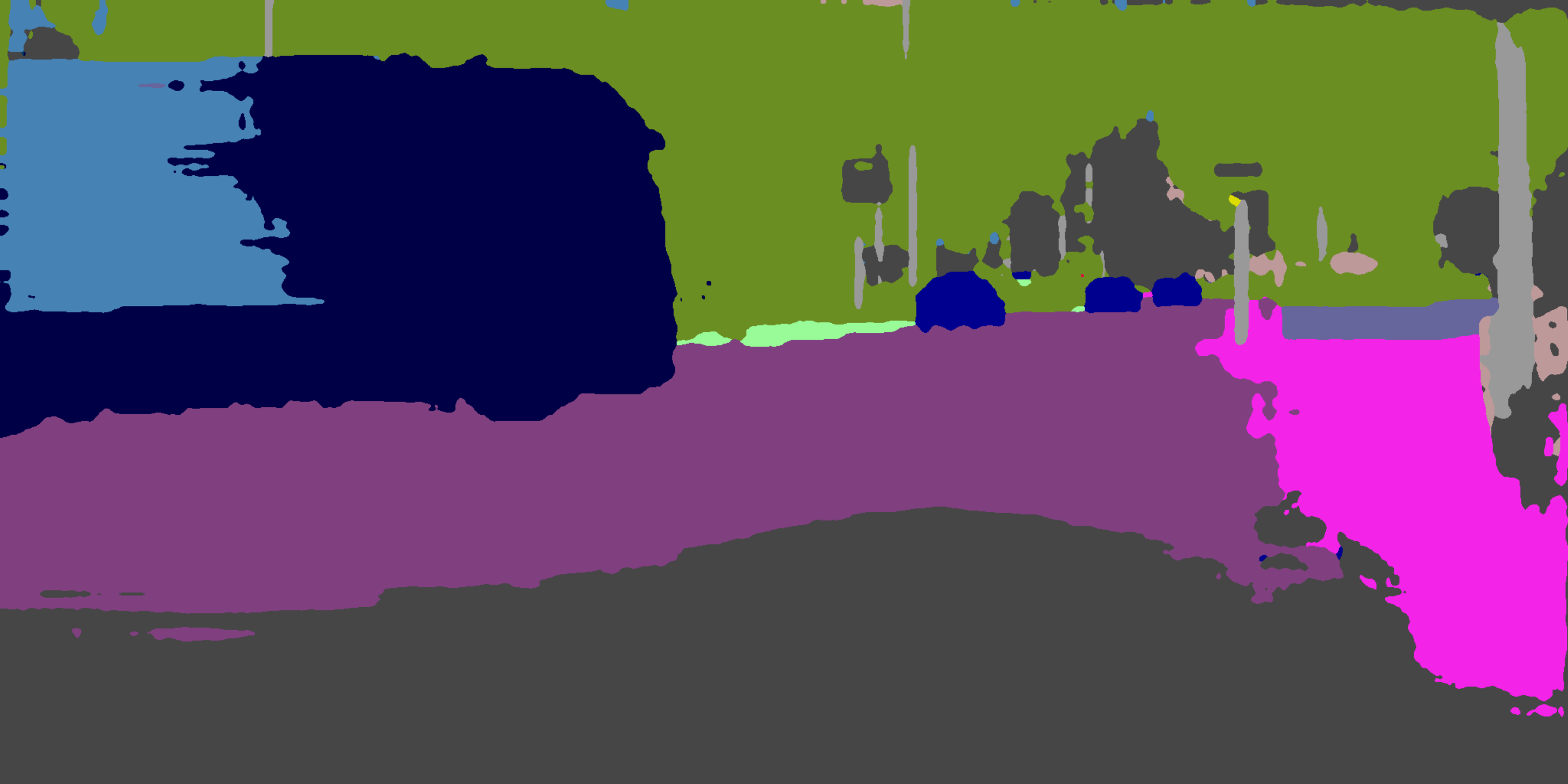}
\includegraphics[width=0.16\textwidth]{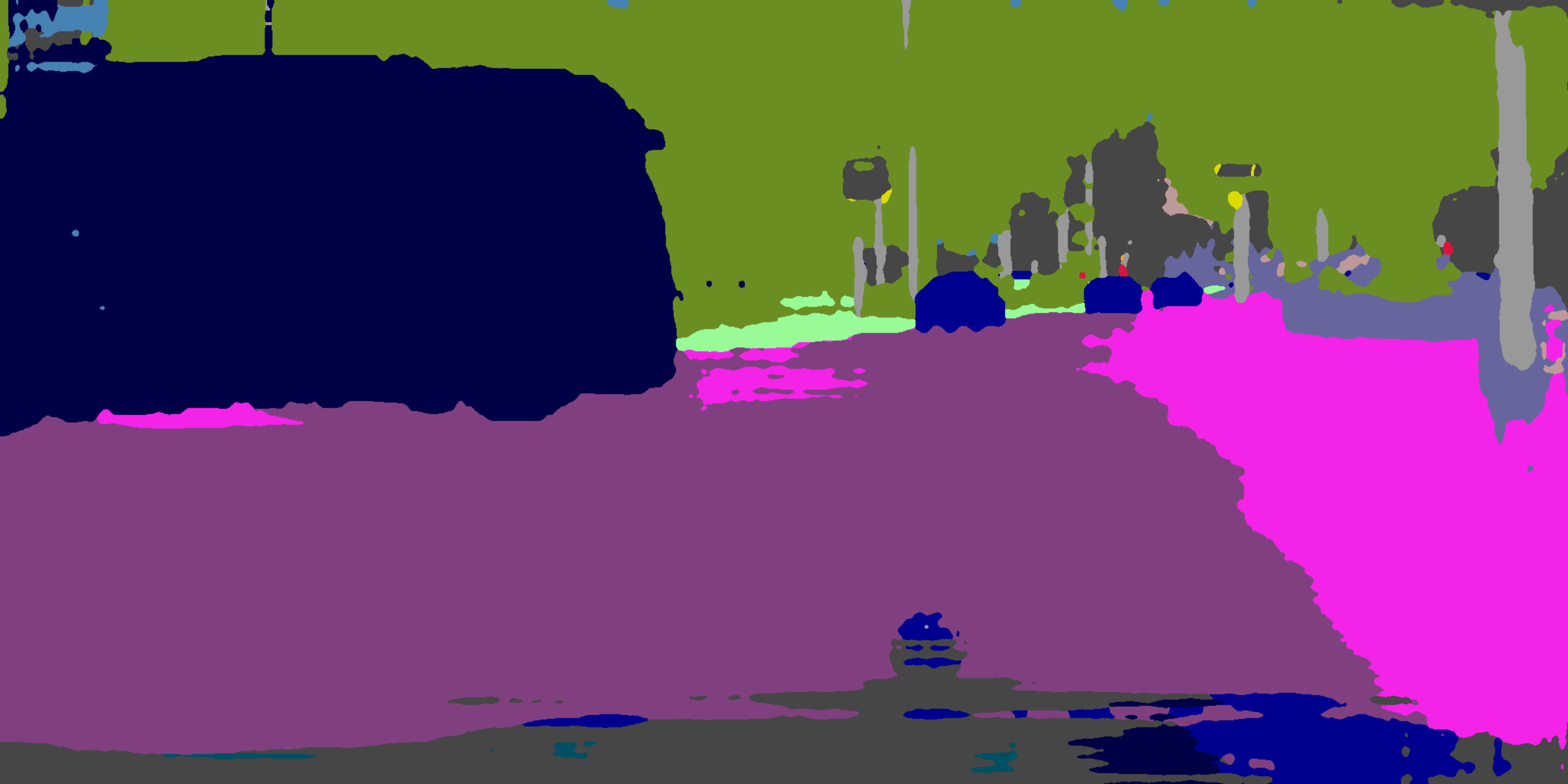}\\
\quad\\\vspace{-0.325cm}
\includegraphics[width=0.16\textwidth]{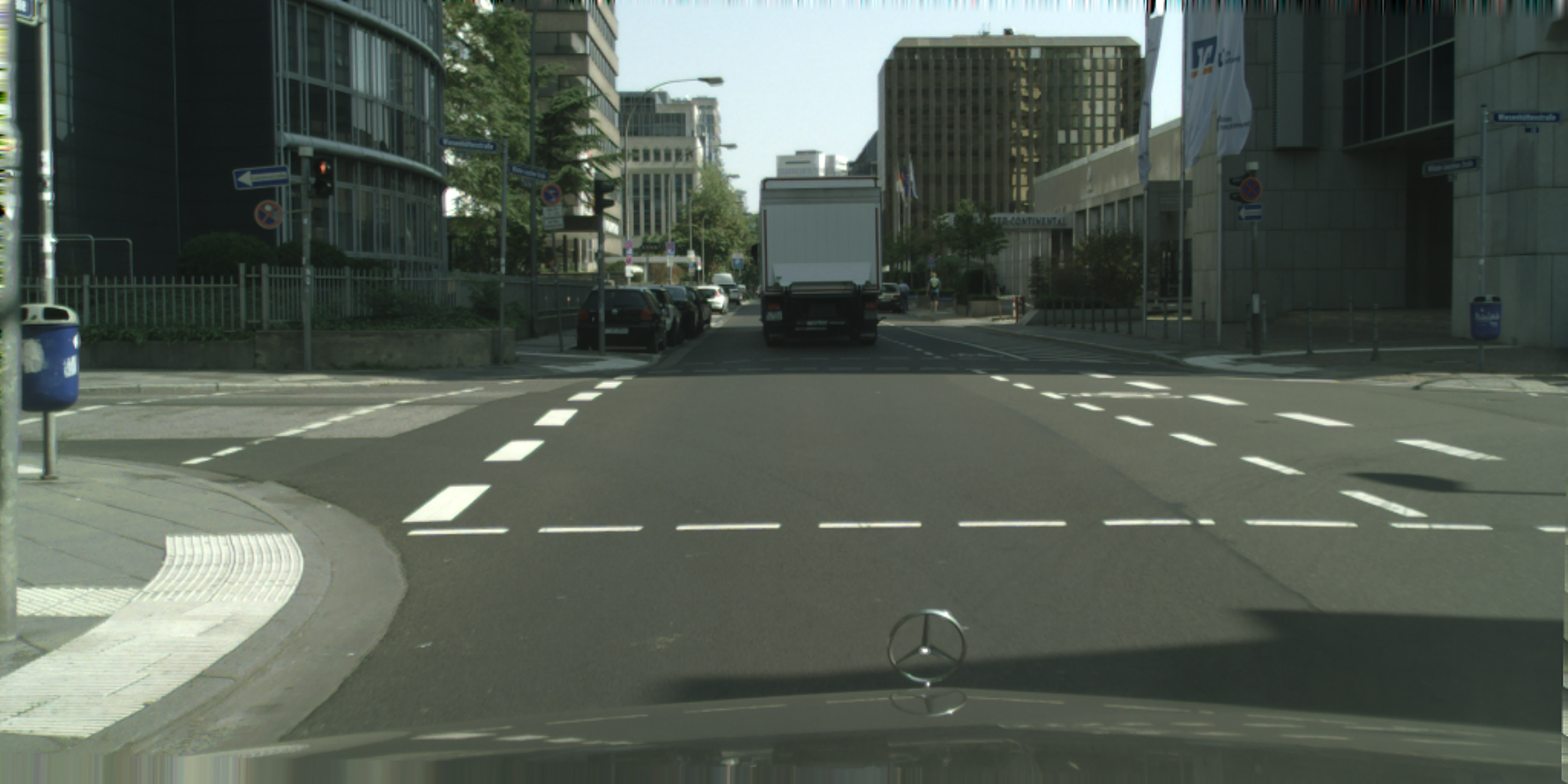}
\includegraphics[width=0.16\textwidth]{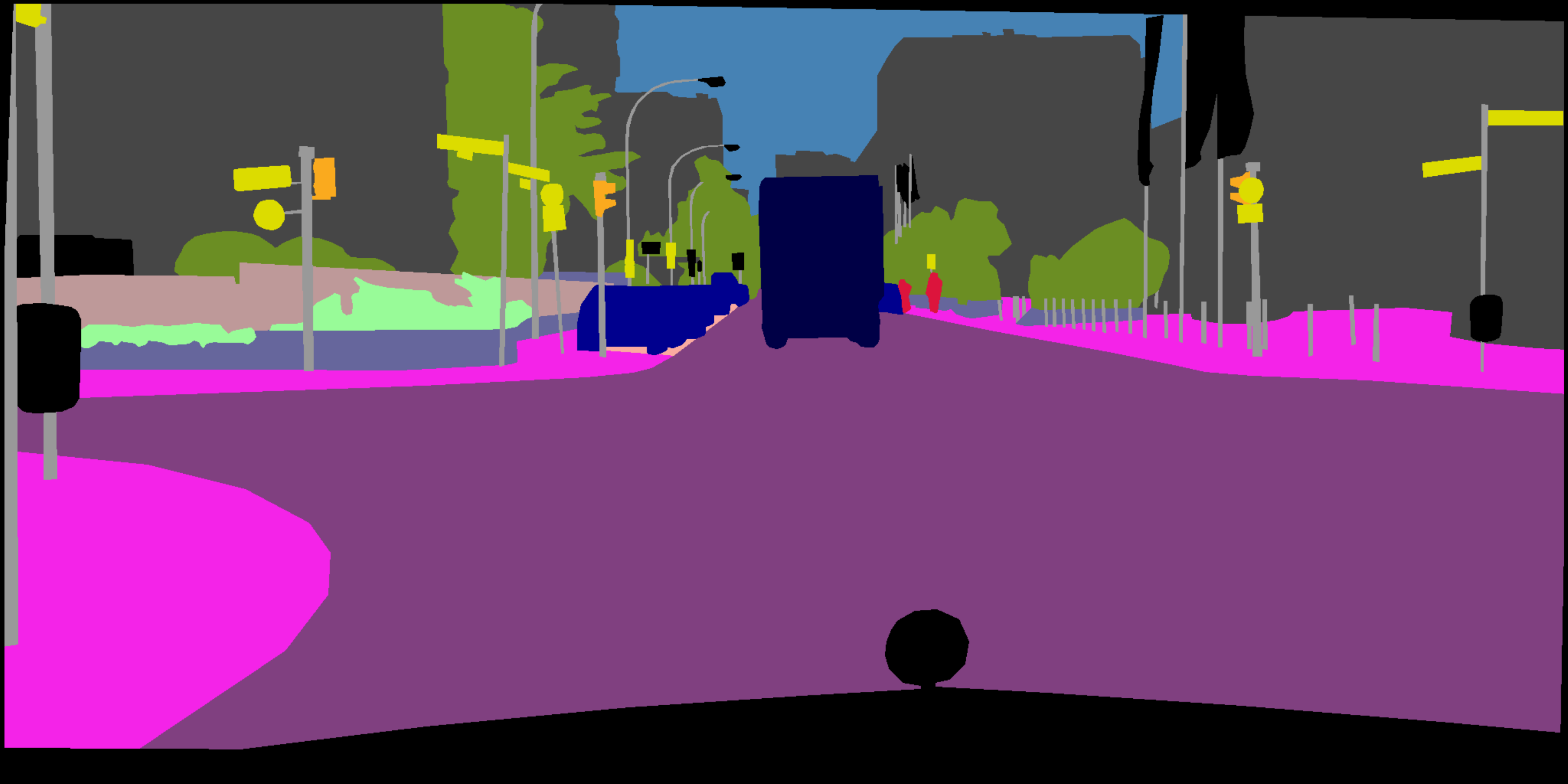}
\includegraphics[width=0.16\textwidth]{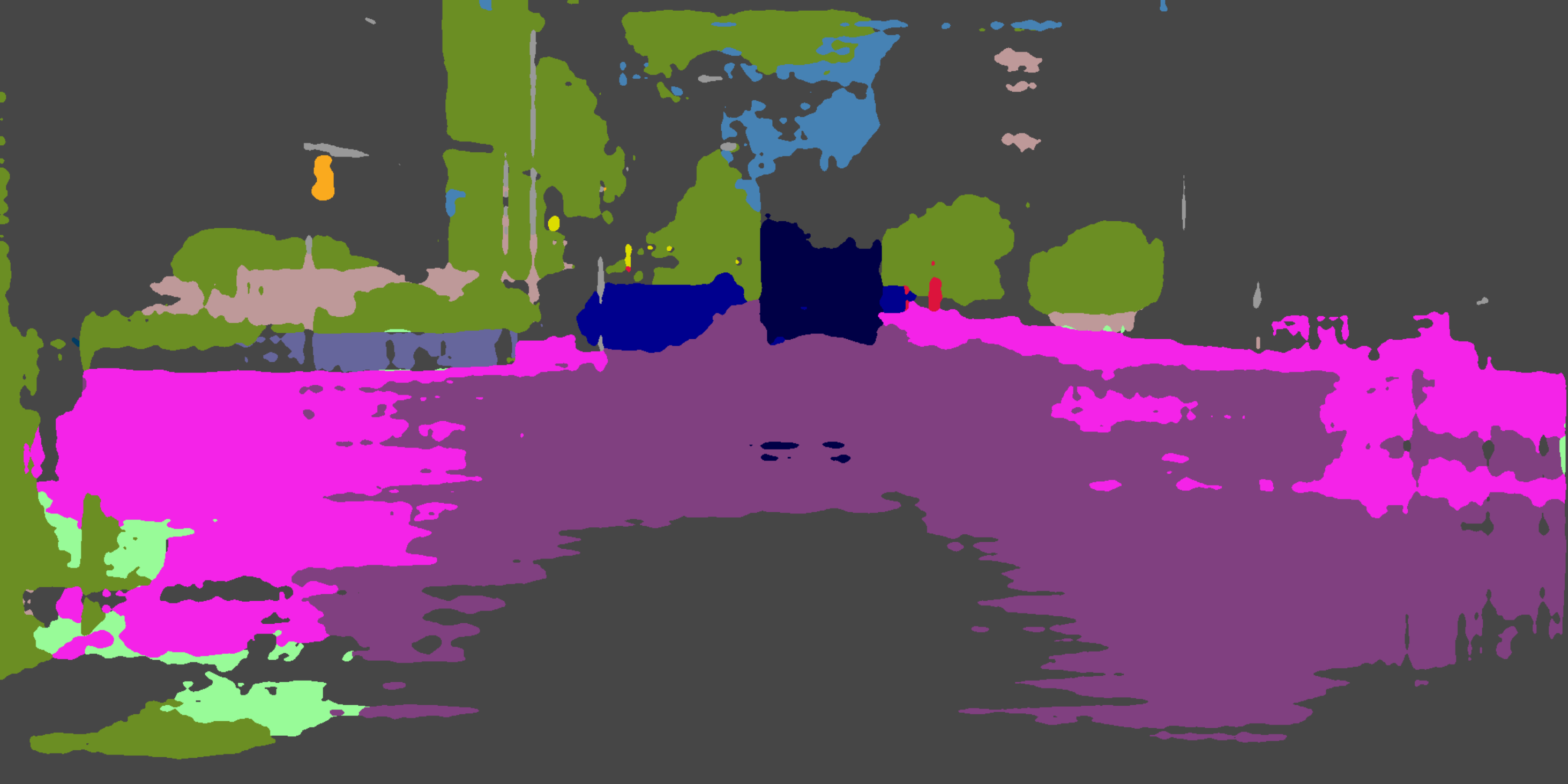}
\includegraphics[width=0.16\textwidth]{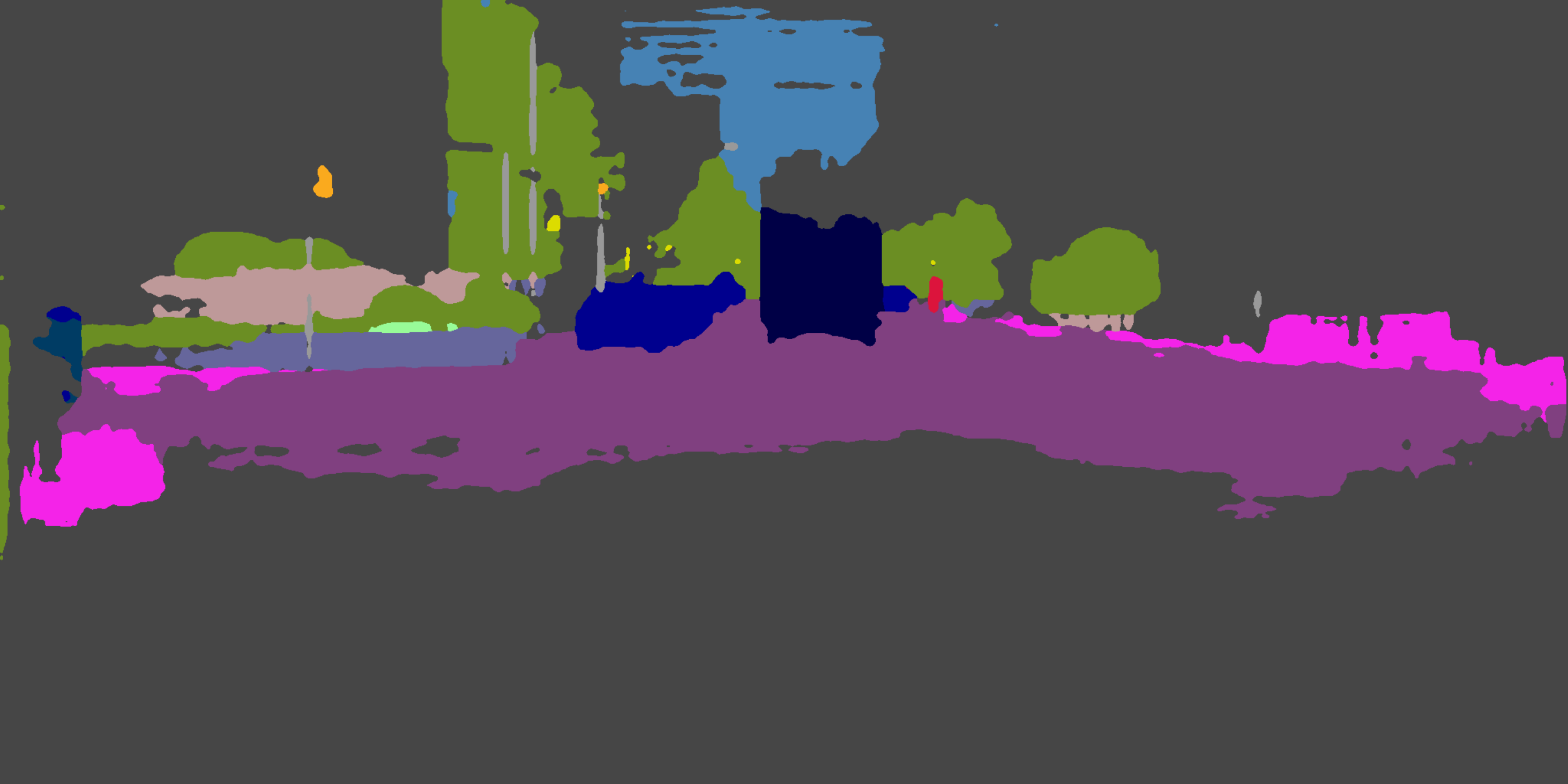}
\includegraphics[width=0.16\textwidth]{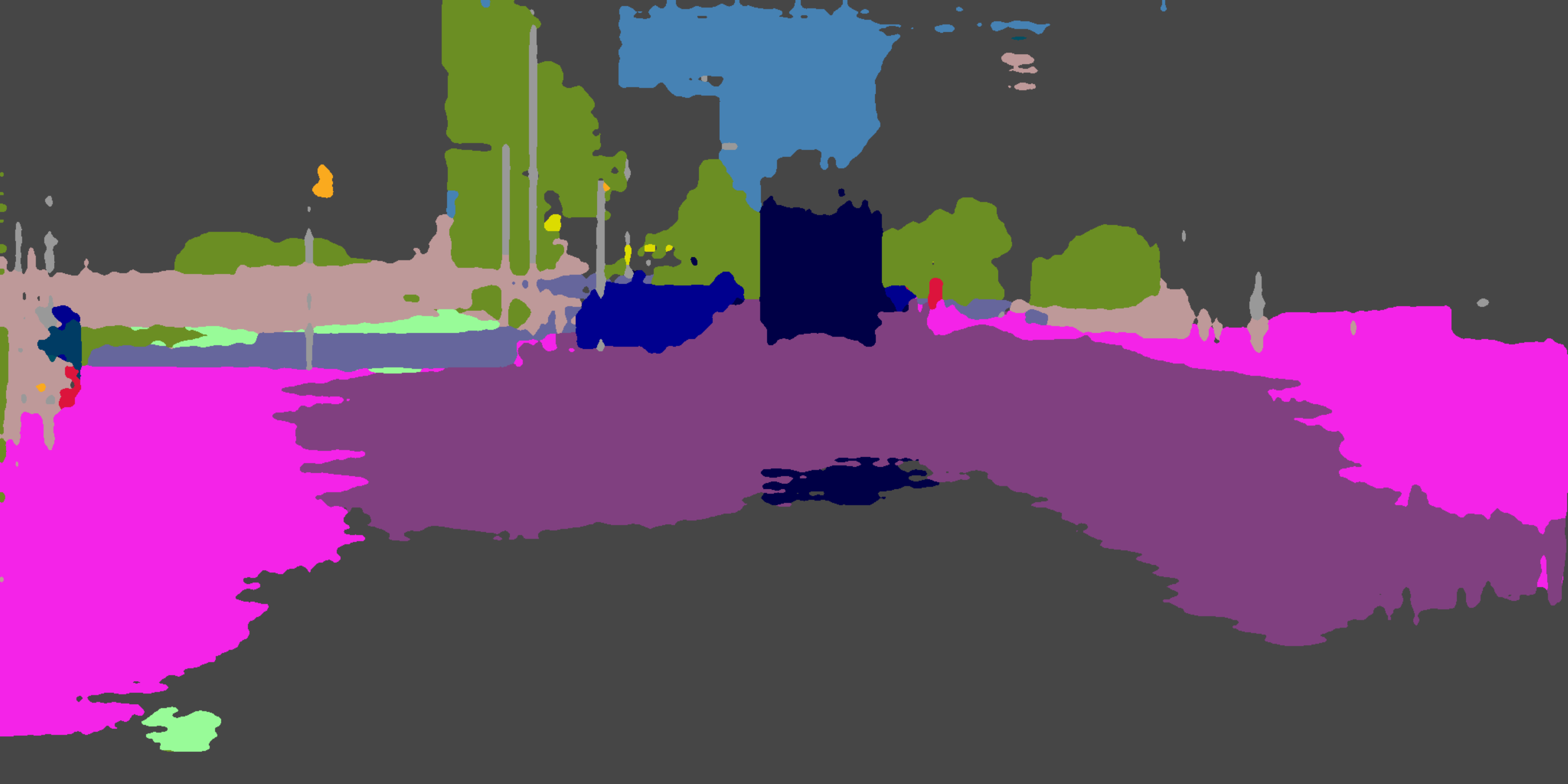}
\includegraphics[width=0.16\textwidth]{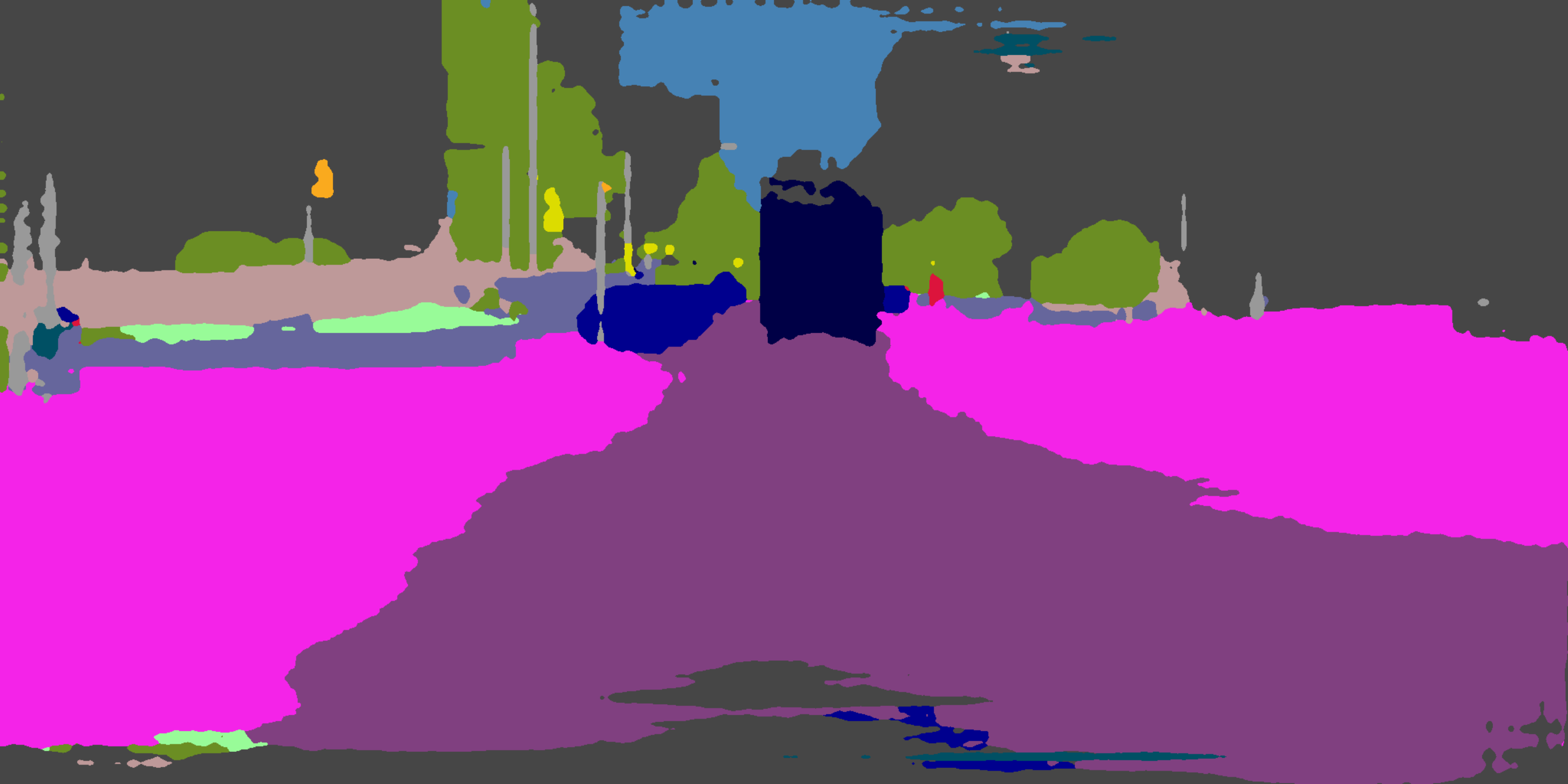}\\
\quad\\\vspace{-0.325cm}
\includegraphics[width=0.16\textwidth]{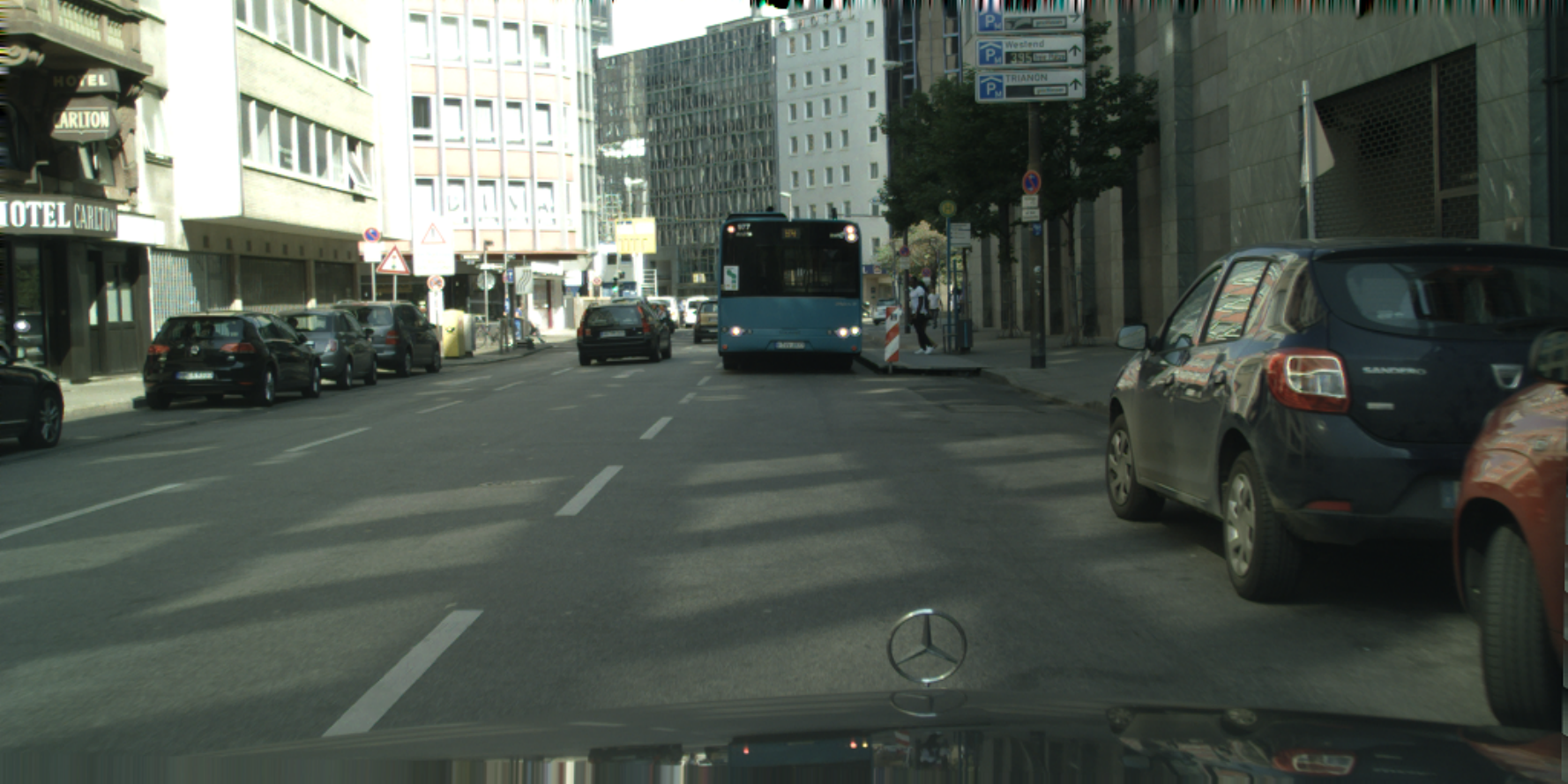}
\includegraphics[width=0.16\textwidth]{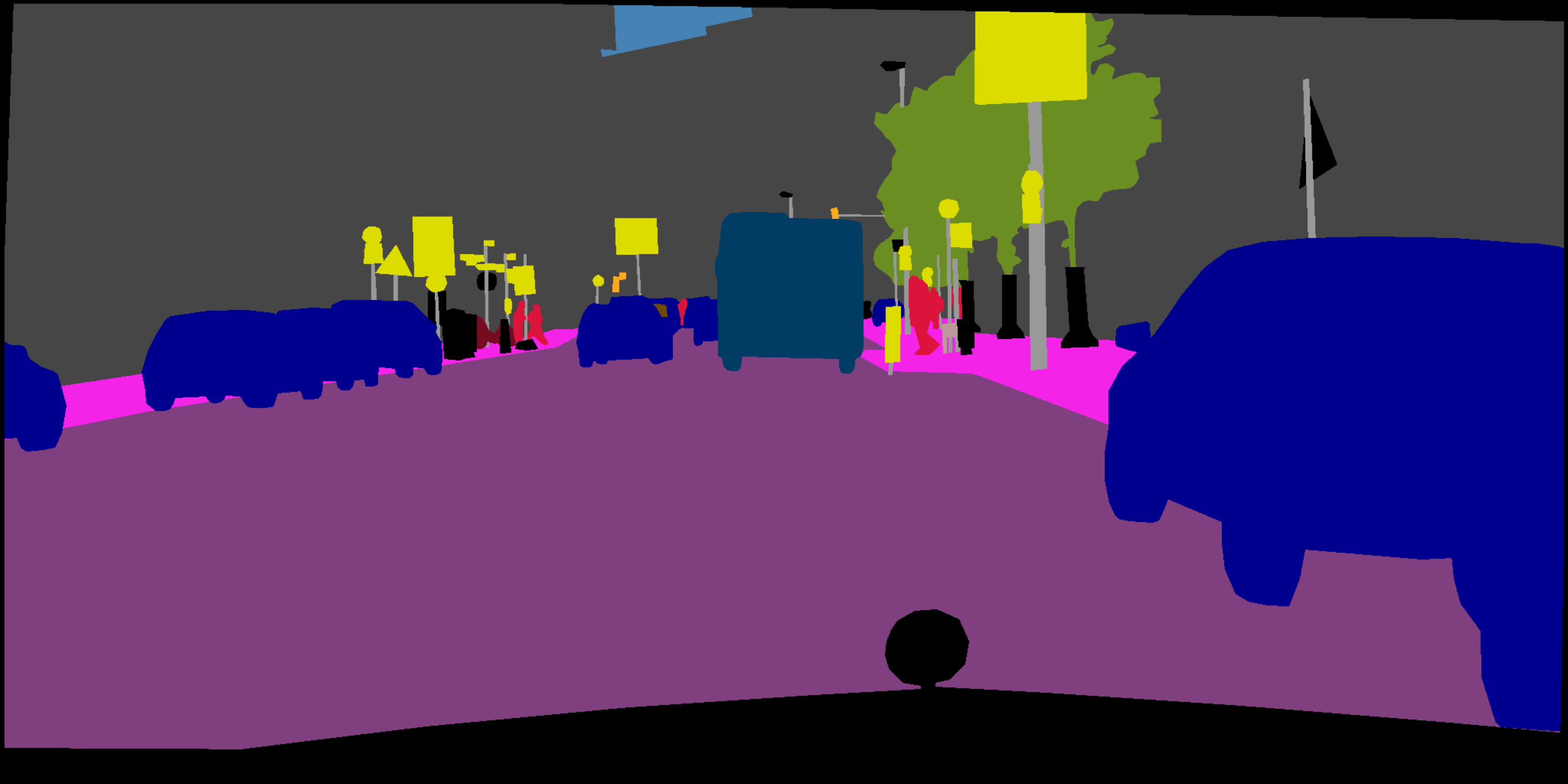}
\includegraphics[width=0.16\textwidth]{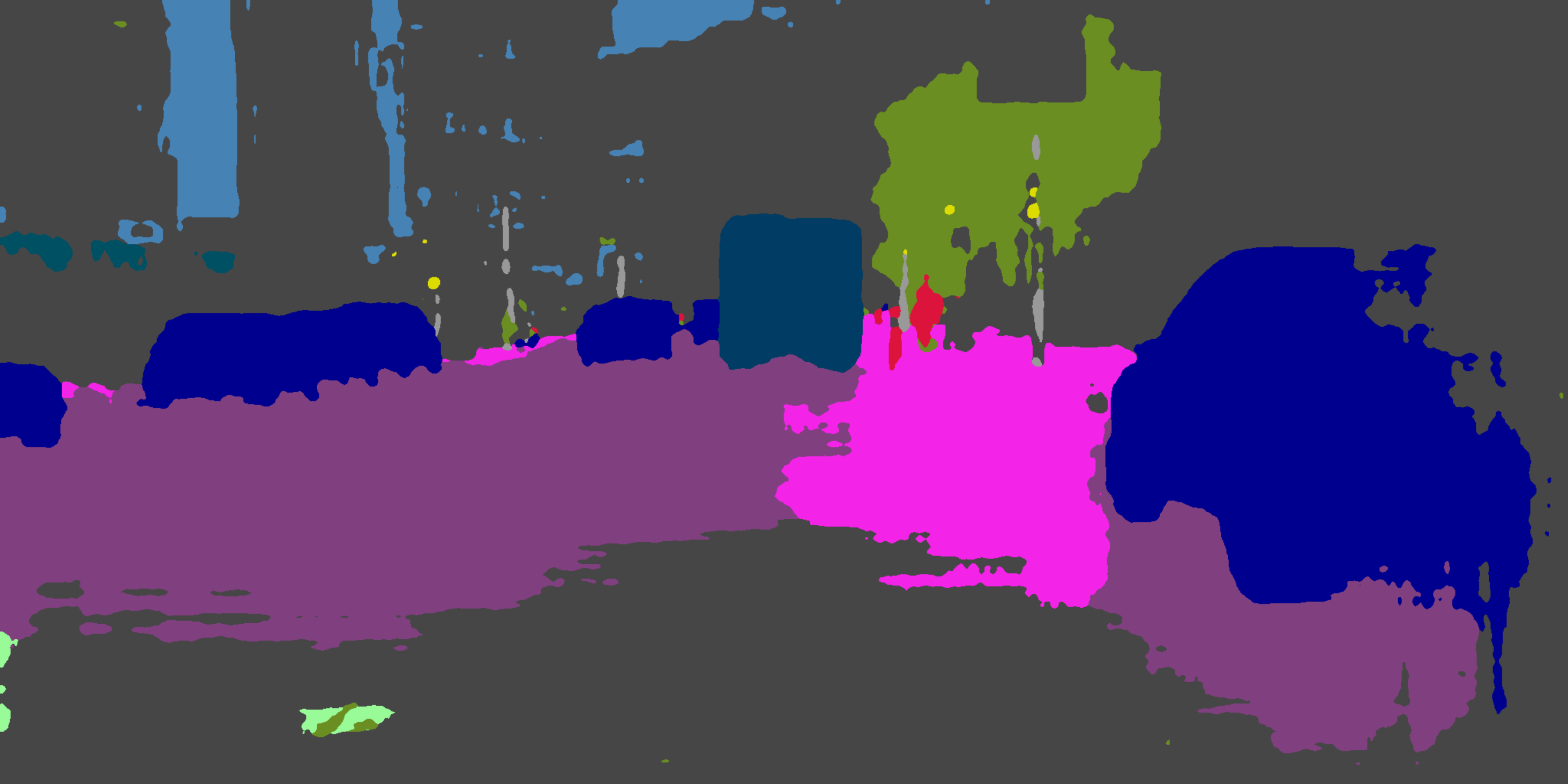}
\includegraphics[width=0.16\textwidth]{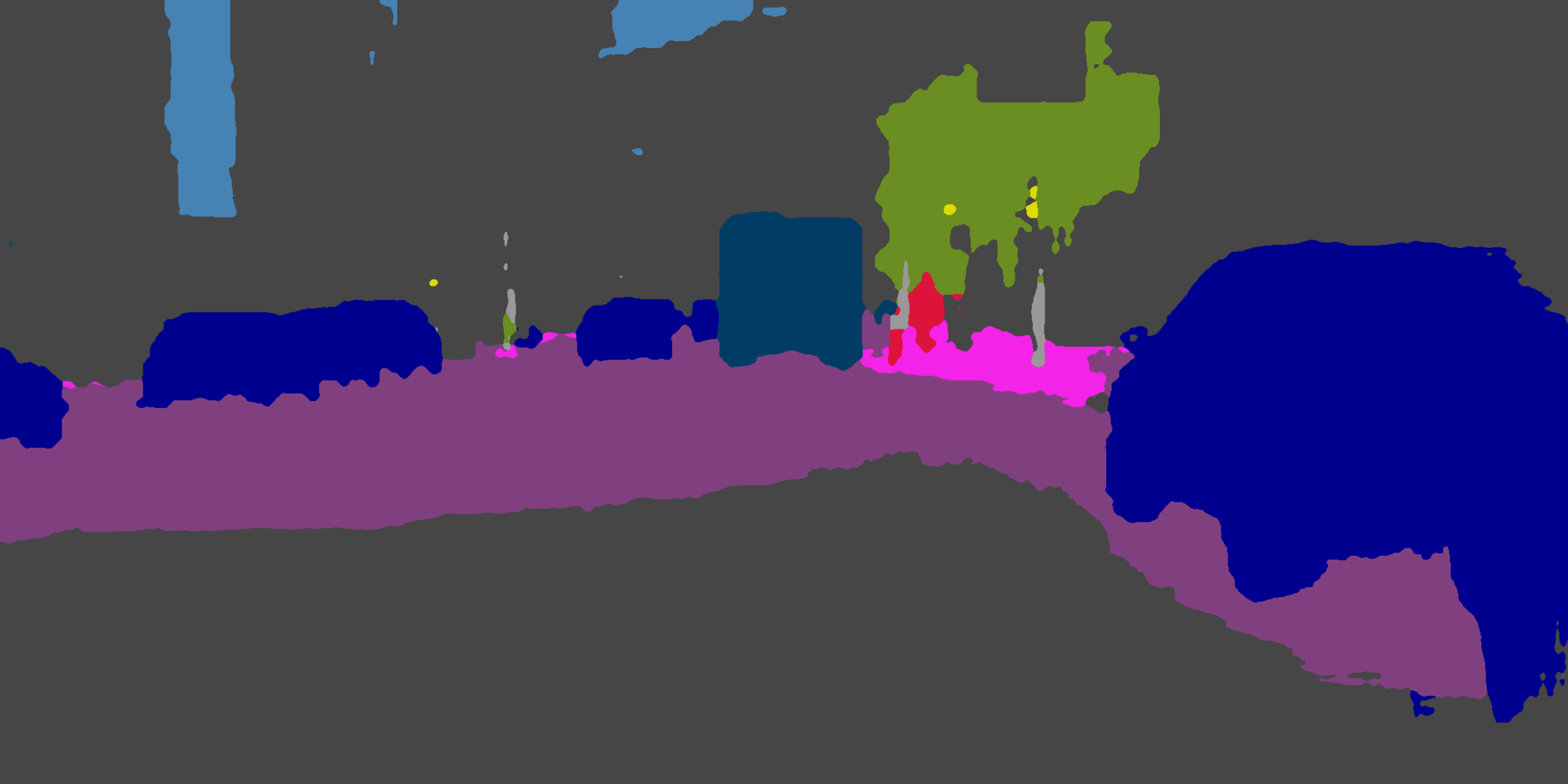}
\includegraphics[width=0.16\textwidth]{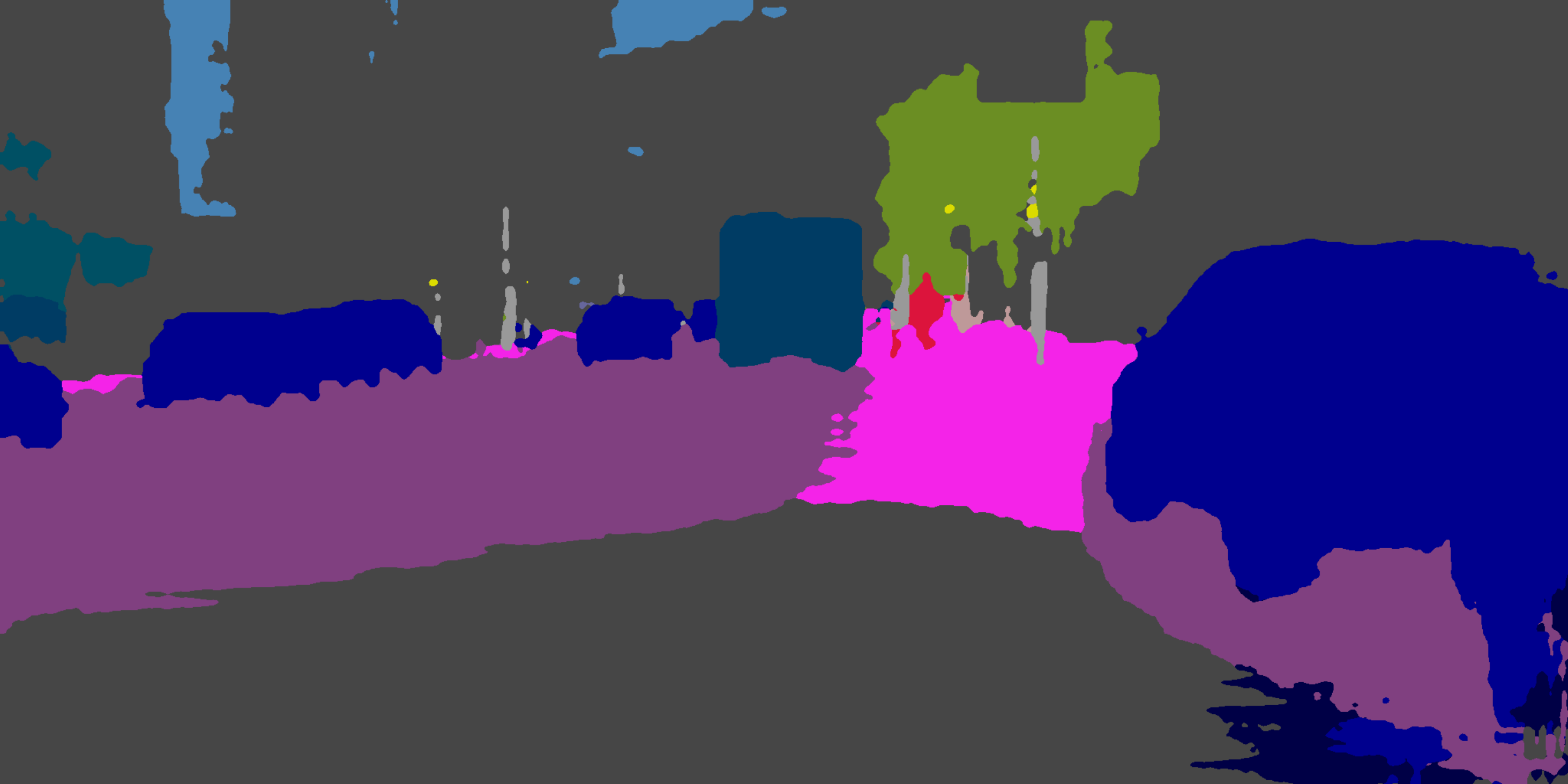}
\includegraphics[width=0.16\textwidth]{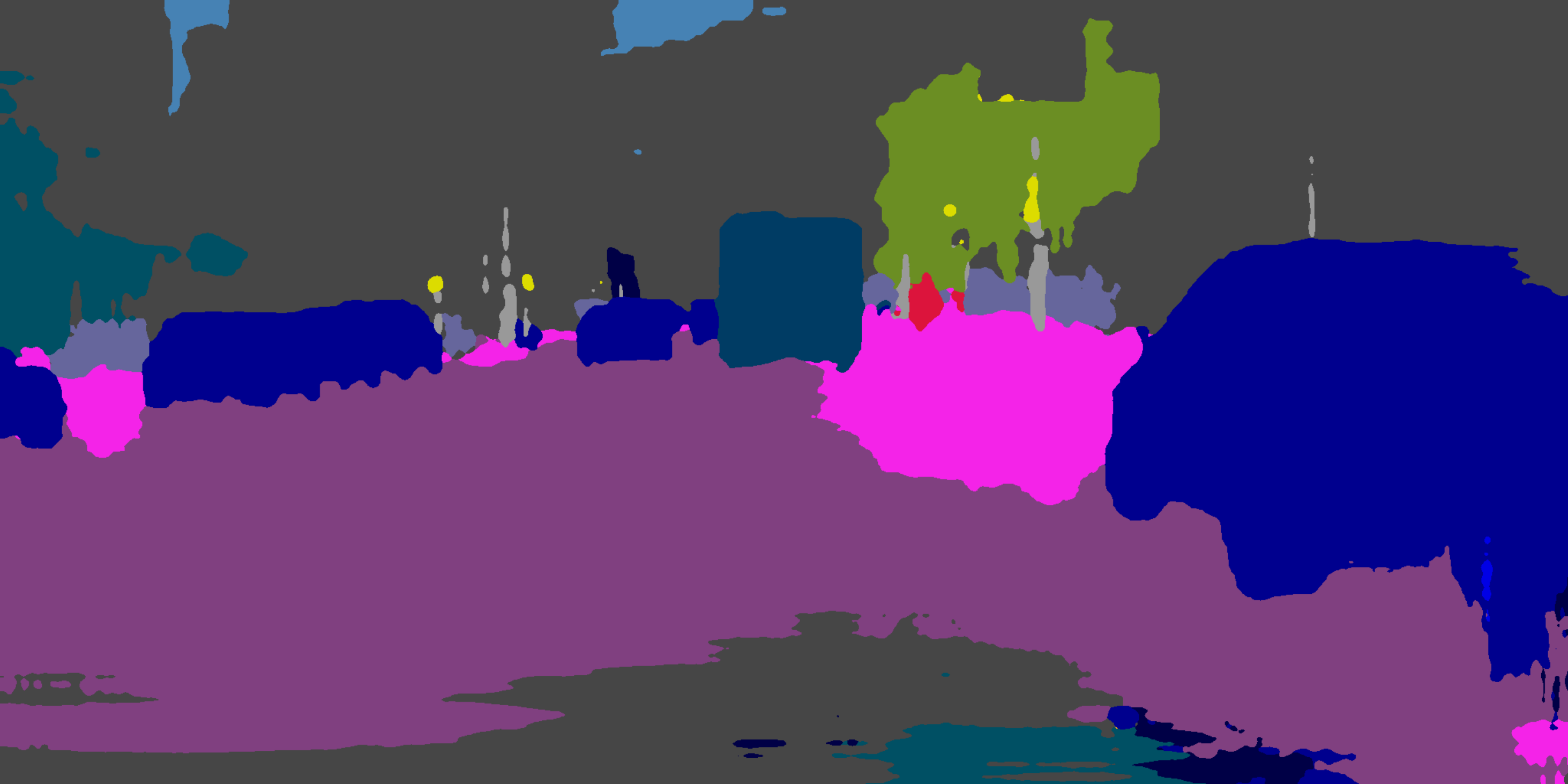}
\caption{Adaptation results on GTA5 $\rightarrow$ Cityscapes. Rows correspond to sampled images and predictions. Columns correspond to original images, ground truth, and results of source ResNet-38, ST, CBST and CBST-SP. Best viewed in color.}\label{gta2city}
\end{figure}

Table \ref{t_gtacity} gives the results on the 19 valid classes. For models with FCN8s-VGG16 as backbone, the performance of ST demonstrates that the adapted model can easily bias to easy-to-transfer classes. However, the CBST not only achieves better mIoU than ST, but also better IoU for these initial hard-to-transfer classes. Since images from GTA5 and Cityscapes have similar structure layouts, we evaluate the performance of CBST-SP, and achieves an mIoU of $36.1\%$ better than some methods using powerful backbones such as ResNet-50~\cite{saito2017adversarial} and DenseNet~\cite{murez2018image}. With a more powerful model ResNet-38, our framework achieves an mIoU of $46.2\%$. Finally, multi-scale testing (MST) at scales 0.5, 0.75, and 1.0 boosts the mIoU to $47.0\%$. Qualitative results on Cityscapes from different comparing methods are visualized in Figure \ref{gta2city}.

\section{Conclusions}
In this paper, we propose a deep self-training frameworks towards unsupervised domain adaptation for semantic segmentation. The framework is formulated as a loss minimization problem and can be learned end-to-end. We also introduce a class-balanced self-training (CBST) framework to overcome the imbalance issue of transferring difficulty among classes via generating pseudo-labels with balanced class distribution. Finally, if both the source and target domains share similar spatial layouts, we can incorporate spatial priors in self-training, which also improves the adaptation quality. In the experiment, we demonstrate that our method achieves good results which outperform some state-of-the-art methods with considerable margins. This empirically suggests that self-training based approaches may provide an effective alternative towards domain adaptation besides adversarial training.

\clearpage
\bibliographystyle{splncs04}
\bibliography{ref}

\section*{Appendix}
The main paper presents our ECCV 2018 camera ready submission. In the appendix, we further present the additional details and results that are not covered by the camera ready paper due to space constraints. We believe these details will benefit successful reproduction of the reported experiments.

\setcounter{section}{0}
\renewcommand{\thesection}{\Alph{section}}

\section{Additional implementation details}
For training FCN8s-VGG16 in our experiments, we use SGD with learning rate of $1 \times 10^{-6}$, batch size of 1, and input image patch with $500 \times 500$ crop size. For training ResNet-38 in our experiments, we use SGD with learning rate of $1 \times 10^{-4}$ and a mini-batch of two image patches with $500 \times 500$ crop size, plus the augmentation of random multi-scale resizing ($0.7 \sim 1.3$) and horizontal flipping. Fine-tuning in each self-training round contains two target epochs. Note that for all experiments in the main paper, we start the pseudo-label selection portion $p$ from $20\%$, and incrementally add $5\%$ to $p$ in each additional round of pseudo-label generation until $p$ reaches $50\%$. All results in the main paper are unified to report the mIoUs of the self-trained models at the 3 round (6 epochs). 

\section{Additional results on GTA5 $\rightarrow$ BDD-V}
\subsection{Ablation Study} 
We show additional ablation studies on GTA5 to Berkeley DeepDrive Video Dataset (BDD-V)~\cite{nexar}, where we evaluate all self-training based methods, including ST, CBST and CBST-SP. In addition, we also evaluate the results of ST-SP which is self-training with spatial priors.

The BDD-V Dataset is a recently released dataset containing 5,561 $1280 \times 720$ images, where the dataset is divided into 3,333 annotated training images, 745 annotated validation images and 1,483 unlabeled test images. The dataset was collected using the NEXAR dashcam interface, and has 41 classes where 19 valid evaluation classes overlap with those in Cityscapes and GTA5. Table~\ref{t_gtabdds} shows the quantitative results of GTA5 to BDD-V validation. The experiment consistently shows the effectiveness of the proposed self-training frameworks.

\begin{table}[t]
\centering
\caption{Ablation study on GTA V $\rightarrow$ BDD-V}\label{t_gtabdds}
	\vspace{1mm}
	\resizebox{\linewidth}{!}{%
		\begin{tabular}{c|cccccccccccccccccccc}
			\hline
			Method           & Road          & SW            & Build         & Wall                              & Fence         & Pole          & TL            & TS            & Veg.          & Terrain       & Sky           & PR            & Rider         & Car           & Truck          & Bus           & Train & Motor         & Bike                              & Mean          \\ \hline
			Source Resnet-38 & 76.7          & 34.1          & 53.8          & 10.2                              & 28.3          & 29.1          & 34.1          & 33.9          & 73.4          & 17.5          & 60.8          & 52.8          & 15.2          & 63.8          & 40.78 & 28.8          & 0.0   & 21.3 & \multicolumn{1}{c|}{2.6}          & 35.0          \\
			ST               & 83.5          & 26.1          & 72.5          & 14.1                              & 27.3          & 26.5          & 32.5          & 28.5          & 74.5          & 35.7          & 88.1          & 51.4          & 15.9          & 67.4          & 26.6           & 35.9          & 0.0   & 8.9           & \multicolumn{1}{c|}{2.9}          & 37.8          \\
			ST-SP            & 88.2          & 40.8 & 74.1          & 14.8                              & 27.1          & 25.8          & 33.1          & 36.1 & 72.2          & 37.4 & 88.8          & 53.8          & 21.2 & 74.2          & 24.5           & 22.9          & 0.0   & 12.9          & \multicolumn{1}{c|}{1.5}          & 39.5          \\
			CBST             & 84.1          & 26.6          & 75.0 & 15.3                              & 28.8 & 28.0          & 33.8          & 29.8          & 76.2          & 35.6          & 90.4 & 54.2          & 18.2          & 69.4          & 28.6           & 36.7 & 0.0   & 13.0          & \multicolumn{1}{c|}{3.8} & 39.3          \\
			CBST-SP          & 89.9 & 39.3          & 73.9          & \multicolumn{1}{c}{14.9}          & 28.0          & 28.7 & 34.1 & 35.6          & 76.7 & 34.9          & 89.6          & 57.4 & 19.8          & 77.3 & 27.1           & 28.1          & 0.0   & 13.8          & \multicolumn{1}{c|}{1.7}          & 40.6 \\ \hline
		\end{tabular}
	}
\end{table}

One could see that both CBST-SP, CBST are better than ST, with CBST-SP being the best among them. Moreover, the class-balanced pseudo label generation strategies particularly benefits the adaptation of classes with small object scales, such as pols, traffic light, pedestrian and motorcycle, etc. We additionally evaluate self-training with spatial priors (ST-SP), and found that the priors also benefit the vanilla self-training but the performance is lower than CBST-SP. This demonstrates the effectiveness of both CBST and SP.

\subsection{Parameter analysis}
\begin{figure}[t]
\centering
\includegraphics[height=0.25\textwidth]{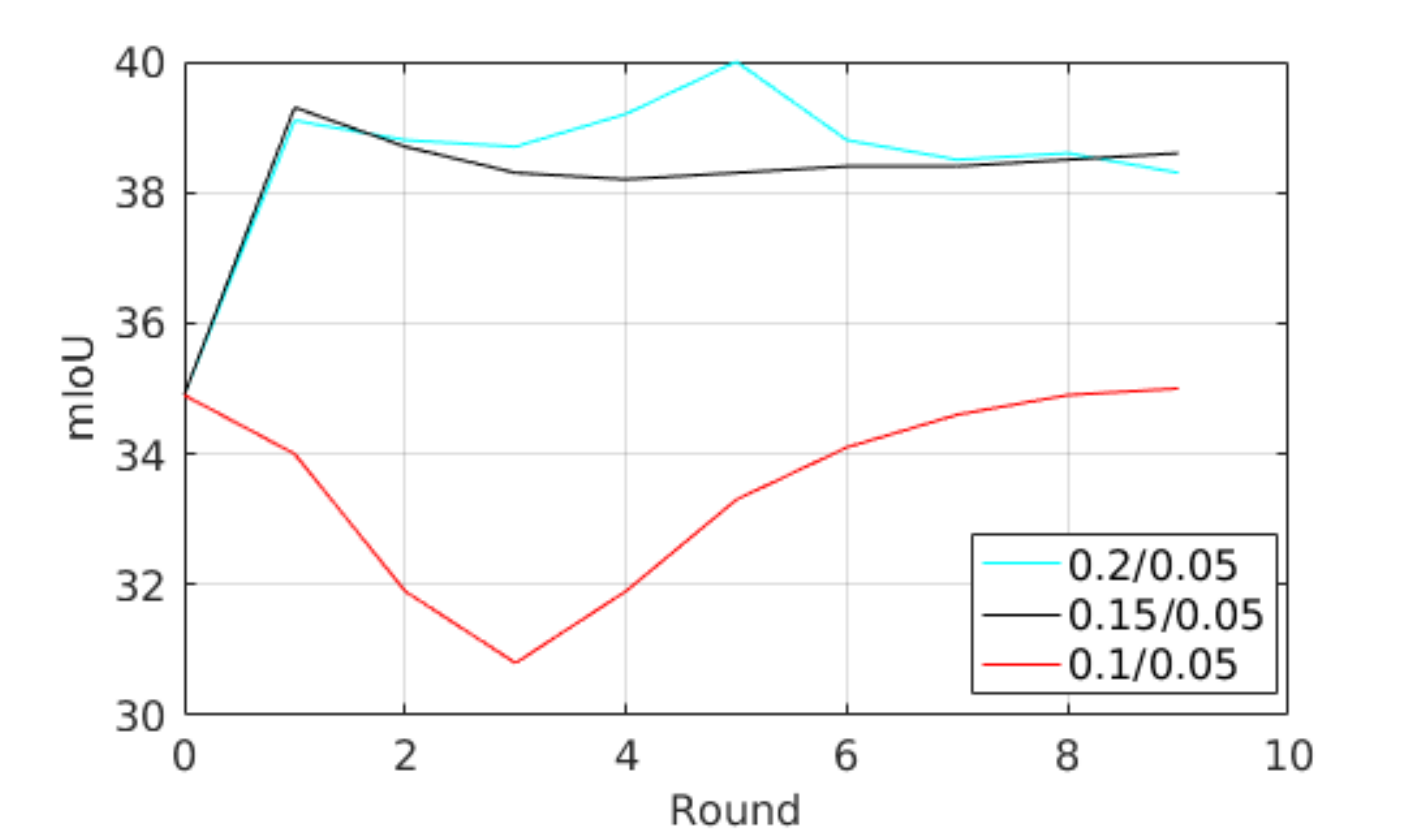}
\includegraphics[height=0.25\textwidth]{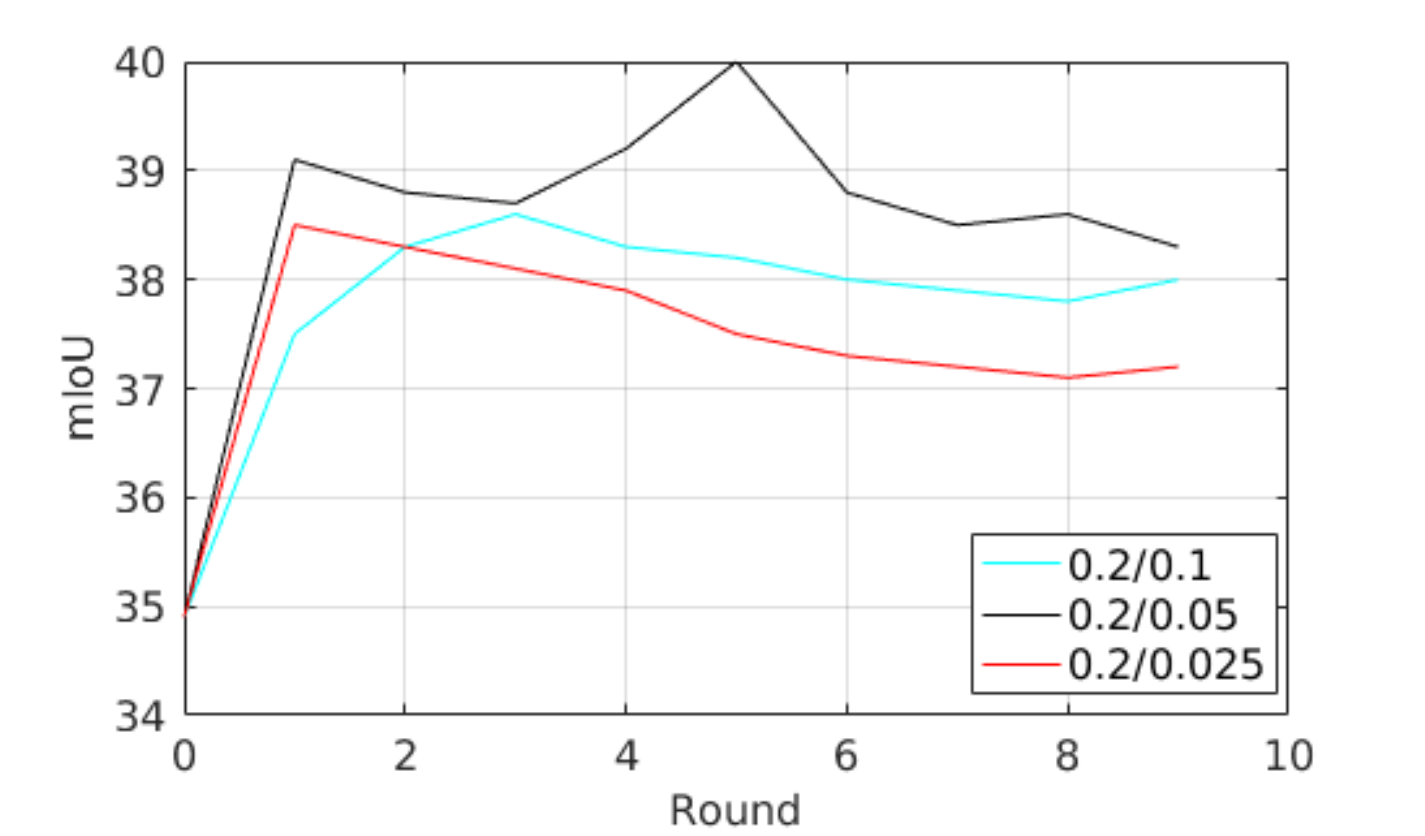}\\
\includegraphics[height=0.25\textwidth]{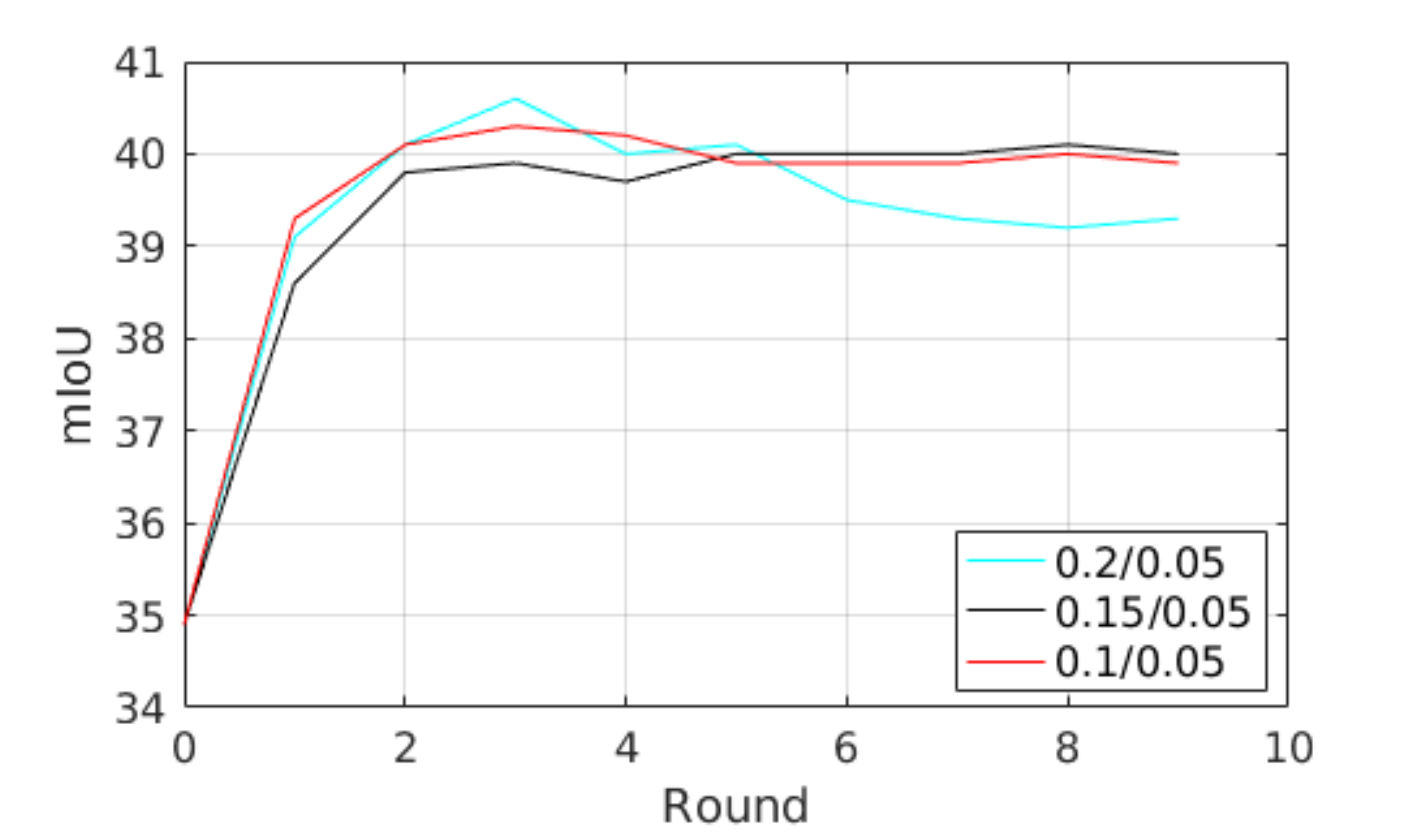}
\includegraphics[height=0.25\textwidth]{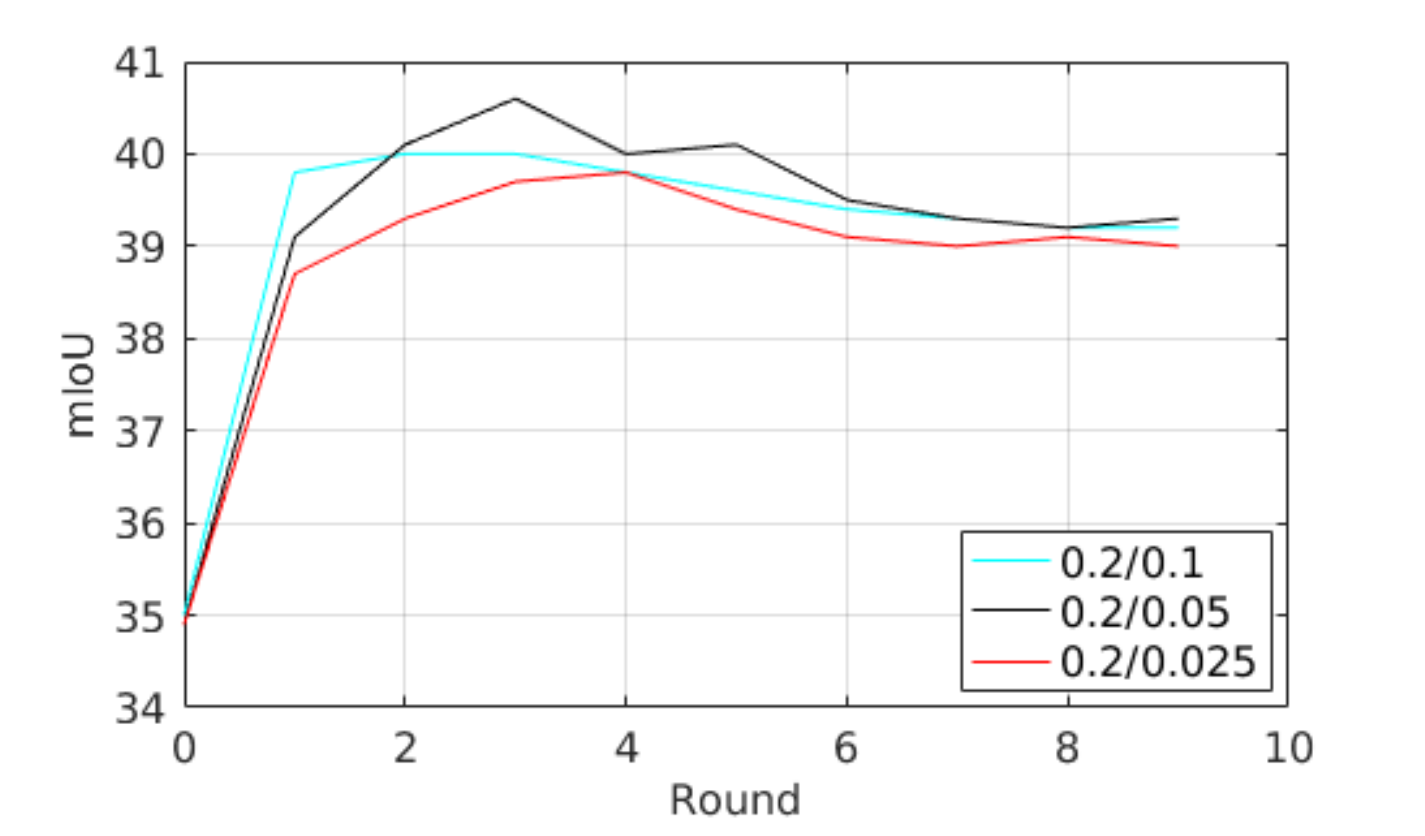}
\caption{Parameter analysis on GTA5 $\rightarrow$ BDD-V. Top: ST-SP. Bottom: CBST-SP. Left: Different $p_0$. Right: Different $\triangle p$. Legends: $p_0/\triangle p$. Best viewed in color.}\label{gtabddv_as}
\end{figure}

We also conduct sensitivity analysis for ST-SP and CBST-SP on the policy parameters of $k$ and $k_c$. Recap that in ST-SP/CBST-SP, both $k$ and $k_c$ are determined by a single policy parameter $p$ which is the portion of pseudo-labels. We conduct analysis by changing both the initial value $p_0$ and the per round increment size $\triangle p$. Fig.~\ref{gtabddv_as} shows the system performance curves of the comparing methods at different self-training rounds with different ($p_0$/$\triangle p$) configurations. It can be observed that CBST-SP is not very sensitive to $k_c$ and shows convergence behavior better than ST-SP.

\end{document}